\documentclass[lettersize,journal]{IEEEtran}
\usepackage{amsmath,amsfonts,amssymb}
\usepackage{algorithm}
\usepackage{algorithmic}
\usepackage{array}
\usepackage{textcomp}
\usepackage{stfloats}
\usepackage{url}
\usepackage{verbatim}
\usepackage{graphicx}
\usepackage{cite}
\usepackage{multirow}
\usepackage{float}
\usepackage{color}
\usepackage{booktabs}
\usepackage[table]{xcolor}
\usepackage[font=footnotesize,labelfont=bf]{caption}
\usepackage{subcaption}
\usepackage{hyperref}    
\hypersetup{
    colorlinks=true,
    linkcolor=red,          
    citecolor=blue,          
    urlcolor=red           
}

\usepackage{tikz,xcolor,hyperref}
\definecolor{lime}{HTML}{A6CE39}
\DeclareRobustCommand{\orcidicon}{%
    \begin{tikzpicture}
    \draw[lime, fill=lime] (0,0) 
    circle [radius=0.16] 
    node[white] {{\fontfamily{qag}\selectfont \tiny ID}};    \draw[white, fill=white] (-0.0625,0.095) 
    circle [radius=0.007];    \end{tikzpicture}
    \hspace{-2mm}}
\foreach \x in {A, ..., Z}{%
    \expandafter\xdef\csname orcid\x\endcsname{\noexpand\href{https://orcid.org/\csname orcidauthor\x\endcsname}{\noexpand\orcidicon}}
    }

\begin{document}

\title{
From Calibration to Refinement: Seeking Certainty via Probabilistic Evidence Propagation for Noisy-Label Person Re-Identification
}

%


\author{Xin Yuan\orcidA{},~\IEEEmembership{Member,~IEEE}, Zhiyong Zhang\orcidB{}, Xin Xu\orcidC{},~\IEEEmembership{Senior Member,~IEEE}, Zheng Wang\orcidD{},~\IEEEmembership{Senior Member,~IEEE}, Chia-Wen Lin\orcidE{},~\IEEEmembership{Fellow,~IEEE}
\thanks{
This research was financially supported by funds from National Natural Science Foundation of China (No. 62376201), Nature Science Foundation of Hubei Province (No. 2025AFB056), Key Project of the Science Research Plan of the Hubei Provincial Department of Education (No. D20241103), and Hubei Provincial Science and Technology Talent Service Enterprise Project (No. 2025DJB059).
(\textit{Corresponding author: Xin Xu})}
\thanks{Xin Yuan, Zhiyong Zhang, and Xin Xu are with School of Computer Science and Technology, Wuhan University of Science and Technology, Wuhan, Hubei 430065, China (email: xinyuan@wust.edu.cn, zhiyong@wust.edu.cn, xuxin@wust.edu.cn). Xin Yuan and Xin Xu are also with Hubei Province Key Laboratory of Intelligent Information Processing and Real-Time Industrial System, Wuhan University of Science and Technology, Wuhan 430065, China.}
\thanks{Zheng Wang is with School of Computer Science, Wuhan University, Wuhan, Hubei 430072, China (email: wangzwhu@whu.edu.cn).}
\thanks{Chia-Wen Lin is with Department of Electrical Engineering and the Institute of Communications Engineering, National Tsing Hua University, and with Electronic and Optoelectronic System Research Laboratories, Industrial Technology Research Institute, Hsinchu, Taiwan 100084‌, China (email: cwlin@ee.nthu.edu.tw).}
}

\markboth{ IEEE Transactions on Multimedia,~Vol.~XX, No.~XX,~2026}%
{Shell \MakeLowercase{\textit{et al.}}: A Sample Article Using IEEEtran.cls for IEEE Journals}


\maketitle

\begin{abstract}
With the increasing demand for robust person re-identification (Re-ID) in unconstrained environments, learning from datasets with noisy labels and sparse per-identity samples remains a critical challenge. Existing noise-robust person Re-ID methods primarily rely on loss-correction or sample-selection strategies using softmax outputs. However, these methods suffer from two key limitations: 1) Softmax exhibits translation invariance, leading to over-confident and unreliable predictions on corrupted labels. 2) Conventional sample selection based on small-loss criteria often discards valuable hard positives that are crucial for learning discriminative features. To overcome these issues, we propose the \underline{\textbf{CA}}libration-to-\underline{\textbf{RE}}finement (\textbf{CARE}) method, a two-stage framework that seeks certainty through probabilistic evidence propagation from calibration to refinement. 
In the calibration stage, we propose the probabilistic evidence calibration (PEC) that dismantles softmax translation invariance by injecting adaptive learnable parameters into the similarity function, and employs an evidential calibration loss to mitigate overconfidence on mislabeled samples. 
In the refinement stage, we design the evidence propagation refinement (EPR) that can more accurately distinguish between clean and noisy samples.  
Specifically, the EPR contains two steps: Firstly, the composite angular margin (CAM) metric is proposed to precisely distinguish clean but hard-to-learn positive samples from mislabeled ones in a hyperspherical space;
Secondly, the certainty-oriented sphere weighting (COSW) is developed to dynamically allocate the importance of samples according to CAM, ensuring clean instances drive model updates. 
Extensive experimental results on Market1501, DukeMTMC-ReID, and CUHK03 datasets under both random and patterned noises show that CARE achieves competitive performance. 
In addition, ablation studies validate the effectiveness of each component and demonstrate superior robustness to noisy labels.
The code is available at https://github.com/YuanXinCherry/CARE.

\end{abstract}

\begin{IEEEkeywords}
Person Re-identification, Label Noise, Probabilistic Evidence Propagation, Noise Robust Learning.
\end{IEEEkeywords}

\section{Introduction}
Person re-identification (Re-ID) addresses the critical task of matching individuals across disjoint camera networks, which is essential for intelligent surveillance and security systems~\cite{xie2022sampling,xu2022rank,yuan2023searching}.
While modern deep learning models achieve high accuracy under controlled conditions, their performance relies heavily on large-scale datasets with precise identity annotations.
In the person Re-ID applications, however, label noise inevitably arises from two primary sources: 1) imperfect automated detection pipelines causing misaligned or fragmented pedestrian bounding boxes, and 2) inconsistent human annotations due to viewpoint variations and occlusions~\cite{yu2019robust,ye2020purifynet,ye2022collaborative}.
In addition, unlike the typical classification task where noisy labels can be mitigated through data abundance, the person Re-ID task faces the acute challenge of sparse per-identity samples, \textit{i.e.}, typically fewer than 30 images per identity~\cite{ye2020purifynet}.
This data scarcity exacerbates the negative impact of noisy labels, as even minor annotation errors distort the learning of subtle discriminative features~\cite{wei2020combating}, ultimately degrading the model performance in complex environments.

\begin{figure*}[ht]
  \centering
  \includegraphics[width=0.855\linewidth]{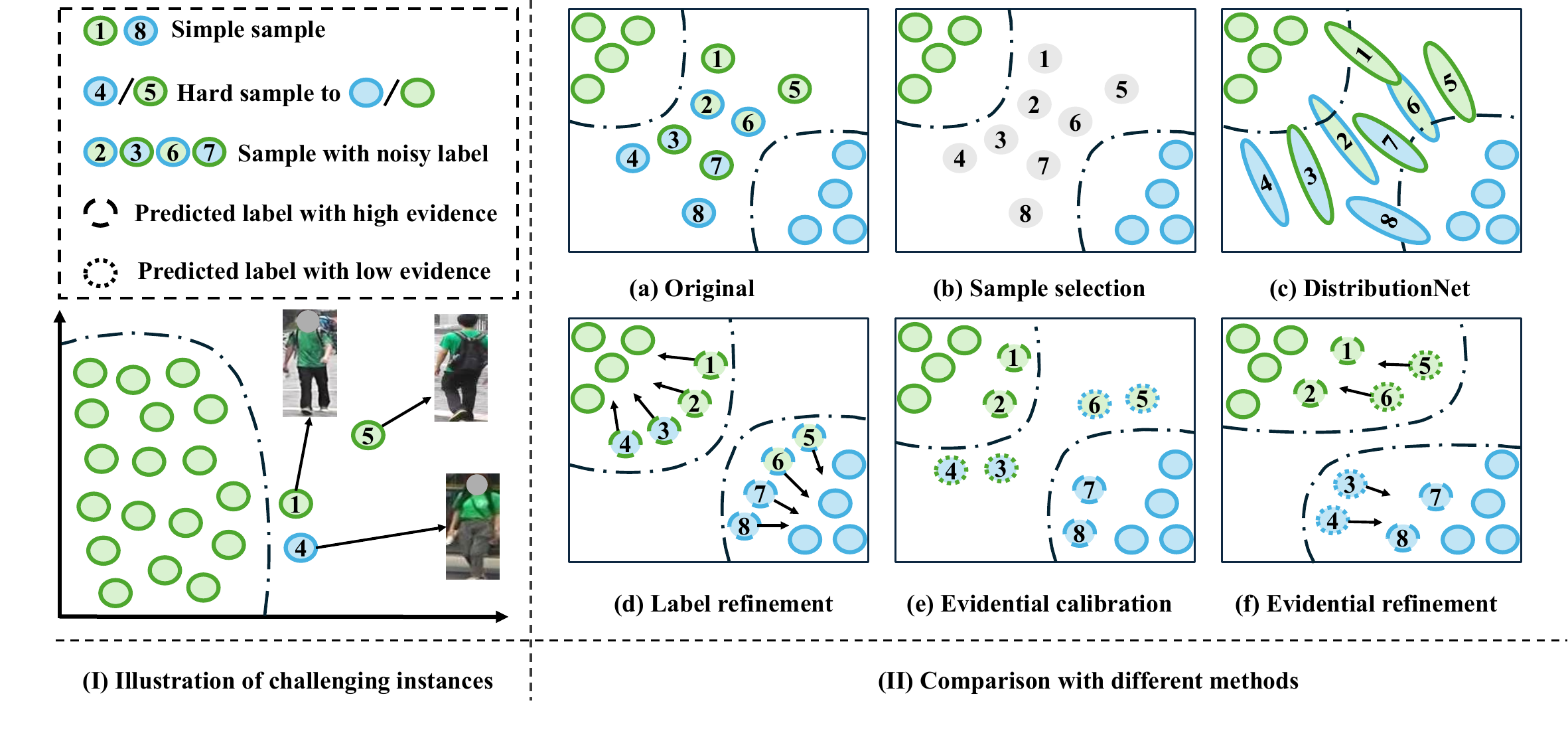}
  \caption{Illustration of core idea.
  \textbf{Left:} two challenging instances in feature space, the sample 4 (clean but close to another identity) and the sample 5 (hard sample with occlusion). \textbf{Right:} (a) Original samples with noisy labels; (b) Sample selection methods filter out noisy but informative samples~\cite{liang2024combating, malach2017decoupling, karim2022unicon}; (c) DistributionNet uses uncertainty to model features, yet it still confuses similar features between clean and noisy labels~\cite{yu2019robust};
  (d) Label refinement methods based on softmax may produce the same probabilities for different samples, resulting in incorrectly refurbished labels~\cite{ye2020purifynet,ye2022collaborative,chen2023refining,zhong2023neighborhood}; 
  \textbf{CARE} contains (e) evidential calibration and (f) evidential refinement: (e) calibrates the high evidential instances in the \textbf{Calibration} stage,
  and (f) will refine the low evidential instances in the \textbf{Refinement} stage.
  \textit{Best viewed in color}.}
  \label{fig:motivation}
  \vspace{-5mm}
\end{figure*}

Existing noise-robust paradigms from image classification typically filter noisy samples using loss-correction \cite{patrini2017making, zhang2018generalized, shu2019meta} or sample-selection \cite{ malach2017decoupling, karim2022unicon,liang2024combating} strategies based on softmax outputs~\cite{han2018co,lee2018cleannet}. Although these strategies are effective in data-abundant domains, they are ill-suited for person Re-ID owing to the vital role of each training instance in sparse identity learning. Meanwhile, overly aggressive removal of samples may result in the loss of learning discriminative features from valuable hard instance samples, whereas retaining wrongly labeled instances may disturb the metric learning objective~\cite{ye2020purifynet}. Moreover, conventional confidence estimation via softmax probabilities exhibits translation invariance, which causes overconfident predictions on corrupted samples~\cite{zong2024dirichlet}. These limitations prove especially damaging for person Re-ID models, which must distinguish fine-grained appearance patterns under substantial intra-identity variations~\cite{wang2018learning}.

To overcome the fundamental limitations of naive filtering (\textit{e.g.}, indiscriminate hard-positive discarding) and unreliable small-loss heuristics (\textit{e.g.}, softmax-induced overconfidence~\cite{zong2024dirichlet}) in noisy-label Re-ID (as shown in Fig.~\ref{fig:motivation} \textcolor{red}{II-(b)}), recent advances have pivoted toward integrated optimization paradigms that co-optimize label refinement and representation learning. Contemporary research addresses noisy-label person Re-ID through four distinct methodological approaches: 1) Probabilistic Feature Modeling, As shown in Fig.~\ref{fig:motivation} \textcolor{red}{II-(c)}, DistributionNet~\cite{yu2019robust} pioneers Gaussian distribution representations of samples, where feature variance explicitly quantifies uncertainty to attenuate noise influence; 2) Integrated Label-Representation Optimization, PurifyNet~\cite{ye2020purifynet} and ICLR~\cite{zhong2024iclr} establish a unified framework jointly optimizing label correction and feature learning, incorporating informative sample reweighting to preserve discriminative samples; 3) Cross-Network Collaborative Refinement, CORE~\cite{ye2022collaborative} and TSNT~\cite{liu2024two} deploy mutual distillation between peer networks, iteratively refining annotations through bidirectional knowledge transfer; 4) Neighborhood-Consistency Adaptation, LRP~\cite{chen2023refining} and LRNI-HSR~\cite{zhong2023neighborhood} dynamically adjust label confidence by exploiting multi-granularity neighborhood agreements and hierarchical prototype alignment.
While these approaches partially alleviate issues stemming from overly aggressive removal of samples (\textit{e.g.}, mitigated discarding of hard positives shown in Fig.~\ref{fig:motivation} \textcolor{red}{II-(d)}), three unresolved limitations persist:
First, Unconstrained Noise Propagation: Joint frameworks treating most samples as reliable propagate residual label noise, severely amplified under sparse per-identity samples; Second, Early-Stage Noise Accumulation: Lacking initial calibration, erroneous signals compound before sufficient corrective evidence emerges; Third, Overconfident Similarity Estimation: Softmax confidences inherit translation invariance, inducing excessive certainty on camera/occlusion artifacts that cause premature convergence to spurious patterns.
In addition, the intrinsic semantic ambiguity of Re-ID data compromises deterministic predictions~\cite{li2025spcl,li2024learning}, necessitating models that are robust to noisy labels and capable of quantifying predictive uncertainty.

To this end, we propose the \underline{\textbf{CA}}libration-to-\underline{\textbf{RE}}finement (\textbf{CARE}) method, a two-stage calibration-to-refinement framework for noisy-label person Re-ID.
This work is designed to tackle two interrelated yet distinct challenges: \textit{unreliable confidence estimation} and \textit{biased sample selection}.
The core idea is shown in Fig.~\ref{fig:motivation} \textcolor{red}{II-(e)} and \textcolor{red}{II-(f)}. 
In the \textbf{Calibration} stage, we present Probabilistic Evidence Calibration (PEC), a Dirichlet-informed prediction calibration that breaks softmax translation invariance by injecting adaptive, learnable terms into the similarity computation and employs an evidential calibration loss to alleviate over-confident predictions on corrupted labels. 
This \textbf{Calibration} stage provides a well-calibrated, evidence-based foundation for sample assessment.
Building upon this foundation, the \textbf{Refinement} stage introduces Evidence Propagation Refinement (EPR). 
First, the Composite Angular Margin (CAM) metric measures angular separability on a hypersphere to precisely distinguish clean but hard-to-learn positives from mislabeled instances; then, Certainty-Oriented Sphere Weighting (COSW) dynamically allocates sample importance based on CAM-derived certainty metrics, amplifying the contribution of genuinely clean examples while suppressing noise. 
This two-stage framework establishes a synergistic loop: precise calibration enables unbiased sample refinement, which then yields purified supervision signals, thereby facilitating robust representation learning.
Our contributions are as follows: 
\begin{itemize}
  \item We propose \textbf{CARE}, a two-stage calibration-to-refinement framework that first calibrates prediction uncertainty and then refines sample contributions for noisy-label person Re-ID, which preserves informative hard positives and substantially improves robustness and generalization.
  \item We design Probabilistic Evidence Calibration (PEC), a Dirichlet-informed calibration with evidential calibration loss that breaks softmax translation invariance and produces more reliable uncertainty estimates on noisy labels.
  \item We introduce Evidence Propagation Refinement (EPR), which leverages the Composite Angular Margin (CAM) and Certainty-Oriented Sphere Weighting (COSW) to separate clean but hard-to-learn positives from mislabeled instances in a hyperspherical space and allocate training sample importance accordingly.
  \item Extensive experimental results demonstrate the competitiveness of \textbf{CARE} compared with state-of-the-art methods with noisy labels on three public datasets under both random and patterned label noise settings. 
\end{itemize}


The rest of this paper is organized as follows: 
Section \ref{sec: Related Work} reviews related works. Section \ref{sec:The Proposed Method} details the proposed method. Section \ref{sec:Experiments} presents extensive experiments, including comparisons with state-of-the-art methods and comprehensive ablation studies. Finally, Section \ref{sec:Conclusion} concludes the paper.

\section{Related Work}
\label{sec: Related Work}
\subsection{Deep Person Re-ID Methods}
Deep person re-identification (Re-ID) aims to retrieve the same person's identity across non-overlapping cameras and has advanced rapidly with the use of deep convolutional neural networks~\cite{xie2022sampling,xu2022rank,yuan2023searching,yu2019robust,ye2020purifynet,ye2022collaborative}. Existing methods are typically grouped into two complementary paradigms. The first paradigm focuses on feature representation learning, which builds robust cross-view descriptors by combining global and local part cues, attention mechanisms, or multi-branch architectures to capture fine-grained appearance patterns~\cite{cheng2024neighbor,chen2021person,wang2018learning}. The second paradigm emphasizes distance metric learning, which learns discriminative similarity measures to reinforce intra-class compactness and enlarge inter-class separability~\cite{xu2022rank,yuan2023searching,yan2021beyond}. Although these supervised person Re-ID approaches achieve good performance under clean annotations, they typically rely on abundant and accurate labels per identity~\cite{xie2022sampling,xu2022rank,yuan2023searching}. Unsupervised person Re-ID, which instead estimates pseudo-labels through clustering to learn from unlabeled data, offers an alternative~\cite{pang2022camera, wang2021camera, pang2025robust}. However, this paradigm may introduce noisy pseudo-labels that can undermine the robustness of the learned Re-ID model. Furthermore, Re-ID data exhibit intrinsic semantic ambiguity due to intra-individual polymorphism and inter-individual commonality, which can be addressed by semantic polymorphism and commonality learning~\cite{li2025spcl,li2024learning} through the modeling of associative relationships to facilitate the learning of distinguishable semantics.
Nevertheless, all the aforementioned methods will suffer from severe degradation in retrieval accuracy when encountering annotation errors.


\subsection{Deep Learning with Noisy-Label}
Existing deep learning with noisy-label methods can be broadly classified into two categories: loss-correction~\cite{patrini2017making,zhang2018generalized, li2024regroup} or sample-selection~\cite{liang2024combating, malach2017decoupling, karim2022unicon}. The former intends to correct the loss by estimating the noise transition matrix, rectifying sample labels, or weights to ensure that the classifier can converge to clean data via mitigating the impact of noisy labels ~\cite{patrini2017making,zhang2018generalized, shu2019meta}. The latter strives to identify and remove noisy samples or reduce the weight of potentially noisy samples based on the small-loss criterion~\cite{garcia2016noise, malach2017decoupling, karim2022unicon}. Despite their effectiveness in standard image classification, these techniques frequently rely on sufficient per-class samples and softmax confidence, serving as proxies for label correctness~\cite{han2018co,lee2018cleannet, lin2020focal,wang2019symmetric,li2021learning,sheng2024enhancing}. However, these assumptions are inapplicable in the person Re-ID task, where identity multiplicity, severe class imbalance, per-identity sample scarcity, and distribution shift vulnerability of softmax confidences collectively undermine the person Re-ID model robustness.

\begin{table*}[t]
\centering
\small
\resizebox{0.82\linewidth}{!}{
\begin{tabular}{lll|c lll}
\hline
\textbf{Notation} & \textbf{Shape/Domain} & \textbf{Description} & & \textbf{Notation} & \textbf{Shape/Domain} & \textbf{Description} \\
\hline
$\mathcal{D}$ & $\{(x_i,y_i)\}_{i=1}^N$ & training dataset & & $N$ & scalar & number of samples \\
$C$ & scalar & number of identity classes & & $x_i$ & $\mathcal{X}$ & $i$-th input image \\
$y_i$ & $\{1,\dots,C\}$ & identity label & & $\mathrm{z}_{ij}$ & scalar & logit score \\
$\mathrm{z}_i$ & $\mathbb{R}^{C}$ & logit vector & & $s_j$ & scalar $\ge0$ & smoothing term \\
$\Theta$ & network params & backbone parameters & & $\varphi(\cdot)$ & mapping & evidence mapping \\
$\upsilon_i$ & $\mathbb{R}_{\ge0}^C$ & evidence vector & & $\kappa$ & scalar $>0$ & concentration offset \\
$\mu_i$ & $\mathbb{R}_{>0}^C$ & Dirichlet parameters & & $S_i$ & scalar & concentration sum \\
$\mathcal{I}_{ij}$ & $\{0,1\}$ & one-hot indicator & & $\psi(\cdot)$ & function & digamma function \\
$\mathcal{L}_{\mathrm{ENLL}}$ & loss scalar & expected NLL & & $\mathcal{L}_{\mathrm{KL}}^{(Dir)}$ & loss scalar & KL divergence \\
$\lambda$ & scalar & trade-off coefficient & & $\alpha,\beta$ & scalars & CAM scaling coefficients \\
$\Delta^{(t)}(x_i)$ & scalar & angular separation & & $\Lambda^{(t)}(x_i)$ & scalar & top-$k$ spread \\
$\mathcal{S}^{(t)}(x_i)$ & scalar & instant CAM score & & $\mathrm{CAM}_t(x_i)$ & scalar & accum. CAM score \\
$w_j$ & scalar & epoch weight & & $T$ & integer & total epochs \\
$\phi(x_i;\Theta)$ & $\mathbb{R}^d$ & embedding feature & & $w_q$ & $\mathbb{R}^d$ & class prototype \\
$d_{q}^{(t)}(x_i;\Theta)$ & scalar & angular distance & & $\widetilde{\Delta}^{(t)}$ & scalar & hyperspherical analog \\
$\widetilde{\Lambda}^{(t)}$ & scalar & hyperspherical analog & & $\mathcal{S}_{\mathrm{COSW}}^{(t)}(x_i)$ & scalar & COSW certainty score \\
$\mathcal{R}(\cdot)$ & scalar & reweighting function & & $p_k(j\mid x_i)$ & $[0,1]$ & predictive probability \\
$\Theta_1,\Theta_2$ & network params & peer networks & & $\mathcal{L}_{\mathrm{WCE}}$ & loss scalar & weighted CE \\
$\mathcal{L}_{\mathrm{WKL}}$ & loss scalar & weighted KL & & $D_{\mathrm{WKL}}(\cdot\|\cdot)$ & scalar & one-sided WKL \\
\hline
\end{tabular}}
\caption{Summary of important notations used in the paper.}  
\label{tab:symbols}
\vspace{-4mm}
\end{table*}

\subsection{Deep Person Re-ID with Noisy-Label}
Recent efforts have adapted noisy-label techniques to the person Re-ID problem by introducing probabilistic modelling, joint label–model optimization, neighbourhood-aware refinement, and credibility-driven reweighting. 
For the first type, 
Yu \textit{et al.}~\cite{yu2019robust} proposed DistributionNet, which models each image as a Gaussian distribution instead of a point embedding, adaptively mitigating the impact of noisy instances by leveraging larger variances for uncertain samples.
For the second type, 
Ye \textit{et al.}~\cite{ye2020purifynet} introduced PurifyNet, which integrates label refinement and representation learning through an alternating optimization framework. This unified loop incorporates hard-aware reweighting to retain informative hard examples during label correction.
Subsequently, Zhong \textit{et al.}~\cite{zhong2024iclr} proposed ICLR, a framework that leverages partially credible labels to simultaneously refine and re-weight unreliable labels.
For the third type, 
Ye \textit{et al.}~\cite{ye2022collaborative} developed CORE, a collaborative co-refinement paradigm where peer networks exchange reliability signals to mutually refine pseudo-labels through distillation. 
After that, Liu \textit{et al.}~\cite{liu2024two} introduced TSNT, a noise-robust Re-ID framework that integrates self-refining and co-training strategies to prioritize refurbished reliable samples via weighted learning and reliable triplet mining.
For the fourth type, 
Chen \textit{et al.}~\cite{chen2023refining} developed LRP, leveraging fused global-local features and multi-modal neighbor consistency checks to assess label reliability and refine unreliable annotations.
Zhong \textit{et al.}~\cite{zhong2023neighborhood} proposed LRNI-HSR, enhancing neighbor-based refinement with hierarchical semantic prototypes for progressive label rectification in sparse identity scenarios.
However, these methods still rely on softmax confidences or small-loss heuristics for noise proxies, which are vulnerable to softmax translation invariance and prone to overconfident misclassification or failure to retain hard positives that are essential for sparse per-identity learning. Inspired by these, our method uses probabilistic evidence calibration to correct the problem of softmax over-confidence and performs evidence propagation refinement to the effective learning of clean but hard-to-learn samples during the training process.



\section{The Proposed Method}
\label{sec:The Proposed Method}

\subsection{The Overall Framework}
 
This paper proposes the \underline{\textbf{CA}}libration-to-\underline{\textbf{RE}}finement (\textbf{CARE}) method, a two-stage calibration-to-refinement framework for noisy-label person Re-ID. In the \textbf{Calibration} stage, we introduce Probabilistic Evidence Calibration (PEC), a Dirichlet-informed calibration that injects adaptive, learnable scale and shift terms into the similarity computation to break softmax translation invariance, and employs an evidential calibration loss to temper over-confident predictions on corrupted labels. In the \textbf{Refinement} stage, Evidence Propagation Refinement (EPR) leverages the Composite Angular Margin (CAM) metric to separate clean but hard positive instances from mislabeled instances in a hyperspherical embedding, and applies Certainty-Oriented Sphere Weighting (COSW) to allocate sample importance according to CAM-derived certainty, amplifying genuinely clean instances while suppressing noisy ones. 

Collectively, PEC and EPR constitute an integrated calibration-to-refinement framework that preserves discriminative hard positive instances while significantly enhancing robustness and discriminability under noisy labels and sparse identity learning.
For clarity, the important notations used in this paper are summarized in Table~\ref{tab:symbols}.
The corresponding motivation of the proposed \textbf{CARE} is visualized in Fig.~\ref{fig:motivation}.
In detail, our proposed \textbf{CARE} framework is illustrated in Fig.~\ref{fig:framework}.  

\begin{figure}[t]
  \centering
  \includegraphics[width=0.9\linewidth]{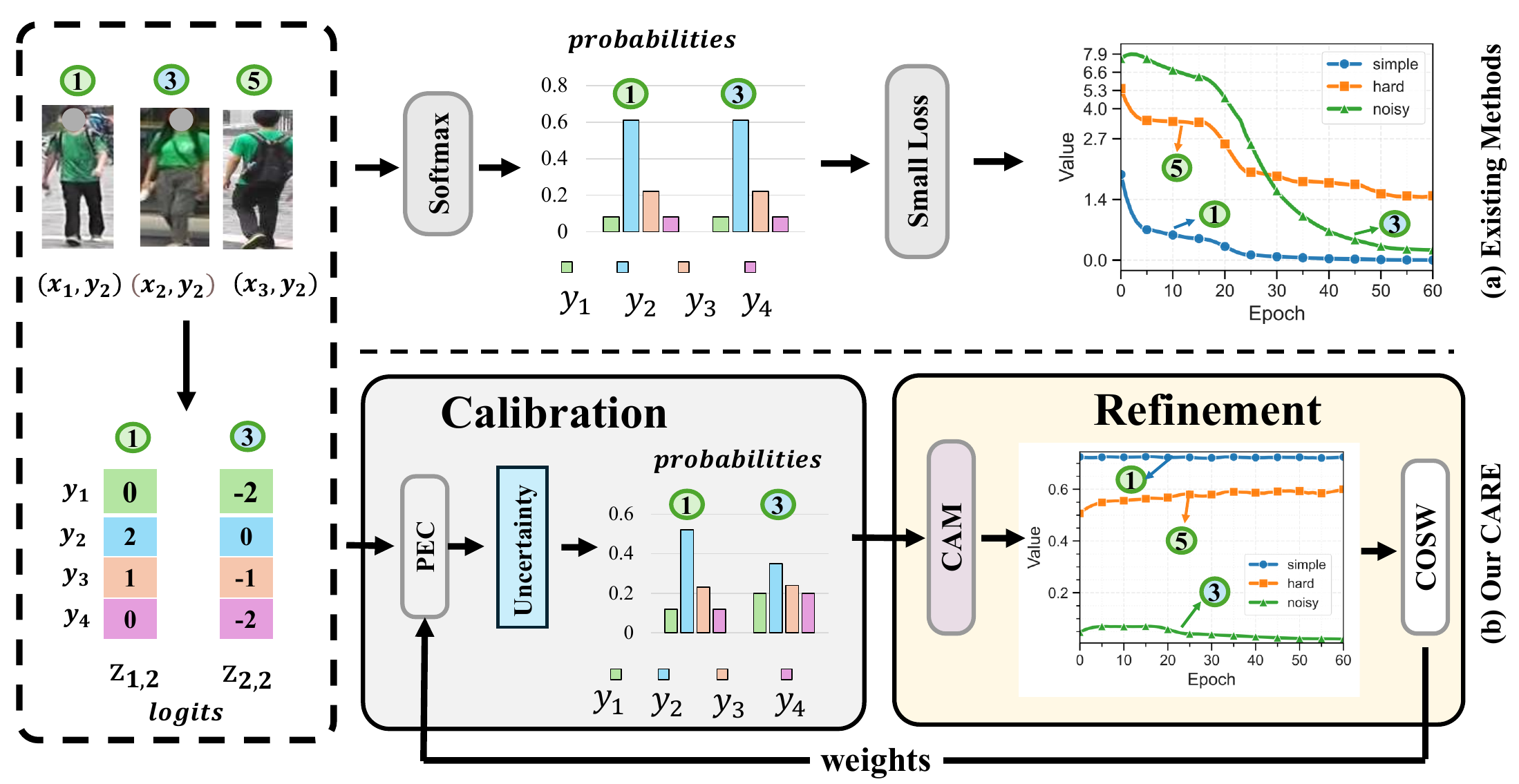}
  \caption{
  Illustration of the core problems and our solution under 20\% random noise. (a) \textbf{Top row}: conventional softmax-based scoring yields over-confident predictions on corrupted labels, while small-loss selection tends to discard informative but hard positives.
  (b) \textbf{Bottom row}: Our two-stage \textbf{CARE} framework addresses these problems by first calibrating uncertainty to isolate noise in the \textbf{Calibration} stage, then refining with angular metrics to preserve hard positives through soft weighting in the \textbf{Refinement} stage. 
  Simple, noisy, and hard positive samples are marked as {\includegraphics[width=0.032\linewidth]{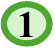}}, {\includegraphics[width=0.032\linewidth]{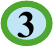}}, and {\includegraphics[width=0.03\linewidth]{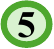}}, respectively.
  }
  \label{fig:comparison}
  \vspace{-5mm}
\end{figure}    

\subsection{Motivation of Our CARE}
\label{sec:motivation}

\subsubsection{The Calibration Stage}

Conventional softmax-based scoring is translation-invariant and prone to overconfident predictions on corrupted labels. Thus, ambiguous or fragmented pedestrian images can receive high scores for incorrect identities, causing mislabeled examples to be mistakenly treated as reliable supervision. As illustrated in Fig.~\ref{fig:comparison}, this issue stems from softmax assigning similarly high confidence to both clean simple samples and mislabeled ones despite their distinct feature patterns, making early-stage discrimination difficult and propagating errors forward.

To overcome this limitation, the \textbf{Calibration} stage introduces the PEC, which replaces deterministic softmax with evidence modeling based on the Dirichlet distribution. By incorporating adaptive, camera-aware smoothing into the similarity measurement, PEC breaks the translation invariance of softmax. Consequently, clean samples accumulate concentrated evidence toward their true class, while noisy samples exhibit diffuse and uncertain evidence profiles, thereby explicitly revealing feature–label mismatches.
These calibrated uncertainty estimates identify samples that are unlikely to belong to their annotated class. This provides a reliable signal for the subsequent \textbf{Refinement} stage to reweight or rectify supervision, effectively mitigating early error propagation.

\subsubsection{The Refinement Stage}

Even with calibrated uncertainty, conventional small-loss selection remains flawed: it risks discarding clean but hard-to-learn positive samples because their high loss resembles that of mislabeled samples, as shown in Fig.~\ref{fig:comparison}. This confusion arises because loss alone cannot separate learning difficulty from label corruption. The key issue is the lack of a stable, geometry-aware measure to distinguish these cases. Hard positives typically lie near the true class boundary with coherent alternative predictions, while mislabeled samples show both larger angular deviation and scattered top-$k$ predictions due to feature-label mismatch.

To address this, the \textbf{Refinement} stage employs a CAM metric within the hyperspherical feature space. CAM jointly evaluates: 1) angular separation from the assigned class and 2) ambiguity among top-$k$ alternatives, clearly separating hard positives (compact ambiguity) from noise (dispersed ambiguity). These epoch-smoothed CAM scores are converted by the COSW module into continuous sample weights, enabling soft, progressive re-weighting rather than hard filtering. The weights then guide a co-training loop via weighted objectives, ensuring the feature representation and uncertainty estimator refine each other.
The \textbf{Refinement} stage transforms uncertainty into robust sample weights, preserving hard positives while suppressing noise. This creates a synergistic cycle with the \textbf{Calibration} stage, where better weights yield cleaner features, and cleaner features yield better uncertainty estimates.

\subsection{Probabilistic Evidence Calibration}
In this subsection, we introduce the PEC in the \textbf{Calibration} stage (see the top in Fig.~\ref{fig:framework}), which injects adaptive learnable parameters into the similarity function to break the translation invariance commonly associated with softmax-based scores.
Formally, suppose that the training set is $\mathcal{D} = \left\{\left(x_{i}, y_{i}\right)\right\}_{i=1}^{N}$ with the corresponding class label $y_{i}$.
Concretely, let $\mathrm{z}_{ij}$
be the logit scores of sample $x_i$ belonging to class $j$. We define a calibrated score \(\tilde{p}(y_j|x_i)\) by introducing an adaptive learnable parameter \(s_j\), \textit{i.e.},
\begin{equation}
\tilde{p}(y_j|x_i) \;=\; \frac{\exp(\mathrm{z}_{ij}) + s_j}{\sum_{k=1}^{C} \bigl(\exp(\mathrm{z}_{ik}) + s_j\bigr)},
\quad j=1,\dots,C,
\label{eq:calib_prob}
\end{equation}
where \(s_j=\mathrm{softplus}(\Theta)\) is a nonnegative scalar associated with camera and \(\Theta\) is optimized jointly with network parameters. Injecting \(s_j\) into both numerator and denominator adaptively smooths the per-class similarity distribution and empirically reduces over-confidence on spurious matches that arise from camera-specific artifacts.
Even with such calibration, the conventional cross-entropy gradients may still be weakened in the presence of noisy labels.
The main reason lies in the fact that the gradient changes caused by the adaptive learnable smoothing term (\textit{i.e.}, $s_j$) are strictly positive for non-target classes, which can reduce the magnitude of the error gradient.
Intuitively, this operation mitigates the harmful influence of corrupted labels and encourages a more lenient probabilistic treatment of prediction results.

Importantly, the robust learning from noisy labels requires not only accurate predictions but also well-calibrated and interpretable uncertainty estimates.
The commonly used techniques for uncertainty estimation include entropy regularization~\cite{liu2021tractable, venkataramanan2023gaussian}, temperature scaling~\cite{ding2021local,joy2023sample}, Bayesian Neural Networks (BNNs)~\cite{franchi2024make, li2025federated}, deep ensembles~\cite{rahaman2021uncertainty, abe2022deep}, and Dirichlet distribution\cite{zong2024dirichlet, yu2024discretization}. The entropy regularization penalizes low-entropy (over-confident) distributions but operates as a post-hoc regularizer, lacking a structured probabilistic model of the underlying evidence. 
The temperature scaling adjusts confidence scores uniformly via a learned scalar, improving calibration metrics on a validation set but without modifying the model's internal learning dynamics. 
The BNNs and deep ensembles provide principled uncertainty by modeling parameter or predictive distributions, yet they often incur substantial computational overhead from multiple forward passes or maintaining multiple models.

In contrast, evidence theory based on the Dirichlet distribution provides a structured framework for uncertainty estimates. Its key advantage is the explicit modeling of evidence strength, where the concentration parameters represent per-class evidence counts, naturally distinguishing between uncertain and confident states. 
This evidence-based perspective aligns with the goal of discerning between reliable semantic commonality and spurious or variable semantic clues in person Re-ID.
Crucially, this evidence-based uncertainty is obtained through a single deterministic forward pass, providing robustness comparable to ensembles at a fraction of the computational cost. Empirically, the uncertainty quality achieved with the Dirichlet evidence model is comparable to that of ensemble methods, while being significantly more efficient. Therefore, considering effectiveness, efficiency, and implementation simplicity, we adopt the Dirichlet distribution to construct our proposed PEC in the \textbf{Calibration} stage.
Let \(\upsilon_i = \varphi(\mathrm{z}_i)\ge 0\) be a nonnegative evidence vector mapped from logits via a rectifying mapping \(\varphi(\cdot)\) (\textit{e.g.}, exponential function). 
Then, we define the Dirichlet distribution parameter as follows:
\begin{equation}
\mu_i \;=\; \upsilon_i + \kappa\mathbf{1},
\label{eq:alpha_def}
\end{equation}
where \(\kappa>0\) is a small concentration offset (typically $ \kappa= 1$) and \(\mathbf{1}\in\mathbb{R}^C\) is a vector containing $C$ ones. 
After that, the predicted probability for class $j$ is then given as:
\begin{equation}
\mathbb{E}_{\mathrm{z}_i\sim\mathrm{Dir}(\mu_i)}[\mathrm{z}_{ij}]
= \frac{\mu_{ij}}{\sum_{k=1}^C \mu_{ik}}.
\label{eq:dir_mean}
\end{equation}

Instead of minimizing point-wise cross-entropy on \(\tilde p_i\), we optimize the Dirichlet parameter \(\mu_i\) to achieve both accurate and well-calibrated predictive distributions.
Therefore, this leads to an expected negative log-likelihood (ENLL) loss under the Dirichlet posterior as the supervision signal:
\begin{equation}
\mathcal{L}_{\mathrm{ENLL}} \;=\; \frac{1}{N}\sum_{i=1}^N \sum_{j=1}^C \mathcal{I}_{ij}\Bigl[\,\psi\!\bigl(S_i\bigr) - \psi(\mu_{ij})\,\Bigr],
\label{eq:nll_evid}
\end{equation}
where \(\mathcal{I}_{ij}\in\{0,1\}\) is the one-hot label, \(S_i=\sum_{j}\mu_{ij}\) represents the concentration sum, and \(\psi(\cdot)\) denotes the Digamma function. 
Minimizing this expectation, \textit{i.e.}, equal to $\mathbb{E}_{\mathrm{z}_i\sim\mathrm{Dir}(\mu_i)}[-\log \mathrm{z}_{ij}]$ under the Dirichlet, yields smoother gradients than a hard point-wise cross-entropy estimation.

To regularize learned uncertainty and prevent degeneration under noisy labels, we introduce a Dirichlet divergence term that quantifies the discrepancy between the inferred Dirichlet distribution and a weakly informative uniform prior. This regularization employs the Kullback-Leibler (KL) Divergence between Dirichlet distributions, formulated as follows:

\begin{figure*}[t]
\centering
\includegraphics[width=0.9\linewidth, height=7cm]{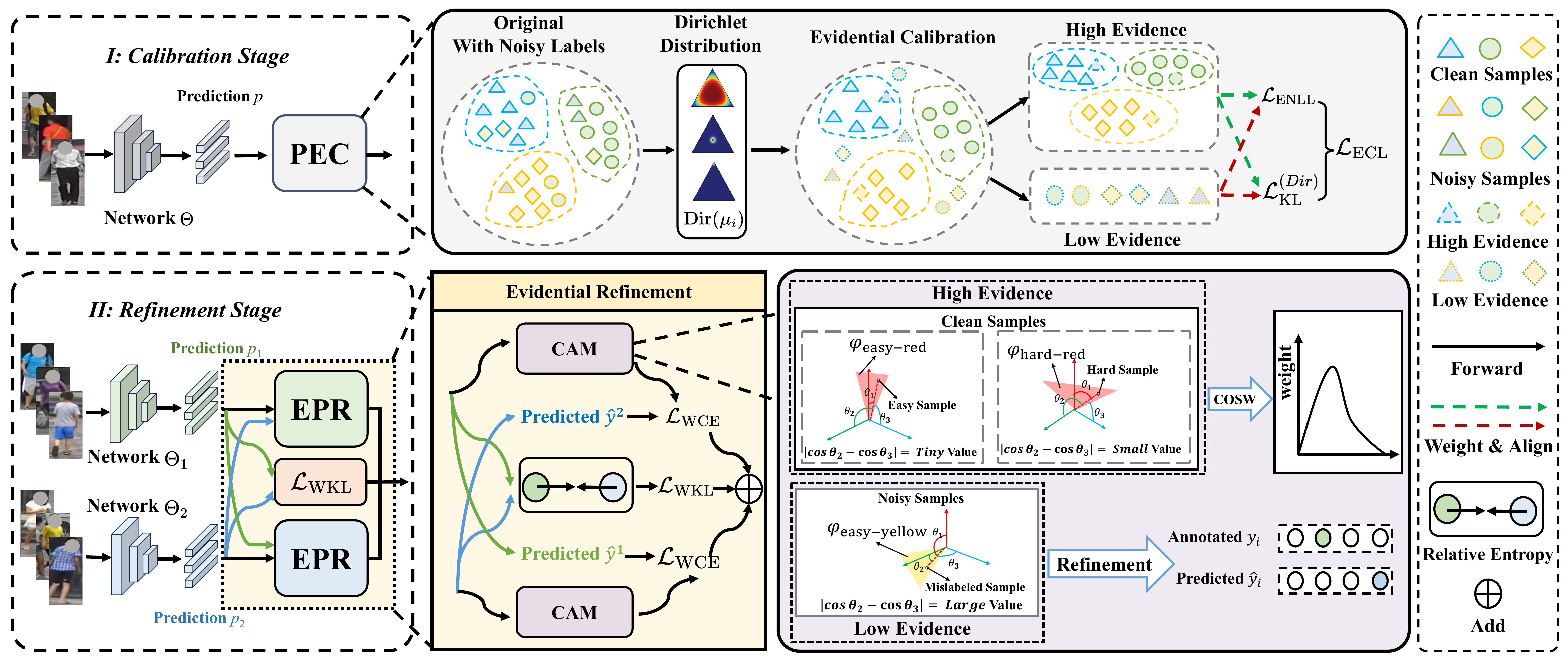}\\
\caption{Illustration of overall \textbf{CARE} framework. 
\textbf{Calibration} stage: PEC integrates Dirichlet‑informed prediction calibration to break softmax's translation invariance to mitigate over‑confidence. 
\textbf{Refinement} stage: EPR, powered by the CAM metric, surpasses small-loss methods in distinguishing clean but hard‑to‑learn samples from mislabeled ones; then COSW dynamically reallocates sample importance to prioritize clean instances over noisy instances.
}
\vspace{-4mm}
\label{fig:framework}
\end{figure*}
\begin{equation}
\resizebox{0.9\linewidth}{!}{$
\begin{split}
\mathcal{L}_{\mathrm{KL}}^{(Dir)} &\;=\; \frac{1}{NC}\sum_{i=1}^N D_{\mathrm{KL}}\bigl(\mathrm{Dir}(\mu_i)\,\|\,\mathrm{Dir}(\mathbf{1})\bigr) \\
&\;=\; \frac{1}{NC}\sum_{i=1}^N \Biggl[ \log\frac{\mathrm{B}(\mathbf{1})}{\mathrm{B}(\mu_i)}
+ \sum_{k=1}^C (\mu_{ik}-1)\bigl(\psi(\mu_{ic})-\psi(S_i)\bigr)\Biggr],
\end{split}
$}   
\label{eq:kl_dir}
\end{equation}
where \(\mathrm{B}(\cdot)\) denotes the multivariate Beta function. This regularization term penalizes overconfident Dirichlet assignments on noisy samples and promotes appropriately uncertain predicted distributions when evidence is insufficient.

The final evidential calibration loss (ECL) for training in the \textbf{Calibration} stage is expressed as follows:
\begin{equation}
\mathcal{L}_{\mathrm{ECL}} \;=\; \mathcal{L}_{\mathrm{ENLL}} + \lambda\,\mathcal{L}_{\mathrm{KL}}^{(Dir)},
\label{eq:ecl_final}
\end{equation}
where $\lambda > 0$ balances fidelity to noisy labels against calibrated uncertainty requirements.
In practice, we integrate Eq.~\eqref{eq:calib_prob}'s camera-aware smoothing into the similarity computation feeding $\varphi(\cdot)$, jointly optimizing network weights $\Theta$ and optional $\kappa$ via standard stochastic gradient descent. Empirically, this evidential calibration (\textit{i.e.}, as shown in Fig.~\ref{fig:motivation}\textcolor{red}{(e)}) mitigates early-stage error accumulation and generates more reliable initial pseudo-labels for the latter \textbf{Refinement} stage (see the bottom in Fig.~\ref{fig:framework}).
By providing well-calibrated, evidence-based uncertainty estimates, the \textbf{Calibration} stage establishes a reliable foundation for the subsequent \textbf{Refinement} stage, where sample-level assessment and re-weighting can be performed with greater accuracy and fairness.

\subsection{Evidence Propagation Refinement}
\label{sec:Evidence Propagation Refinement}
\subsubsection{Composite Angular Margin}
\label{sec:Composite Angular Margin}

Robust person Re-ID requires mitigating the influence of corrupted label annotations during training. Conventional approaches that rely on large-loss heuristics risk mistakenly discarding legitimate hard positives (\textit{i.e.}, clean but hard-to-learn samples), such as heavily occluded or extreme viewpoint samples, which can provide useful information. To overcome this limitation, we propose a composite angular margin (CAM) metric integrating two complementary robustness indicators: 1) an inter-class angular margin measuring the separation ($\Delta$) between the true identity and its noisy label, and 2) a top-$k$ ambiguity measure quantifying uncertainty ($\Lambda$) among most probable alternative classes.
$\Delta$ measures separation from the assigned class, while $\Lambda$ quantifies uncertainty among competing alternatives. Crucially, clean but hard-to-learn positive samples tend to show small $\Delta$ and small $\Lambda$ (confined ambiguity), whereas mislabeled samples typically exhibit small $\Delta$ but large $\Lambda$ (dispersed predictions). This complementary synergy enables CAM to more reliably discriminate true hard positives from label noise.

At epoch \(t\), let \(\mathrm{z}_i^{(t)}\) denote the logit vector for sample \(x_i\), with \(y_i\) as its assigned identity. The instantaneous CAM score is defined as a linear combination:
\begin{equation}
\mathcal{S}^{(t)}(x_i) \;=\; \alpha\cdot \Delta^{(t)}(x_i) \;-\; \beta\cdot \Lambda^{(t)}(x_i),
\label{eq:cam_inst}
\end{equation}
where \(\alpha,\beta>0\) are scaling coefficients. The first term \(\Delta^{(t)}\) (angular separation) quantifies the margin by which the target logit exceeds the strongest non-target logit:
\begin{equation}
\Delta^{(t)}(x_i) \;=\; \mathrm{z}^{(t)}_{y_i}(x_i; \Theta) \;-\; \max_{j\neq y_i} \mathrm{z}^{(t)}_{j}(x_i; \Theta).
\label{eq:ang_sep}
\end{equation}

Intuitively, the operative role of this item can maintain positive angular separation between clean samples (both easy-to-learn and hard-to-learn examples) and mislabeled samples.

The second component \(\Lambda^{(t)}\) (top-\(k\) spread) quantifies the divergence between the dominant competitor and the average of the next $k-1$ candidates:
\begin{equation}
\resizebox{0.88\linewidth}{!}{$
\Lambda^{(t)}(x_i) \;=\; \max_{j \neq y_i} \mathrm{z}^{(t)}_{j}(x_i; \Theta)
\;-\; \frac{1}{k}\sum_{l\in\mathcal{T}_{k}(x_i)} \mathrm{z}^{(t)}_{l}(x_i; \Theta),
$}
\label{eq:topk_spread}
\end{equation}
where \(\mathcal{T}_{k}(x_i)\) indexes the top-\textit{k} logits (including the dominant competitor \(l^\star\)). 
For correctly labeled hard samples, the remaining top-\(k\) logits typically concentrate around the leading competitor, reducing \(\Lambda\); conversely, mislabeled samples exhibit dispersion in this set, increasing \(\Lambda\).

To derive robust sample-wise reliability scores, we aggregate epoch-level CAM metrics via a temporal decay weighting. Define \(w_j \geq 0\) as the epoch-\textit{j} weight (\textit{e.g.}, from a cosine decay schedule). The accumulated CAM score at epoch $t$ is:
\begin{equation}
\mathrm{CAM}_t(x_i) = \sum_{j=1}^{t} w_j \mathcal{S}^{(j)}(x_i),
\label{eq:cam_accum}
\end{equation}
\begin{equation}
    w_j = \frac{\exp\bigl(1+\cos(\frac{\pi j}{T})\bigr)} {\sum_{r=1}^t \exp\bigl(1+\cos(\frac{\pi r}{T})\bigr)},
    \label{eq:w_j}
\end{equation}
where $T$ denotes the total epochs.
This normalized aggregation stabilizes noisy instantaneous signals, producing a robust separator between challenging clean samples (high CAM) and likely mislabeled instances (low CAM).

\subsubsection{Certainty-Oriented Sphere Weighting}
\label{sec:Certainty-Oriented Sphere Weighting}
Given the reliable CAM metrics, we compute hyperspherical sample weights with COSW that prioritize trustworthy instances while preserving hard positives.
The key role of COSW is to normalize the CAM score into a bounded weight $\mathcal{S}_{\mathrm{COSW}}^{(t)}(x_i) \in [0,1]$, thereby enabling soft, adaptive sample weighting instead of hard threshold or binary selection. This is crucial for Re-ID, given the high informational value of each training instance.

Let \(\phi(x_i; \Theta)\in\mathbb{R}^d\) represent the extracted feature of the sample $x_i$ and \(w_q\in\mathbb{R}^d\) the class prototype (classifier weight) for identity $q$. The normalized angular distance to identity $q$ at epoch $t$ is computed as:
\begin{equation}
d^{(t)}_{q}(x_i;\Theta) = \frac{1}{2} \cdot ( 1- \frac{\langle \phi^{(t)}(x_i; \Theta), w_q\rangle}
{\|\phi^{(t)}(x_i; \Theta)\|_2 \cdot \|w_q\|_2}),
\label{eq:ang_dist}
\end{equation}
where \( \langle \cdot, \cdot \rangle \) denotes the inner product operation and \( \| \cdot \|_2 \) is the $\ell_2$ norm.
This distance normalization ensures scale invariance and provides an interpretable distance metric directly mapped from cosine similarity.
Then, we transform angular separation and top-$k$ spread into the hyperspherical domain by substituting logits with normalized distances $d^{(t)}_{q}(x_i;\Theta)$, deriving hyperspherical analogues $\widetilde{\Delta}^{(t)}(x_i)$ and $\widetilde{\Lambda}^{(t)}(x_i)$ in Eqs.~\eqref{eq:ang_sep} and \eqref{eq:topk_spread}.
Since higher prediction probabilities correspond to smaller angular distances, we modify the optimization direction from maximization to minimization for consistency.

In order to obtain a bounded certainty score for each sample, we put the transformed components through a smooth compression function and combine them:
\begin{equation}
\resizebox{0.88\linewidth}{!}{$
\mathcal{S}_{\mathrm{COSW}}^{(t)}(x_i) \;=\; \frac{1}{2} \cdot \Bigl( \frac{1}{1 + \exp{(\alpha \cdot \widetilde{\Delta}^{(t)})}} + \exp{(\beta \cdot \widetilde{\Lambda}^{(t)})}\Bigr),
$}
\label{eq:cosw_inst}
\end{equation}
where \(\mathcal{S}_{\mathrm{COSW}}^{(t)}(x_i)\in [0,1]\) represents epoch-wise certainty score. It signifies: values near 1 denote reliable and clean positive samples; values near 0 suggest potential mislabeling or low-confidence samples.
Then, we integrate epoch-wise certainties with the same cosine decay
schedule (\textit{i.e.}, Eqs.~(\ref{eq:cam_accum}) and (\ref{eq:w_j})) for CAM to derive the stable hyperspherical CAM. 

\subsubsection{Training Objective for Refinement Stage}
In the \textbf{Refinement} stage, we initialize network $\Theta_1$ and network $\Theta_2$ using the trained network weights from the \textbf{Calibration} stage. To maintain the robustness of these two networks against noisy labels, a co-training scheme is employed in the \textbf{Refinement} stage. Specifically, we use Weighted Cross-Entropy and Weighted KL Divergence to align the two peer networks.
For the Weighted Cross-Entropy objective, the predicted labels of the networks are used as supervisory information to constrain the network training. For network $\Theta_1$ and network $\Theta_2$, the corresponding Weighted Cross-Entropy objective is defined as:
\begin{equation}
\resizebox{0.88\linewidth}{!}{$
\mathcal{L}_{\mathrm{WCE}} = -\frac{1}{N} \sum_{i=1}^{N}  \sum_{j=1}^{C} \Bigl[ \mathcal{R}_1(j, \tilde{y}^{2}_{i}) \log p_1(j \mid x_i)\bigr) + \mathcal{R}_2(j, \tilde{y}^{1}_{i}) \log p_2(j \mid x_i)\bigr) \Bigr],
$}
\end{equation}

\begin{equation}
\resizebox{0.65\linewidth}{!}{$
\mathcal{R}(j, \tilde{y}^{k}_{i}) =
\begin{cases}
\text{CAM}(x_i) & \text{if $x_i$ is clean} \\
p(\tilde{y}^{k}_{i} \mid x_i) & \text{if $x_i$ is noisy}
\end{cases},
$}
\label{eq:R}
\end{equation}
where $\mathcal{R}(j, \tilde{y}^{k}_{i})$ is a weighted function and $k$ is 1 or 2. In other words, the predicted labels of network $\Theta_1$ are used to supervise network $\Theta_2$, while the predicted labels of network $\Theta_2$ are utilized to supervise network $\Theta_1$.

Meanwhile, to further facilitate knowledge sharing and transfer between the two networks, we use the Weighted KL Divergence to constrain the two peer networks as follows:
\begin{equation}
\resizebox{0.88\linewidth}{!}{$
\begin{aligned}     
\mathcal{L}_{\mathrm{WKL}} \;=\; & \frac{1}{N}\sum_{i=1}^N \sum_{j=1}^C \Bigl[ D_{\mathrm{WKL}}\bigl(p_2(j \mid x_i) \,\|\, p_1(j \mid x_i)\bigr)
+ \\
& D_{\mathrm{WKL}}\bigl(p_1(j \mid x_i) \,\|\, p_2(j \mid x_i)\bigr)\Bigr] 
\end{aligned}
$}
\label{eq:kl_net}
\end{equation}
where $D_{\mathrm{WKL}}$ constrains the relationship between two networks, formalized as follows: 
\begin{equation}
\resizebox{0.88\linewidth}{!}{$
\begin{aligned} 
D_{\mathrm{WKL}} \bigl(p_2(j \mid x_i) \,\|\, p_1(j \mid x_i)\bigr) = & -\mathcal{R}\left(j, \tilde{y}_i^2\right) \log p_1\left(j \mid x_i\right)+ \\
& p_2\left(j \mid x_i\right) \log p_1\left(j \mid x_i\right),
\end{aligned}
$}
\end{equation}
\begin{equation}
\resizebox{0.88\linewidth}{!}{$
\begin{aligned} 
D_{\mathrm{WKL}}\bigl(p_1(j \mid x_i) \,\|\, p_2(j \mid x_i) = & -\mathcal{R}\left(j, \tilde{y}_i^1\right) \log p_2\left(j \mid x_i\right)+ \\
& p_1\left(j \mid x_i\right) \log p_2\left(j \mid x_i\right).
\end{aligned}
$}
\end{equation}

In this manner, this interactive guidance mechanism combines collaborative learning for each network by optimizing itself together with the predicted labels from its peer network, while matching its probability distribution to its peer network. Ultimately, the networks and the predicted labels are gradually improved simultaneously.

\section{Experiments}
\label{sec:Experiments}

\begin{table*}[t]  
\tiny
\begin{center}
\resizebox{\linewidth}{!}{
\begin{tabular}{c|l|c|c|cccc|cccc|cccc}
\cline{1-16}
\multicolumn{4}{c|}{Datasets}&\multicolumn{4}{c|}{Market1501}&\multicolumn{4}{c|}{DukeMTMC-ReID}&\multicolumn{4}{c}{CUHK03}\\\hline

Noise&Methods&Venue&Fields&Rank-1&Rank-5&Rank-10&mAP&Rank-1&Rank-5&Rank-10&mAP&Rank-1&Rank-5&Rank-10&mAP\\\hline

\multirow{16}*{{10\%}}
&SL*\cite{wang2019symmetric}&ICCV19&IC&79.2&91.8&94.8&57.0&68.6&81.7&86.7&50.2&33.9&54.7&65.2&33.5\\
  
& MMT* \cite{ge2020mutual}&ICLR20&IC&81.1&93.5&95.8&63.2&73.0&86.1&90.3&59.6&27.2&49.2&60.7&29.7\\
& JoCoR*\cite{wei2020combating} &CVPR20&IC&81.5&93.4&96.0&60.8&68.2&81.2&86.3&48.3&28.8&49.2&59.6&29.4\\  
&Focal*\cite{lin2020focal}&TPAMI20&IC&78.7&91.5&94.8&57.3&68.3&82.1&86.5&50.4&-&-&-&-\\ 

&UbiW*\cite{li2021learning}&TPAMI21&IC& 79.1 & 91.9 & 95.0 & 56.8 & 68.9 & 81.8 & 86.7 & 49.6&30.1&50.4&60.9&29.5\\

& ACT\cite{sheng2024enhancing} &ACMMM24&IC&88.4&96.7&98.1&64.6&74.3&86.3&89.6&54.6&31.7&50.6&63.6&30.6\\   
& SED\cite{sheng2024foster} & ECCV24 & IC & 88.5 & 96.7 & 98.2 & 64.8 & 74.6 & 86.7 & 89.7 & 55.0 & 31.9 & 50.7 & 64.0 & 30.9\\ 
& CA2C\cite{sheng2025ca2c} & ICCV25 & IC & 88.7 & 96.8 & 98.2 & 65.1 & 74.9 & 87.1 & 89.7 & 55.4 & 32.2 & 51.2 & 64.2 & 31.2\\  
& PLReMix\cite{liu2025plremix} & WACV25 & IC & 88.8 & 96.8 & 98.1 & 65.6 & 75.0 & 87.2 & 89.8 & 56.2 & 32.1 & 51.4 & 64.8 & 33.0\\  
& DULC\cite{xu2025revisiting} & AAAI25 & IC & 89.1 & 96.9 & 98.3 & 66.3 & 75.1 & 87.3 & 89.7 & 57.3 & 32.5 & 51.6 & 64.6 & 33.2\\  

&DistributionNet*\cite{yu2019robust}&ICCV19&PR&82.3&93.1&95.8&61.5&68.6&81.9&86.1&48.0&32.3&51.8&61.7&31.8\\
&PurifyNet*\cite{ye2020purifynet}&TIFS20&PR& 84.2 & 93.7 & 95.6 & 64.3 & 74.6 & 85.6 & 89.6 & 56.8&32.8&54.8&65.0&32.8\\    

&CORE*\cite{ye2022collaborative}&TIP22&PR& 85.5 & 94.0 & 96.4 & 67.7 & 75.4 & 86.7 & 90.8 & 60.1&40.4&60.8&70.9&39.6\\
&LRNI-HSR*\cite{zhong2023neighborhood}&ICASSP23&PR&86.9 & 94.7 & 96.6 & 68.8 & 76.1 & 87.3 & 90.4 & 57.7&-&-&-&-\\
&ICLR*\cite{zhong2024iclr}&PR24&PR& 88.0 & 95.2 & 97.0 & 70.8 & \textbf{77.6} & \textbf{88.1} & 90.8 & 59.4&-&-&-&-\\
\cline{2-16}
& \textbf{CARE} (Ours) & This work &PR& \textbf{90.8} & \textbf{97.3} & \textbf{98.4} & \textbf{71.0} & 76.5 & 87.5 & \textbf{90.8} & \textbf{60.1} & \textbf{41.1} & \textbf{61.4} & \textbf{72.0} & \textbf{40.6}\\\hline

\multirow{16}*{{20\%}}
&Co-teaching*\cite{han2018co}&NIPS18&IC&72.7&88.3&92.6&52.3&56.3&71.0&77.2&37.9&19.0&31.0&38.9&18.8\\
&CleanNet*\cite{lee2018cleannet}&CVPR18&IC&71.4&87.3&91.9&44.2&55.1&71.7&76.5&34.0&12.1&25.9&34.0&12.9\\  
&SL*\cite{wang2019symmetric}&ICCV19&IC&76.8&90.9&93.4&54.0&67.6&81.9&86.2&47.8&29.4&49.2&58.2&29.2\\
& UbiW* \cite{li2021learning}&TPAMI21&IC&76.0&90.0&93.4&52.9&64.8&78.4&83.8&45.4&23.1&41.6&52.1&21.9\\   
& ACT\cite{sheng2024enhancing} &ACMMM24&IC&87.2&95.8&97.4&60.6&69.9&82.8&87.2&50.0&25.0&42.0&52.6&23.7\\      
& SED\cite{sheng2024foster} & ECCV24 & IC & 87.4 & 96.0 & 97.5 & 61.6 & 70.4 & 83.0 & 87.9 & 52.3 & 25.1 & 42.4 & 53.0 & 24.1\\  
& CA2C\cite{sheng2025ca2c} & ICCV25 & IC & 88.1 & 96.1 & 97.8 & 63.1 & 70.3 & 83.2 & 88.1 & 53.1 & 25.7 & 43.3 & 53.2 & 25.8\\  
& PLReMix\cite{liu2025plremix} & WACV25 & IC & 88.2 & 96.2 & 97.9 & 64.3 & 71.1 & 84.1 & 88.2 & 53.3 & 26.2 & 44.1 & 55.2 & 27.1\\  
& DULC\cite{xu2025revisiting} & AAAI25 & IC & 88.4 & 96.4 & 97.8 & 65.5 & 72.0 & 84.8 & 88.7 & 54.9 & 27.4 & 45.2 & 56.1 & 28.6\\

&DistributionNet*\cite{yu2019robust}&ICCV19&PR&77.0&90.6&94.0&53.4&62.4&77.4&82.5&40.9&24.3&41.6&53.1&24.2\\    
&PurifyNet*\cite{ye2020purifynet}&TIFS20&PR&83.1&93.3&95.9&63.1&74.1&85.6&89.2&55.8&30.2&49.8&58.4&29.2\\
&CORE*\cite{ye2022collaborative} &TIP22&PR&84.1&93.1&95.5&66.2&74.4&85.9&89.7&55.8&34.4&55.0&64.1&35.0\\
&LRNI-HSR*\cite{zhong2023neighborhood} &ICASSP23&PR&84.5&93.6&96.0&65.6&76.0&87.8&91.0&56.9&-&-&-&-\\
&LRP*\cite{chen2023refining}&TMM23&PR&88.8&95.1&97.0&70.5&\textbf{77.7}&\textbf{88.0}&91.2&\textbf{60.3}&-&-&-&-\\
&TSNT*~\cite{liu2024two}&TPAMI24&PR& 85.7 & 94.1 & 96.1 & 68.1 & 75.2 & 87.0 & 90.6 & 57.4 &-&-&-&-\\
&ICLR*~\cite{zhong2024iclr}&PR24&PR&87.2&94.9&96.9&70.3&77.6&87.8&90.0&59.2&-&-&-&-\\
\cline{2-16}
&\textbf{CARE} (Ours) & This work &PR&\textbf{90.4}&\textbf{97.1}&\textbf{98.1}&\textbf{70.5}&77.1&87.8&\textbf{91.2}&59.7&\textbf{37.7}&\textbf{58.8}&\textbf{67.2}&\textbf{37.2}\\\hline

\multirow{16}*{{30\%}}
&SL*\cite{wang2019symmetric}&ICCV19&IC& 73.9 & 88.9 & 92.7 & 51.5 & 63.7 & 79.6 & 84.2 & 43.7&23.6&40.2&52.4&22.1\\
 &MMT*\cite{ge2020mutual}&ICLR20&IC& 64.9 & 83.4 & 89.0 & 43.5 & 58.9 & 75.9 & 80.3 & 44.4&13.4&27.4&38.1&15.4\\
& JoCoR* \cite{wei2020combating}&CVPR20&IC& 74.6 & 89.2 & 92.9 & 52.8 & 63.2 & 79.1 & 84.3 & 44.6&22.9&40.9&52.0&22.8\\
&Focal*\cite{lin2020focal}&TPAMI20&IC& 71.0 & 87.4 & 91.7 & 48.5 & 62.6 & 78.2 & 82.9 & 44.0&-&-&-&-\\
&UbiW*\cite{li2021learning}&TPAMI21&IC& 70.9 & 87.1 & 91.5 & 49.0 & 61.2 & 77.2 & 82.4 & 41.8&20.2&36.0&46.0&20.1\\ 
& ACT\cite{sheng2024enhancing} &ACMMM24&IC&85.1&95.5&97.1&57.2&66.1&81.2&85.6&45.1&16.3&32.2&42.0&16.2\\ 
& SED\cite{sheng2024foster} & ECCV24 & IC & 85.7 & 95.8 & 97.4 & 58.8 & 66.4 & 81.7 & 86.1 & 46.1 & 16.7 & 33.1 & 43.0 & 17.0\\  
& CA2C\cite{sheng2025ca2c} & ICCV25 & IC & 86.2 & 95.9 & 97.5 & 59.3 & 68.3 & 82.5 & 87.1 & 47.3 & 19.2 & 33.9 & 43.2 & 18.1\\  
& PLReMix\cite{liu2025plremix} & WACV25 & IC & 86.6 & 95.8 & 97.5 & 60.9 & 69.2 & 83.3 & 87.7 & 49.0 & 20.1 & 35.1 & 44.5 & 20.2\\  
& DULC\cite{xu2025revisiting} & AAAI25 & IC & 86.4 & 95.7 & 97.4 & 60.7 & 69.4 & 83.5 & 87.8 & 49.6 & 21.2 & 35.2 & 44.7 & 21.3\\

& PurifyNet*\cite{ye2020purifynet} &TIFS20&PR & 81.4 & 92.3 & 94.9 & 60.2 & 71.5 & 84.1 & 87.8 & 52.8&26.6&44.6&54.2&26.4\\      
&CORE*\cite{ye2022collaborative}&TIP22&PR & 81.2 & 92.2 & 94.5 & 60.6 & 72.5 & 85.6 & 88.9 & 55.5&27.7&50.3&60.1&29.3\\
&LRNI-HSR*\cite{zhong2023neighborhood}&ICASSP23&PR& 81.9 & 92.5 & 95.2 & 62.2 & 72.8 & 85.2 & 89.1 & 53.9&-&-&-&-\\
&LRP*\cite{chen2023refining}&TMM23&PR& 86.3 & 94.5 & 96.4 & 67.0 & 75.9 & 86.6 & 90.2 & 57.9&-&-&-&-\\
&ICLR*\cite{zhong2024iclr}&PR24&PR& 86.4 & 94.5 & 96.6 & \textbf{69.7} & \textbf{77.1} & 87.2 & 90.4 & 58.1&-&-&-&-\\
\cline{2-16}
&\textbf{CARE} (Ours)& This work &PR&\textbf{89.4}&\textbf{96.6}&\textbf{97.7}&67.4&74.9&\textbf{87.5}&\textbf{90.4}&\textbf{58.8}&\textbf{33.8}&\textbf{54.3}&\textbf{64.0}&\textbf{33.6}\\\hline

\multirow{12}*{{50\%}}
&Co-teaching*\cite{han2018co}&NIPS18&IC&59.4&77.3&84.0&35.3&44.3&62.6&69.2&24.8&7.4&17.0&23.6&7.5\\
&CleanNet* \cite{lee2018cleannet}&CVPR18&IC&52.7&72.9&80.0&26.1&39.0&55.5&62.2&19.0&4.9&11.5&16.1&5.2\\  
& UbiW* \cite{li2021learning}&TPAMI21&IC&58.7&79.6&86.7&33.1&44.8&63.4&68.5&24.6&9.4&21.1&29.7&10.8\\   
& ACT\cite{sheng2024enhancing} &ACMMM24&IC&81.8&92.6&95.4&47.7&58.6&73.9&79.8&37.7&8.4&21.2&29.5&10.1\\   
& SED\cite{sheng2024foster} & ECCV24 & IC & 81.9 & 92.8 & 95.8 & 48.0 & 58.9 & 74.2 & 79.7 & 37.9 & 9.2 & 22.6 & 30.5 & 12.1\\ 
& CA2C\cite{sheng2025ca2c} & ICCV25 & IC & 82.3 & 93.8 & 96.0 & 48.9 & 59.1 & 75.1 & 80.1 & 38.5 & 10.7 & 23.1 & 30.6 & 13.0\\ 
& PLReMix\cite{liu2025plremix} & WACV25 & IC & 82.5 & 93.7 & 95.8 & 50.3 & 59.2 & 75.2 & 80.6 & 38.8 & 10.8 & 23.8 & 31.6 & 13.9\\  
& DULC\cite{xu2025revisiting} & AAAI25 & IC & 82.8 & 94.0 & 95.9 & 51.2 & 59.7 & 75.7 & 81.4 & 39.8 & 11.1 & 24.5 & 32.6 & 14.6\\

&DistributionNet*\cite{yu2019robust}&ICCV19&PR&61.1&81.1&87.1&35.1&46.0&64.0&70.9&25.8&10.1&21.8&29.9&10.6\\    
&PurifyNet*\cite{ye2020purifynet}&TIFS20&PR&63.6&81.6&87.5&40.1&56.4&73.2&78.9&38.0&10.9&21.5&31.9&11.4\\
&CORE*\cite{ye2022collaborative} &TIP22&PR&66.3&83.9&89.4&42.2&59.2&73.4&78.9&39.2&14.4&28.4&37.0&14.8\\
\cline{2-16}
&\textbf{CARE} (Ours)& This work &PR&\textbf{83.7}&\textbf{94.6}&\textbf{96.1}&\textbf{54.6}&\textbf{62.4}&\textbf{77.2}&\textbf{82.0}&\textbf{43.4}&\textbf{20.7}&\textbf{38.1}&\textbf{47.4}&\textbf{21.2}\\\hline

\end{tabular}}
\end{center}
\caption{
Performance results with state-of-the-art methods under different random noise ratios (10\%, 20\%, 30\%, and 50\%). 
``IC'' denotes the methods designed for the image classification task, whereas ``PR'' refers to those methods developed for the person Re-ID task. ``*'' indicates the results are reported from their original papers, while the remaining results are reproduced under the same experimental settings. 
Best results are in \textbf{bold}.}
\label{tab:results_random}
\vspace{-4mm}
\end{table*}

\subsection{ Datasets and Experimental Settings}

\subsubsection{Datasets}
The proposed method is evaluated on three person Re-ID datasets. Market1501 dataset \cite{zheng2015scalable} comprises 32,668 images of 1,501 identities captured by 6 disjoint cameras, split into a training set with 12,936 images of 751 identities and a testing set containing 19,732 gallery images and 3,368 query images of 750 identities. DukeMTMC-ReID \cite{zheng2017unlabeled} includes 1,812 identities from 8 cameras, with 16,522 training images of 702 identities and a testing set of 2,228 query images and 17,661 gallery images. DukeMTMC-ReID is used only for evaluation and is not public. CUHK03~\cite{li2014deepreid} provides 14,097 images of 1,467 identities from 5 camera pairs, utilizing manually labeled and DPM-detected bounding boxes. We use the detected person images for evaluation.

\subsubsection{Evaluation Metrics}
For fair comparison, performance is measured by cumulative matching characteristics (CMC) at Rank-1, Rank-5, and Rank-10, alongside mean average precision (mAP). Rank-$k$ measures the probability that the top-$k$ retrieval results contain the correct match. mAP represents the average retrieval performance of multiple correct matches.

\subsubsection{Noisy Label Generation}
Following~\cite{yu2019robust}, 
we simulate noisy labels by assigning incorrect identities with a predefined percentage. 
The noise ratio varies across experiments to evaluate robustness under diverse contamination levels. We conduct the experiments under two different noisy label settings: Random Noise (10\%, 20\%, 30\%, and 50\%) and Patterned Noise (10\% and 20\%). 
Random Noise is to randomly select training images and then assign incorrect labels. Patterned Noise first trains the model using clean labels, then assigns randomly selected samples to the label of their most similar samples.

\subsubsection{Baseline}
For fair comparison, we adopt the IDE model, a standard baseline in robust Re‑ID research, and its online co‑refinement variant CORE, which iteratively refines noisy labels during training. While CORE enhances IDE with label co‑refining, our \textbf{CARE} directly builds upon CORE by integrating PEC and EPR through probabilistic evidence propagation.

\subsubsection{Implementation Details}
We employ ResNet50 on the PyTorch framework, which is a standard backbone in current person Re-ID systems.
The network is initialized with ImageNet pretrained weights, augmented by a 512-dimensional FC layer after pool5 with batch normalization and ReLU activation following common practice. Data augmentation includes random cropping (256 $\times$ 128 from resized 288 $\times$ 144 inputs) and horizontal flipping. Training uses a fixed batch size of 32 across all datasets. 
The initial learning rate is set to 0.01 during the first 20 epochs to establish stable feature representations in the \textbf{Calibration} stage, followed by a decay factor of 0.1 applied during the next 40 epochs to facilitate refined metric calibration in the \textbf{Refinement} stage. 
This phased approach progressively adapts to label noise while preserving discriminative capability. 
We set $\lambda = 0.5$ in Eq.~\ref{eq:ecl_final} and $\alpha=\beta =100$ in Eq.~\ref{eq:cam_inst}, and they are discussed in the ablation study.
All other hyperparameters align with baseline configurations for fair comparisons. Additionally, all experimental results reported in this paper were obtained on an NVIDIA RTX 2080Ti GPU.

\begin{table*}[t]  
\tiny
\begin{center}
\resizebox{0.96\linewidth}{!}{
\begin{tabular}{c|l|c|c|cccc|cccc|cccc}
\cline{1-16}
\multicolumn{4}{c|}{Datasets}&\multicolumn{4}{c|}{Market1501}&\multicolumn{4}{c|}{DukeMTMC-ReID}&\multicolumn{4}{c}{CUHK03}\\\hline

Noise & Methods & Venue & Fields & Rank-1 & Rank-5 & Rank-10 & mAP & Rank-1 & Rank-5 & Rank-10 & mAP & Rank-1 & Rank-5 & Rank-10 & mAP\\\hline

\multirow{13}*{{10\%}}

&SL*\cite{wang2019symmetric}&ICCV19&IC& 80.2 & 92.0 & 94.8 & 59.1 & 70.4 & 82.9 & 87.3 & 50.6&32.6&55.9&64.9&31.5\\   
& JoCoR* \cite{wei2020combating}&CVPR20&IC& 80.9 & 92.2 & 95.0 & 60.2 & 71.4 & 83.6 & 87.8 & 52.5&31.4&55.2&64.2&30.8\\
& ACT\cite{sheng2024enhancing} &ACMMM24&IC&88.2&96.2&97.6&64.5&72.7&85.6&89.4&53.4&32.9&52.3&61.4&30.4\\ 

& SED\cite{sheng2024foster} & ECCV24 & IC & 88.0 & 96.1 & 97.5 & 63.8 & 72.8 & 85.6 & 89.3 & 53.2 & 32.7 & 52.1 & 61.1 & 30.1\\ 
& CA2C\cite{sheng2025ca2c} & ICCV25 & IC & 88.3 & 96.2 & 97.6 & 64.9 & 73.1 & 85.8 & 89.6 & 55.3 & 33.0 & 52.5 & 61.9 & 31.0\\
& PLReMix\cite{liu2025plremix} & WACV25 & IC & 88.2 & 96.2 & 97.5 & 64.7 & 73.2 & 85.8 & 89.7 & 55.8 & 33.2 & 52.6 & 61.8 & 31.4\\  
& DULC\cite{xu2025revisiting} & AAAI25 & IC & 88.5 & 96.3 & 97.8 & 64.8 & 74.8 & 86.2 & 90.1 & 56.3 & 33.6 & 53.0 & 62.4 & 31.9\\

&DistributionNet*\cite{yu2019robust}&ICCV19&PR & 52.4 & 71.2 & 77.7 & 27.0 & 37.7 & 53.2 & 60.0 & 20.7&10.5&20.1&26.8&10.9\\  
& PurifyNet*\cite{ye2020purifynet} &TIFS20&PR & 81.8 & 92.6 & 95.4 & 63.2 & 73.5 & 84.8 & 90.1 & 56.1&33.6&56.4&66&32.9\\      
&CORE*\cite{ye2022collaborative}&TIP22&PR  & 84.2 & 93.9 & 96.3 & 66.1 & 76.5 & 87.3 & 90.9 & 59.2 & 35.0 & 55.1 & 64.9 & 33.6 \\
&LRNI-HSR*\cite{zhong2023neighborhood}&ICASSP23&PR& 85.3 & 93.9 & 96.3 & 67.7 & 75.9 & 86.9 & 90.4 & 56.3&-&-&-&-\\
&ICLR*\cite{zhong2024iclr}&PR24&PR & 86.8 & 94.5 & 96.5 & \textbf{68.0} & 78.4 & 88.4 & 91.2 & 59.8&-&-&-&-\\
\cline{2-16}
&\textbf{CARE} (Ours)& This work &PR&\textbf{89.3}&\textbf{96.4}&\textbf{98.0}&66.9&\textbf{79.2}&\textbf{89.5}&\textbf{92.4}&\textbf{60.5}&\textbf{35.3}&\textbf{55.6}&\textbf{65.1}&\textbf{33.9}\\\hline

\multirow{13}*{{20\%}}

&SL*\cite{wang2019symmetric}&ICCV19&IC & 76.5 & 90.3 & 93.6 & 54.5 & 65.9 & 82.5 & 87.0 & 45.6&29.8&50.2&60.9&28.7\\   
& JoCoR* \cite{wei2020combating}&CVPR20&IC & 75.2 & 89.0 & 92.6 & 53.8 & 67.4 & 81.6 & 87.1 & 46.9&24.8&42.9&53.9&25.2\\
& ACT~\cite{sheng2024enhancing} &ACMMM24&IC&85.2&95.5&96.6&58.5&61.9&78.4&83.6&40.7&25.9&43.6&53.7&23.8\\ & SED\cite{sheng2024foster} & ECCV24 & IC & 84.6 & 94.4 & 96.5 & 57.3 & 61.7 & 78.3 & 83.5 & 40.1 & 25.4 & 42.9 & 53.2 & 23.6\\ 
& CA2C\cite{sheng2025ca2c} & ICCV25 & IC & 86.3 & 95.7 & 96.8 & 60.1 & 62.8 & 79.3 & 84.5 & 42.3 & 26.3 & 44.8 & 54.1 & 25.2\\ 
& PLReMix\cite{liu2025plremix} & WACV25 & IC & 86.2 & 95.5 & 96.6 & 59.6 & 62.7 & 79.1 & 84.2 & 42.1 & 26.0 & 43.9 & 54.0 & 24.8\\  
& DULC\cite{xu2025revisiting} & AAAI25 & IC & 86.5 & 95.7 & 96.8 & 60.4 & 63.3 & 80.0 & 84.8 & 42.9 & 26.7 & 45.2 & 54.9 & 25.7\\

&DistributionNet*\cite{yu2019robust}&ICCV19&PR& 49.3 & 68.3 & 75.4 & 24.4 & 34.5 & 50.2 & 57.1 & 18.5&8.8&18.8&25.3&9.5\\  
& PurifyNet*\cite{ye2020purifynet} &TIFS20&PR  & 77.8 & 91.1 & 93.8 & 56.2 & 67.2 & 82.0 & 86.4 & 47.2&29.2&50.4&60.6&29.2\\      
&CORE*\cite{ye2022collaborative}&TIP22&PR & 81.8 & 92.5 & 95.8 & 63.2 & 73.0 & 85.4 & 89.0 & 54.4 & 31.7 & 51.8 & 61.4 & 30.0 \\
&LRNI-HSR*\cite{zhong2023neighborhood}&ICASSP23&PR& 84.1 & 93.6 & 95.9 & \textbf{65.6} & 75.0 & 86.3 & 89.7 & 55.9&-&-&-&-\\
&ICLR*\cite{zhong2024iclr}&PR24&PR & 83.6 & 93.5 & 95.7 & 64.9 & 76.8 & 86.8 & 90.3 & \textbf{56.5} &-&-&-&-\\
\cline{2-16}
& \textbf{CARE} (Ours) & This work & PR &\textbf{86.8}&\textbf{95.9}&\textbf{97.2}&64.6&\textbf{76.8}&\textbf{87.3}&\textbf{91.0}&56.2&\textbf{32.4}&\textbf{52.8}&\textbf{63.1}&\textbf{30.6}\\\hline

\end{tabular}}
\label{tab:5}
\end{center}
\caption{Performance results with state-of-the-art methods under different patterned noise ratios (10\% and 20\%). 
``IC'' denotes the methods designed for the image classification task, whereas ``PR'' refers to those methods developed for the person Re-ID task. ``*'' indicates the results are reported from their original papers, while the remaining results are reproduced under the same experimental settings. 
Best results are in \textbf{bold}.
}
\label{tab:results_patterned}
\end{table*}
\begin{table*}[t]  
\tiny
\label{tab:results_different_stage}
\begin{center}
\resizebox{0.96\linewidth}{!}{
\begin{tabular}{c|c|c|c|cccc|cccc|cccc}
\cline{1-16}
\multicolumn{4}{c|}{Datasets}&\multicolumn{4}{c|}{Market1501}&\multicolumn{4}{c|}{DukeMTMC-ReID}&\multicolumn{4}{c}{CUHK03}\\\hline

Noise Type & Noise &Stages & Net & Rank-1 & Rank-5 & Rank-10 & mAP & Rank-1 & Rank-5 & Rank-10 & mAP & Rank-1 & Rank-5 & Rank-10 & mAP\\\hline

\multirow{12}{*}{Random}

&\multirow{4}*{{0\%}}

&CORE (S1)&1&90.0&97.1&98.4&69.8&76.3&87.7&91.0&58.8&40.6&60.3&70.1&39.8\\
   
&& \textbf{CARE} (S1) &1&90.8&97.6&98.4&71.0&78.1&88.4&91.2&60.3&44.6&63.4&72.1&42.3\\

&& CORE (S1+S2) &2&\textbf{90.8}&97.6&\textbf{98.6}&72.0&78.4&88.1&91.3&\textbf{62.1}&43.6&64.4&73.1&\textbf{43.4}\\

&& \textbf{CARE} (S1+S2) &2&90.7&\textbf{97.7}&98.5&\textbf{72.5}&\textbf{78.4}&\textbf{88.2}&\textbf{91.3}&61.2&\textbf{45.2}&\textbf{64.5}&\textbf{73.9}&43.1\\
         
\cline{2-16}

&\multirow{4}*{{20\%}}

&CORE (S1) &1&86.4&95.4&97.0&57.7&67.6&81.3&85.1&47.0&24.9&43.3&54.1&23.7\\
   
&& \textbf{CARE} (S1) &1&87.1&96.0&97.6&60.9&69.7&82.8&87.7&50.5&25.3&42.5&53.2&24.0\\

&& CORE (S1+S2) &2&89.1&96.8&98.1&67.4&74.9&85.9&89.6&57.0&33.6&54.2&62.9&33.3\\

&&\textbf{CARE} (S1+S2)&2&\textbf{90.4}&\textbf{97.1}&\textbf{98.1}&\textbf{70.5}&\textbf{77.1}&\textbf{87.8}&\textbf{91.2}&\textbf{59.7}&\textbf{37.7}&\textbf{58.8}&\textbf{67.2}&\textbf{37.2}\\

\cline{2-16}

&\multirow{4}*{{50\%}}

&CORE (S1) &1&78.3&91.4&94.0&43.3&51.6&68.8&74.8&32.2&9.0&19.9&28.1&9.3\\
   
&& \textbf{CARE} (S1) &1&82.3&93.6&95.9&48.8&57.0&73.5&79.7&37.1&9.9&22.6&30.6&10.4\\

&& CORE (S1+S2) &2&81.0&92.3&94.8&48.3&56.3&72.2&77.9&37.8&13.0&27.5&35.9&14.4\\

&&\textbf{CARE} (S1+S2) &2&\textbf{83.7}&\textbf{94.6}&\textbf{96.1}&\textbf{54.6}&\textbf{62.4}&\textbf{77.2}&\textbf{82.0}&\textbf{43.4}&\textbf{20.7}&\textbf{38.1}&\textbf{47.4}&\textbf{21.2}\\
\hline\hline

\multirow{8}{*}{Patterned}

&\multirow{4}*{{10\%}}

& CORE (S1) &1&82.4&91.3&95.3&62.9&71.3&84.9&88.6&52.2&27.3&46.9&57.6&26.7\\
   
&& \textbf{CARE} (S1) &1&83.0&92.6&96.0&64.5&73.4&86.2&89.1&55.1&29.6&50.8&61.7&28.3\\

&& CORE (S1+ S2) &2&84.2&93.9&96.3&66.1&76.5&87.3&90.9&56.2&35.0&55.1&64.9&33.6\\

&& \textbf{CARE} (S1+S2) &2&\textbf{89.3}&\textbf{96.4}&\textbf{98.0}&\textbf{66.9}&\textbf{79.2}&\textbf{89.5}&\textbf{92.4}&\textbf{60.5}&\textbf{35.3}&\textbf{55.6}&\textbf{65.1}&\textbf{33.9}\\

\cline{2-16}

&\multirow{4}*{{20\%}}

& CORE (S1) &1&80.3&90.6&94.8&55.6&61.9&77.4&82.4&40.6&26.8&46.6&57.6&25.3\\
   
&& \textbf{CARE} (S1) &1&81.3&91.2&95.5&56.7&66.1&81.2&85.8&46.1&27.4&47.9&58.1&26.0\\

&& CORE (S1+S2) &2&81.8&92.5&95.8&63.2&73.0&85.4&89.0&54.4&31.7&51.8&61.4&30.0\\

&& \textbf{CARE} (S1+S2)&2&\textbf{86.8}&\textbf{95.9}&\textbf{97.2}&\textbf{64.6}&\textbf{76.8}&\textbf{87.3}&\textbf{91.0}&\textbf{56.2}&\textbf{32.4}&\textbf{52.8}&\textbf{63.1}&\textbf{30.6}\\

\hline
\end{tabular}}
\end{center}
\caption{Performance results in different stages under different random and patterned noise ratios on three person Re-ID datasets.
``S1'' and ``S2'' correspond to the \textbf{Calibration} stage and the \textbf{Refinement} stage, respectively. ``Net'' denotes the number of networks used for training. Best results are in \textbf{bold}.
}
\label{tab:results_sef}
\vspace{-5mm}
\end{table*}

\subsection{Comparison With the State-of-The-Art Methods}

\subsubsection{Competing Methods}
We compare the proposed \textbf{CARE} with state-of-the-art methods in the field of combating noisy labels, including: Co-teaching\cite{han2018co}, CleanNet\cite{lee2018cleannet}, SL\cite{wang2019symmetric}, MMT\cite{ge2020mutual}, Focal\cite{lin2020focal}, UbiW \cite{li2021learning}, ACT\cite{sheng2024enhancing}, SED~\cite{sheng2024foster}, CA2C~\cite{sheng2025ca2c}, PLReMix~\cite{liu2025plremix}, DULC~\cite{xu2025revisiting}, 
DistributionNet\cite{yu2019robust}, PurifyNet\cite{ye2020purifynet}, CORE\cite{ye2022collaborative}, LRNI-HSR\cite{zhong2023neighborhood}, LRP\cite{chen2023refining}, TSNT~\cite{liu2024two}, ICLR~\cite{zhong2024iclr}.
Note that the former eleven methods are designed for the image classification task, while the latter seven methods are proposed for the robust person Re-ID task. The comparison results on three person Re-ID datasets under different random and patterned noise ratios are shown in Tables~\ref{tab:results_random} and \ref{tab:results_patterned}, respectively.

\subsubsection{Random Noise}
The results presented in Table~\ref{tab:results_random} show that our \textbf{CARE} method achieves competitive results compared to all state-of-the-art methods under 10\%, 20\%, 30\%, and 50\% random noise ratio conditions. 
From the results in Table~\ref{tab:results_random}, we can observe that the methods designed for the image classification task show similarly limited performance results on person Re-ID datasets. The main reason lies in the fact that these methods simply filter out the noisy samples; thus, these methods are not well-suited for the pedestrian re-recognition task (the challenge of sparse per-identity samples).
Meanwhile, with the increasing of noise ratio, the performance of most methods is dramatically degraded. However, our proposed CARE method still maintains better performance with relatively minor degradation.
Moreover, the \textbf{CARE} method demonstrates significant superiority over noisy-label person Re-ID methods. Under 50\% random noise ratio, our method achieves the highest mAP and Rank‑1 accuracy on Market1501, DukeMTMC-ReID, and CUHK03, outperforming DistributionNet\cite{yu2019robust}, PurifyNet\cite{ye2020purifynet}, and CORE\cite{ye2022collaborative} by substantial margins. 
These significant improvements under noisy conditions demonstrate the efficacy of the \textbf{Calibration} stage, as well as the CAM and COSW in the \textbf{Refinement} stage. 
In summary, \textbf{CARE} achieves competitive performance and exhibits good robustness as label noise increases.



\subsubsection{Patterned Noise}
As shown in Table~\ref{tab:results_patterned}, we also evaluate the performance results under 10\% and 20\% patterned noise ratio conditions on three person Re-ID datasets. We find that our proposed \textbf{CARE} method also exhibits robust performance.
Notably, under 10\% patterned noise, our \textbf{CARE} method shows significant improvements in Rank-1 accuracy. It outperforms the latest method DULC~\cite{xu2025revisiting} by 0.8\% on Market1501 and by 4.4\% on DukeMTMC-ReID, and also surpasses the specialized Re-ID method ICLR~\cite{zhong2024iclr} by 2.5\% and 0.8\% on these two datasets, respectively.
Similarly, under 20\% patterned noise, the Rank-1 accuracy of our method is 3.2\% higher than ICLR~\cite{zhong2024iclr} and competitive with other leading methods on Market1501. It also shows a clear advantage over the latest methods in 2025, including DULC~\cite{xu2025revisiting}, PLReMix~\cite{liu2025plremix}, and CA2C~\cite{sheng2025ca2c} on DukeMTMC-ReID and CUHK03.
However, the mAP of our method on Market1501 under 10\% patterned noise ratio remains unsatisfactory, likely due to insufficient top-$k$ sample optimization in CAM.
Overall, the competitive performance improvement of our method under different patterned noise ratios demonstrates its robustness.

\subsection{Self Evaluation}

As shown in Table~\ref{tab:results_sef}, the proposed two-stage \textbf{CARE} method substantially outperforms the CORE baseline across all person datasets under different random (0\%, 20\%, and 50\%) and patterned (10\% and 20\%) noise ratios, respectively. 
Even under clean annotations (\textit{i.e.}, 0\% noise ratio), \textbf{CARE} (S1) achieves relatively higher accuracy than CORE (S1), and the full method \textbf{CARE} (S1 + S2) further improves performance. 
As the noise ratio increases, our method performs significantly better than CORE, showing the good performance.
For example, under 50\% random noise ratio on Market1501, \textbf{CARE} (S1 + S2) attains 83.7\% Rank-1 and 54.6\% mAP, 
outperforming CORE (S1 + S2) by 2.7\% and 6.3\% in Rank-1 and mAP, respectively.
Similar improvements are observed on DukeMTMC-ReID and CUHK03, where \textbf{CARE} consistently delivers higher Rank-$k$ accuracy and mAP than both CORE (S1) and CORE (S1 + S2).
Notably, the \textbf{Refinement} stage contributes significantly to mAP under heavy noise on three person Re-ID datasets (\textit{e.g.}, 4.6\%$\sim$10.8\% improvement in mAP under 50\% random and 20\% patterned noise ratios), demonstrating that the \textbf{Refinement} stage enhances discriminative capability. 
Overall, \textbf{CARE} consistently achieves the better retrieval accuracy, confirming stronger robustness to noisy labels compared to CORE. 
This can be attributed to its probabilistic evidence propagation from
calibration to refinement: the \textbf{Calibration} stage employs Dirichlet-informed calibration to produce more reliable initial predictions, while the \textbf{Refinement} stage uses these as pseudo-labels and applies hyperspherical weighting to refine feature distribution. 
\begin{table*}[t]  
\tiny
\begin{center}
\resizebox{\linewidth}{!}{
\begin{tabular}{c|c|ccc|cccc|cccc|cccc}
\cline{1-17}
\multicolumn{5}{c|}{Datasets} & \multicolumn{4}{c|}{Market1501} & \multicolumn{4}{c|}{DukeMTMC-ReID} & \multicolumn{4}{c}{CUHK03} \\\hline

Noise Type & Stages & $\mathcal{L}_{\mathrm{ENLL}}$ & $\mathcal{L}_{\mathrm{KL}}^{(Dir)}$ & & Rank-1 & Rank-5 & Rank-10 & mAP & Rank-1 & Rank-5 & Rank-10 & mAP & Rank-1 & Rank-5 & Rank-10 & mAP \\\hline

\multirow{7}{*}{Random}
& \multirow{2}{*}{\textbf{Calibration}}
 & \checkmark &&& 85.4 & 95.2 & 97.1 & 56.8 & 67.5 & 81.6 & 85.6 & 46.8 & 19.4 & 37.0 & 46.8 & 19.2 \\
&  & \checkmark & \checkmark && 87.1 & 96.0 & 97.6 & 60.9 & 69.7 & 82.8 & 87.7 & 50.5 & 25.3 & 42.5 & 53.2 & 24.0 \\\cline{2-17}

&&$\mathcal{L}_{\mathrm{WCE}}$ (w/o \text{Noisy})&$\mathcal{L}_{\mathrm{WCE}}$ (w/ \text{Noisy}) &$\mathcal{L}_{\mathrm{WKL}}$ & Rank-1 & Rank-5 & Rank-10 & mAP & Rank-1 & Rank-5 & Rank-10 & mAP & Rank-1 & Rank-5 & Rank-10 & mAP\\\cline{2-17}
& \multirow{3}{*}{\textbf{Calibration} + \textbf{Refinement}}
 & \checkmark &&& 88.0 & 96.2 & 97.5 & 64.0 & 73.0 & 85.6 & 89.3 & 55.0 & 28.3 & 47.1 & 57.1 & 27.2 \\
&  & & \checkmark && 89.0 & 97.0 & 98.0 & 66.0 & 73.7 & 86.0 & 90.1 & 55.5 & 29.2 & 47.3 & 58.1 & 27.7 \\
&  & \checkmark && \checkmark & 89.2 & 96.9 & 98.1 & 67.1 & 73.6 & 86.4 & 89.7 & 56.3 & 31.3 & 52.4 & 62.0 & 30.7 \\\cline{2-17}
& \textbf{CARE} (Ours) & & \checkmark & \checkmark & \textbf{90.4} & \textbf{97.1} & \textbf{98.1} & \textbf{70.5} & \textbf{77.1} & \textbf{87.8} & \textbf{91.2} & \textbf{59.7} & \textbf{37.7} & \textbf{58.8} & \textbf{67.2} & \textbf{37.2} \\\hline\hline

\multirow{8}{*}{Patterned}

&Stages&$\mathcal{L}_{\mathrm{ENLL}}$ & $\mathcal{L}_{\mathrm{KL}}^{(Dir)}$ & & Rank-1 & Rank-5 & Rank-10 & mAP & Rank-1 & Rank-5 & Rank-10 & mAP & Rank-1 & Rank-5 & Rank-10 & mAP \\\cline{2-17}
& \multirow{2}{*}{\textbf{Calibration}}
 & \checkmark &&& 82.9 & 94.5 & 96.5 & 55.9 & 62.3 & 78.9 & 84.5 & 42.0 & 23.5 & 42.9 & 52.8 & 22.1 \\
 && \checkmark & \checkmark && 83.3 & 94.2 & 96.5 & 56.7 & 66.1 & 81.2 & 85.8 & 46.1 & 27.4 & 47.9 & 58.1 & 26.0 \\\cline{2-17}
&& $\mathcal{L}_{\mathrm{WCE}}$ (w/o \text{Noisy})&$\mathcal{L}_{\mathrm{WCE}}$ (w/ \text{Noisy}) & $\mathcal{L}_{\mathrm{WKL}}$ & Rank-1 & Rank-5 & Rank-10 & mAP & Rank-1 & Rank-5 & Rank-10 & mAP & Rank-1 & Rank-5 & Rank-10 & mAP\\\cline{2-17}
& \multirow{3}{*}{\textbf{Calibration} + \textbf{Refinement}}
 & \checkmark &&& 85.6 & 95.6 & 97.2 & 59.7 & 70.6 & 82.7 & 87.9 & 52.1 & 29.9 & 50.4 & 61.4 & 28.4 \\
&  &  & \checkmark && 85.5 & 95.0 & 96.9 & 59.8 & 72.3 & 83.5 & 86.3 & 51.4 & 30.6 & 51.2 & 62.3 & 29.8 \\
&  & \checkmark && \checkmark & 86.4 & 95.9 & 97.2 & 62.9 & 73.9 & 84.0 & 88.0 & 55.2 & 31.2 & 52.0 & 62.5 & 30.1 \\\cline{2-17}
& \textbf{CARE} (Ours) & & \checkmark & \checkmark & \textbf{86.8} & \textbf{95.9} & \textbf{97.2} & \textbf{64.6} & \textbf{76.8} & \textbf{87.3} & \textbf{91.0} & \textbf{56.2} & \textbf{32.4} & \textbf{52.8} & \textbf{63.1} & \textbf{30.6} \\\hline

\end{tabular}}
\end{center}
\caption{Performance results with different losses under 20\% random and patterned noises on three person Re-ID datasets. Best results are in \textbf{bold}.}
\label{tab:results_ablation}
\end{table*}


\begin{table*}[t]  
\tiny
\label{tab:results_different_batchsize}
\begin{center}
\resizebox{\linewidth}{!}{
\begin{tabular}{c|c|c|cccc|cccc|cccc}
\cline{1-15}
\multicolumn{3}{c|}{Datasets}&\multicolumn{4}{c|}{Market1501}&\multicolumn{4}{c|}{DukeMTMC-ReID}&\multicolumn{4}{c}{CUHK03}\\\hline

Noise Type & Noise &Batch Size & Rank-1 & Rank-5 & Rank-10 & mAP & Rank-1 & Rank-5 & Rank-10 & mAP & Rank-1 & Rank-5 & Rank-10 & mAP\\\hline

\multirow{15}{*}{Random}

&\multirow{5}*{{0\%}}

&8 &79.0&92.1&94.8&45.2 &46.5&64.9&71.2&30.0 &22.9&44.0&54.4&24.6 \\
   
&&16 &90.6&97.2&97.9&69.7 &73.9&86.0&89.8&57.1 &\textbf{45.9}&\textbf{65.9}&\textbf{74.5}&\textbf{45.0} \\

&&32 &\textbf{90.7} &\textbf{97.7} &\textbf{98.5} &\textbf{72.5} &\textbf{78.4} &\textbf{88.2} &\textbf{91.3} &\textbf{61.2} &45.2 &64.5 &73.9 &43.1 \\

&&64 &90.1&97.5&98.2&71.3 &77.4&87.6&90.9&59.5 &35.5&56.9&66.4&36.2 \\

&&128 &89.8&97.2&97.9&68.6 &75.8&87.3&90.4&58.1 &30.2&48.6&58.6&30.0 \\
         
\cline{2-15}

&\multirow{5}*{{20\%}}

&8 &76.2&88.9&92.9&41.8 &42.7&60.5&67.5&26.3 &17.3&35.3&45.6&18.6 \\
   
&&16 &86.3&95.0&96.4&59.5 &69.0&81.2&85.8&50.9 &34.9&56.6&66.1&35.0 \\

&&32 &\textbf{90.4}&\textbf{97.1}&\textbf{98.1}&\textbf{70.5}&\textbf{77.1}&\textbf{87.8}&\textbf{91.2}&\textbf{59.7}&\textbf{37.7}&\textbf{58.8}&\textbf{67.2}&\textbf{37.2} \\

&&64 &89.2&96.8&98.0&67.0 &73.8&85.9&89.5&56.3 &25.7&45.8&55.1&25.8 \\ 

&&128 &88.0&96.6&97.9&63.7 &71.0&83.5&87.9&51.6 &15.2&29.0&38.2&15.0 \\

\cline{2-15}

&\multirow{5}*{{50\%}}

&8 &57.2&77.1&83.1&25.4 &16.7&30.6&37.6&9.0 &7.3&16.1&23.4&8.5 \\
   
&&16 &80.0&92.5&95.2&50.1 &57.0&73.5&79.3&38.2 &19.2&36.4&46.6&20.2 \\

&&32 &83.7&\textbf{94.6}&\textbf{96.1}&\textbf{54.6}&\textbf{62.4}&\textbf{77.2}&\textbf{82.0}&\textbf{43.4}&\textbf{20.7}&\textbf{38.1}&\textbf{47.4}&\textbf{21.2}\\

&&64 &\textbf{84.0}&93.9&96.1&50.8 &62.2&76.9&81.6&42.3 &10.6&23.7&31.9&11.5 \\

&&128 &81.6&92.9&95.4&46.8 &56.3&72.5&78.1&36.1 &6.2&13.0&18.1&6.3 \\

\hline\hline

\multirow{10}{*}{Patterned}

&\multirow{5}*{{10\%}}

&8 &79.2&90.6&93.7&47.3 &54.5&70.8&77.4&35.0 &21.8&42.6&53.4&22.8 \\ 
   
&&16 &\textbf{89.6}&\textbf{96.8}&97.5&\textbf{68.3} &74.4&85.9&89.7&57.0 &34.3&\textbf{57.2}&\textbf{67.6}&33.3 \\

&&32 &89.3&96.4&\textbf{98.0}&66.9&\textbf{79.2}&\textbf{89.5}&\textbf{92.4}&\textbf{60.5}&\textbf{35.3}&55.6&65.1&\textbf{33.9}\\

&&64 &88.8&96.5&97.8&67.0 &73.2&85.4&89.3&54.2 &31.2&51.3&61.5&29.9 \\

&&128 &88.3&96.6&98.0&64.5 &73.3&84.6&89.2&53.5 &19.9&37.9&47.9&20.0 \\ 

\cline{2-15}

&\multirow{5}*{{20\%}}

&8 &78.3&91.2&94.2&45.6 &46.8&65.6&72.3&29.2 &17.6&36.4&47.4&19.0 \\
   
&&16 &\textbf{87.6}&95.0&96.7&61.6 &68.3&81.2&86.0&49.7 &32.2&52.7&61.8&30.5 \\ 

&&32 &86.8&95.9&97.2&\textbf{64.6}&\textbf{76.8}&\textbf{87.3}&\textbf{91.0}&\textbf{56.2}&\textbf{32.4}&\textbf{52.8}&\textbf{63.1}&\textbf{30.6}\\ 

&&64 &87.2&\textbf{96.6}&\textbf{97.9}&63.1 &70.1&82.9&87.3&49.7 &24.1&43.7&54.4&23.3 \\

&&128 &86.3&95.9&97.7&60.3 &69.7&83.9&87.6&49.1 &18.7&35.4&44.4&18.3 \\

\hline
\end{tabular}}
\end{center}
\caption{Performance results with different batch sizes (8, 16, 32, 64, 128) under random and patterned noise ratios on three person Re-ID datasets.
``Batch Size'' denotes the number of samples in each training iteration. 
Best results are in \textbf{bold}.}
\label{tab:results_batchsize}
\vspace{-2mm}
\end{table*}

\subsection{Ablation Studies}

To validate the effectiveness of different losses, we evaluate our method on three person Re-ID datasets under 20\% random and patterned noise ratios, as shown in Table~\ref{tab:results_ablation}.
Meanwhile, the results of hyperparameter sensitivity experiments are shown in Fig.~\ref{fig:hyperparams_compact}. The detailed analyses are as follows:

\subsubsection{Effect of Losses in the Calibration Stage}
As shown in Table~\ref{tab:results_ablation}, under random noise, the Rank-1 and mAP based on $\mathcal{L}_{\mathrm{ENLL}}$ correspond to 85.4\%/67.5\%/19.4\% and 56.8\%/46.8\%/19.2\% on three person Re-ID datasets. 
Then, after we adopt $\mathcal{L}_{\mathrm{KL}}^{(Dir)}$, Rank-1 and mAP improved by 1.7\%/2.2\%/5.9\% and 4.1\%/3.7\%/4.8\% respectively on three person Re-ID datasets. 
Similarly, under patterned noise, the performance results are improved by 0.7\%/3.8\%/3.9\% in Rank-1 and 0.8\%/4.1\%/3.9\% in mAP on three person Re-ID datasets. All these experimental results demonstrate the effectiveness of the \textbf{Calibration} stage.

\subsubsection{Effect of Losses in the Refinement Stage}
From Table~\ref{tab:results_ablation}, after finishing the \textbf{Refinement} stage, the performance results of our method show significant improvement on all three person Re-ID datasets, regardless of whether the random noise or patterned noise is considered.
As well, for $\mathcal{L}_\mathrm{WCE}$, the performance results show improvement regardless of whether the noisy samples are included in Eq.~(\ref{eq:R}).
Compared to $\mathcal{L}_\mathrm{WCE}$ (w/o Noisy), the performance results using $\mathcal{L}_\mathrm{WCE}$ (w/ Noisy) are improved across all datasets under random noise.
Under patterned noise, $\mathcal{L}_\mathrm{WCE}$ (w/ Noisy) shows limited improvement over $\mathcal{L}_\mathrm{WCE}$ (w/o Noisy) on Market1501 and CUHK03, but gains significant improvement on DukeMTMC-ReID. 
It is because our CAM demonstrates more noticeable effectiveness on diverse datasets, and DukeMTMC-ReID contains sufficient diverse data to substantiate this point.
Obviously, combining $\mathcal{L}_\mathrm{WCE}$ (w/ CAM) and $\mathcal{L}_\mathrm{WKL}$ achieves improvements of 1.4\%/2.7\%/8.5\% and 5.2\%/4.2\%/9.5\% in Rank-1 and mAP, respectively, under random noise; under patterned noise, it accomplishes improvements of 1.3\%/4.5\%/1.8\% and 4.8\%/4.8\%/0.8\% in Rank-1 and mAP.
These experiments clearly demonstrate the effectiveness of each loss in the \textbf{Refinement} stage under random noise and patterned noise.

\begin{figure*}[t]
  \centering
  \includegraphics[width=0.23\linewidth, height=2.5cm]{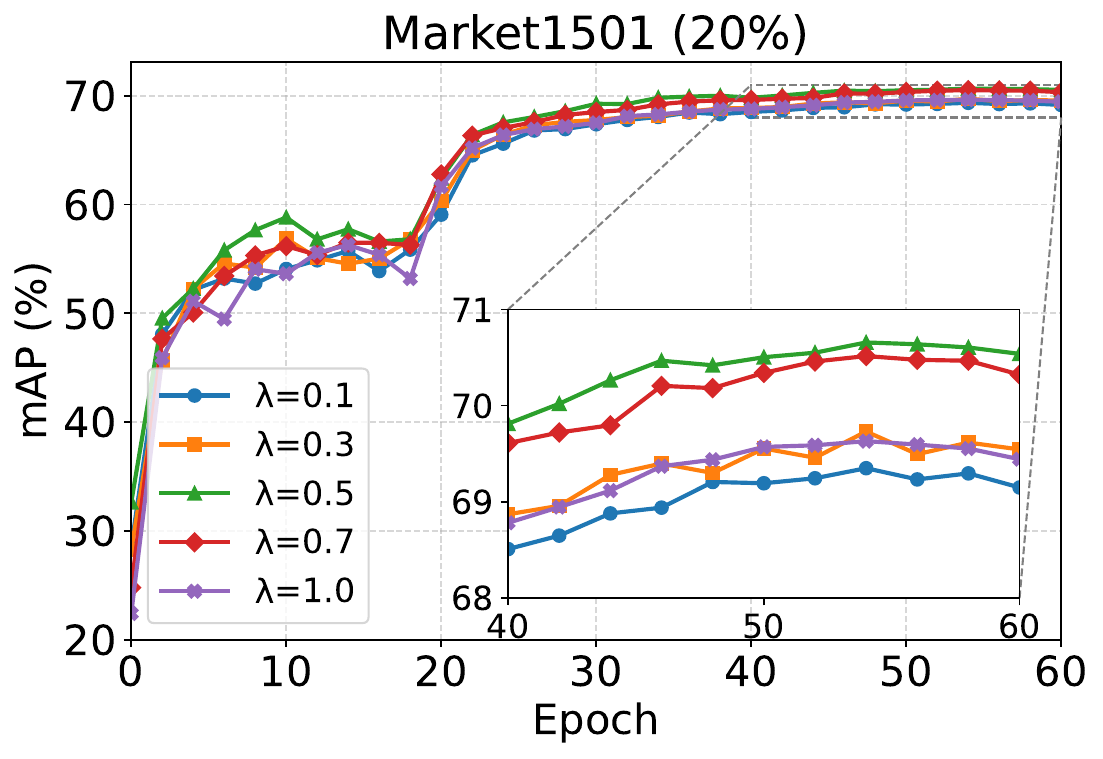}%
  \hspace{2pt}
  \includegraphics[width=0.23\linewidth, height=2.5cm]{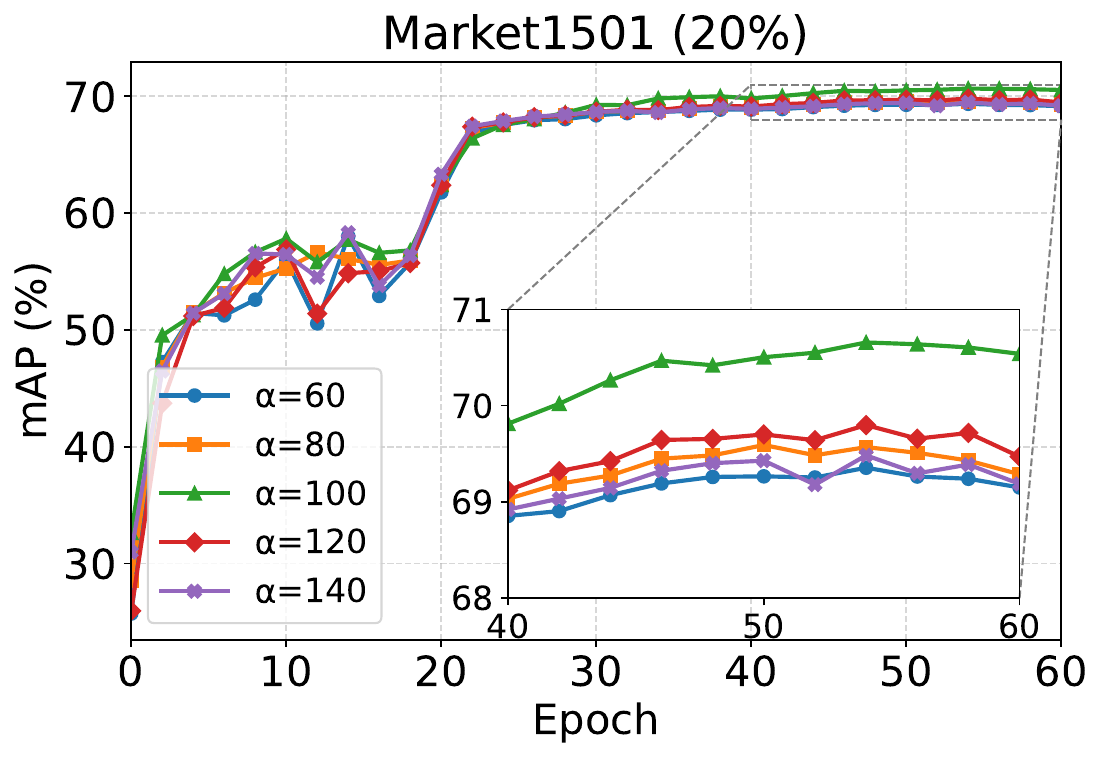}%
  \hspace{2pt}
  \includegraphics[width=0.23\linewidth, height=2.5cm]{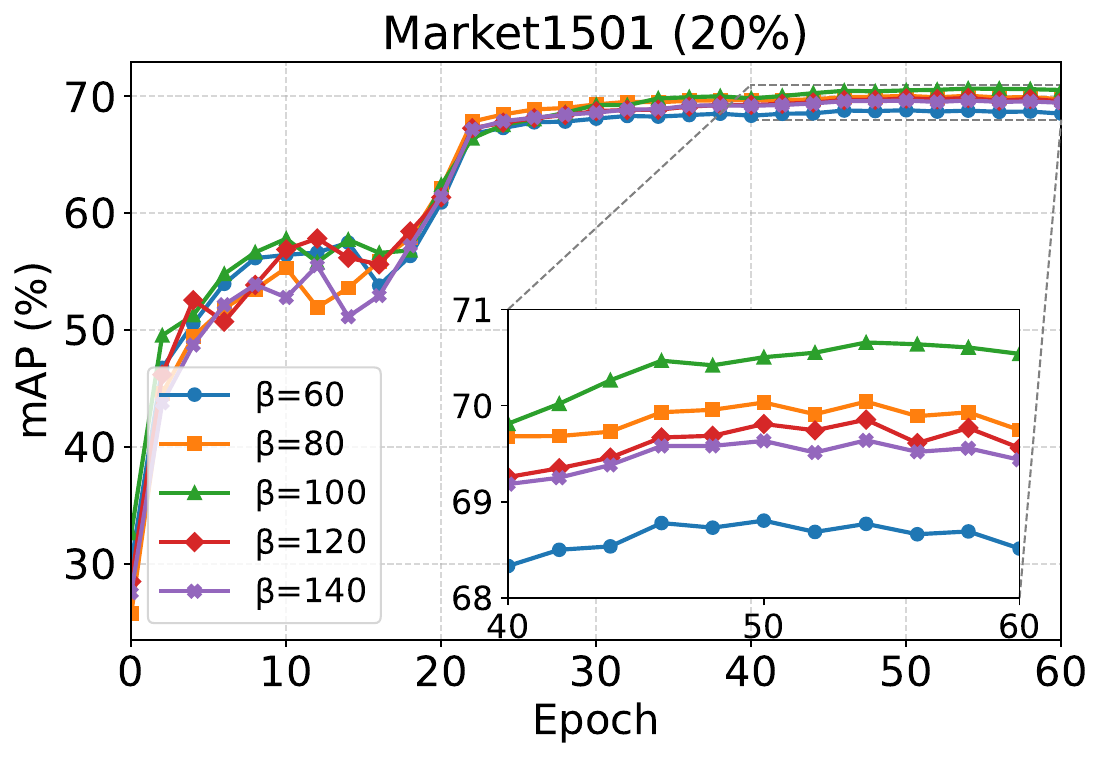}%
  \hspace{2pt}
  \includegraphics[width=0.23\linewidth, height=2.5cm]{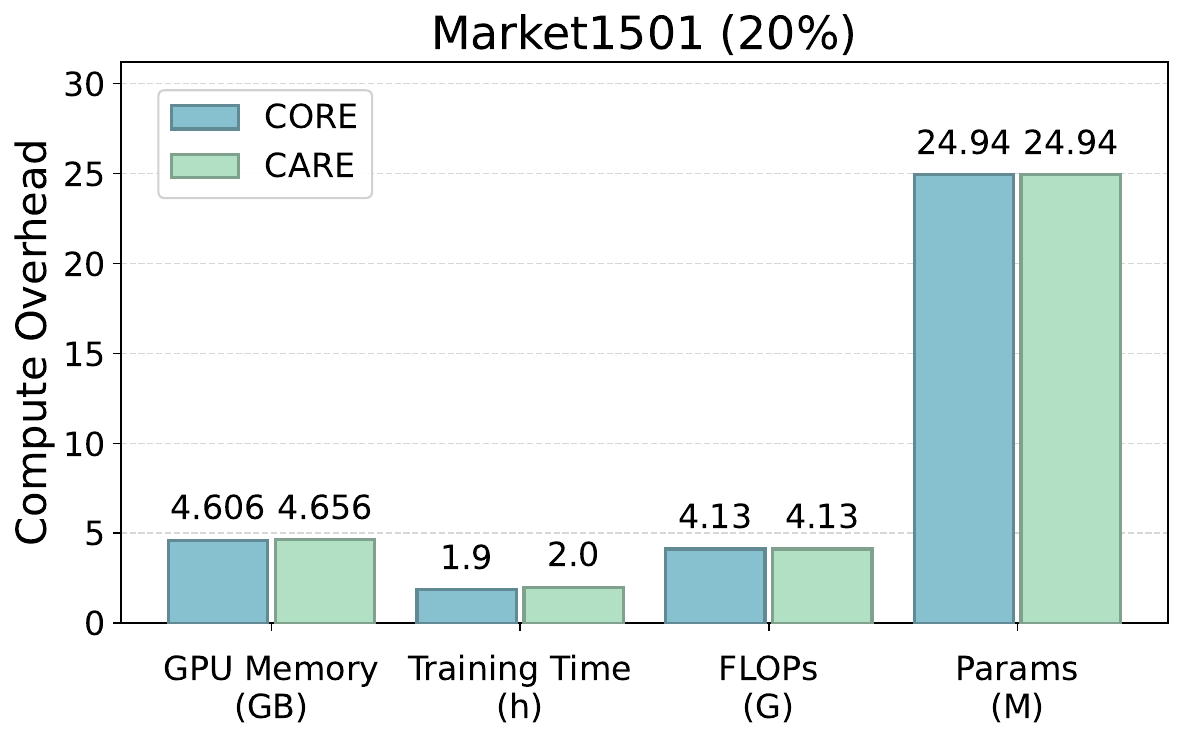}\\[\smallskipamount]
  
  \includegraphics[width=0.23\linewidth, height=2.5cm]{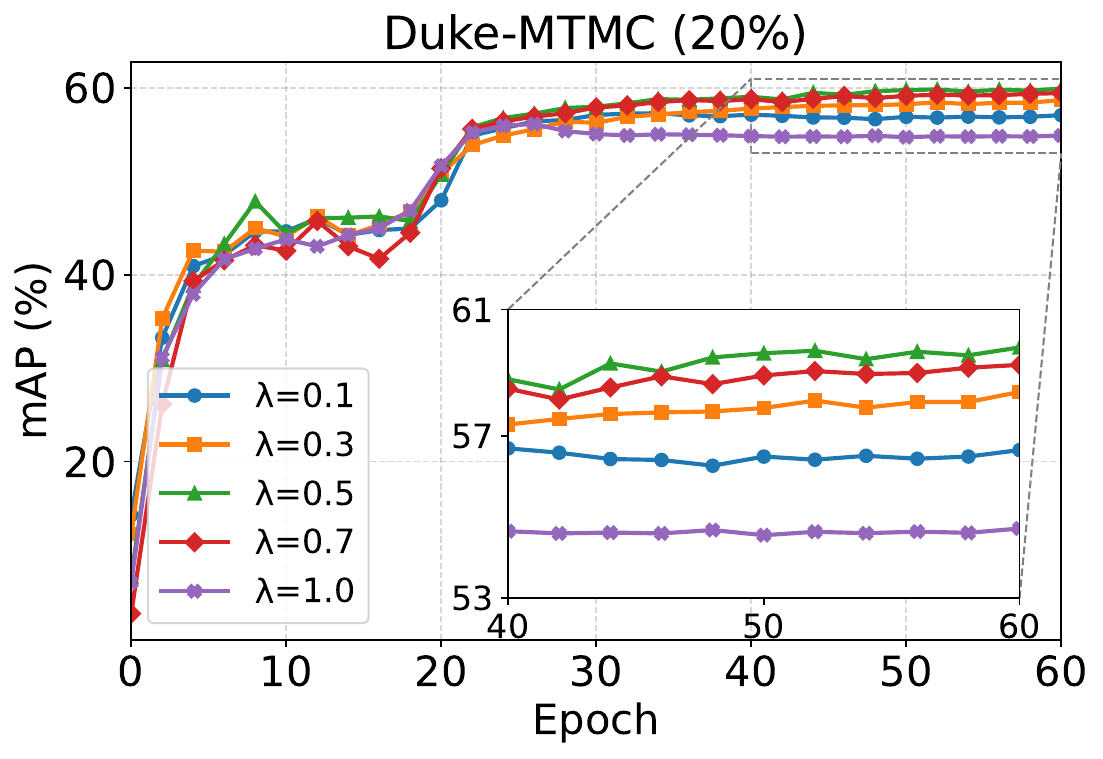}%
  \hspace{2pt}
  \includegraphics[width=0.23\linewidth, height=2.5cm]{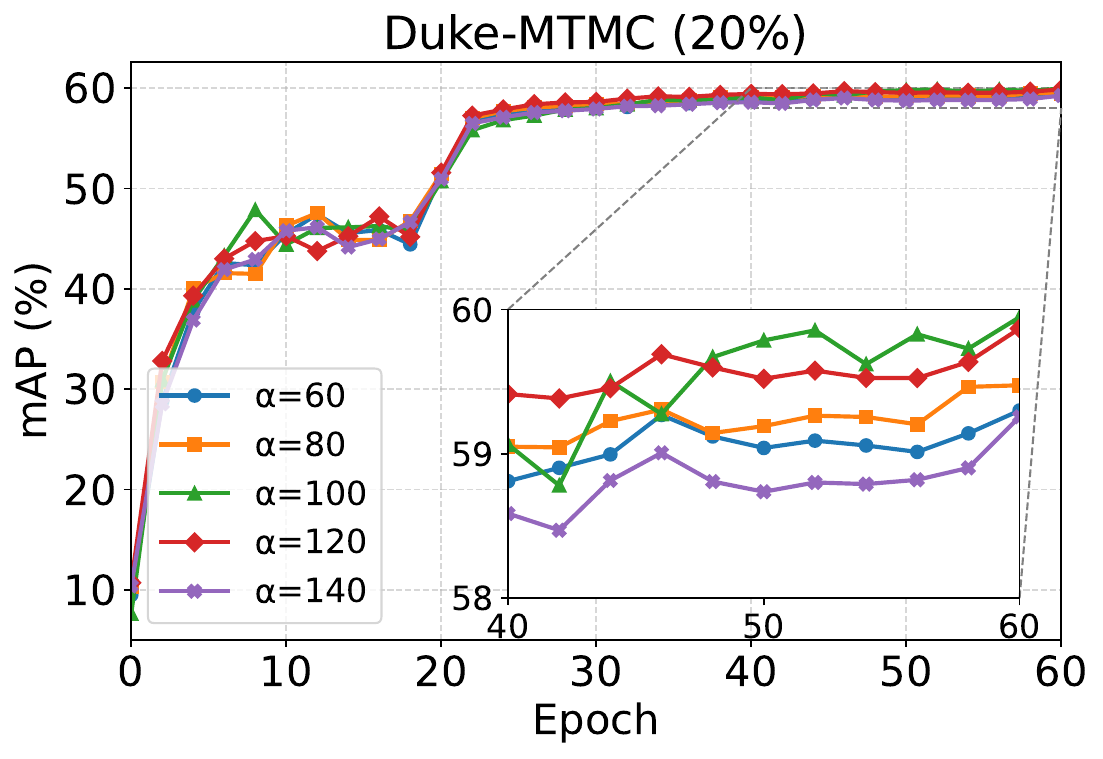}%
  \hspace{2pt}
  \includegraphics[width=0.23\linewidth, height=2.5cm]{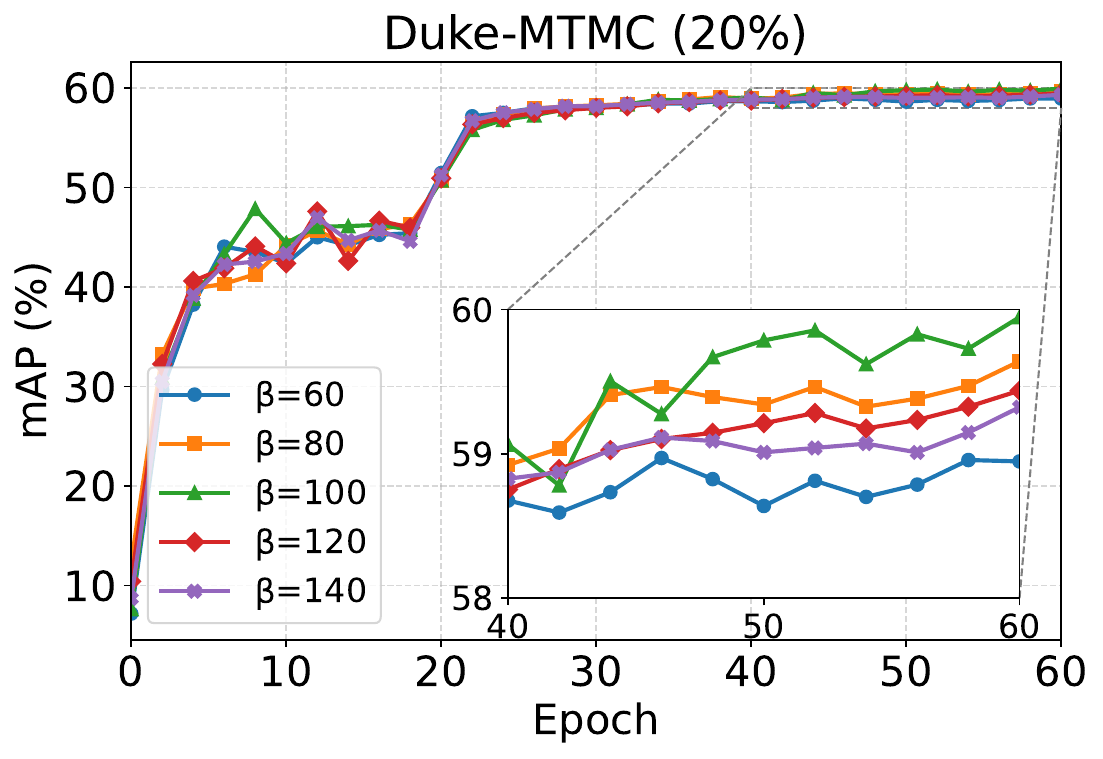}%
  \hspace{2pt}
  \includegraphics[width=0.23\linewidth, height=2.5cm]{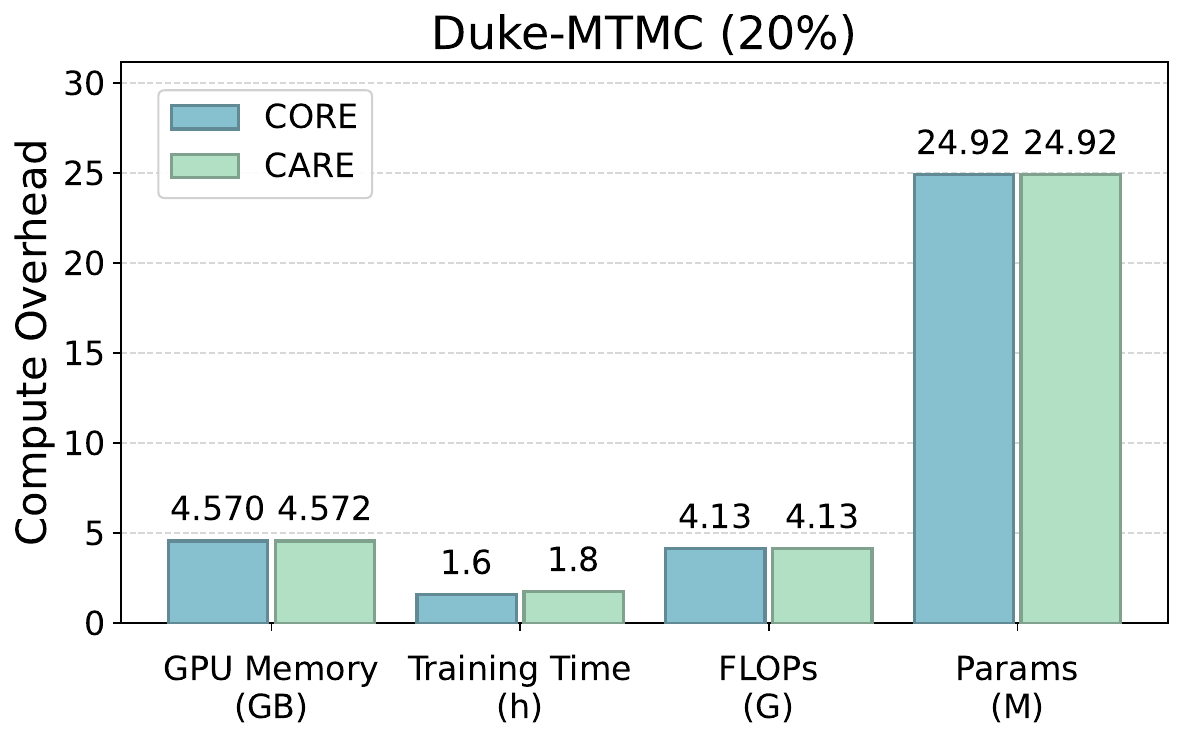}\\[\smallskipamount]
  
  \includegraphics[width=0.23\linewidth, height=2.5cm]{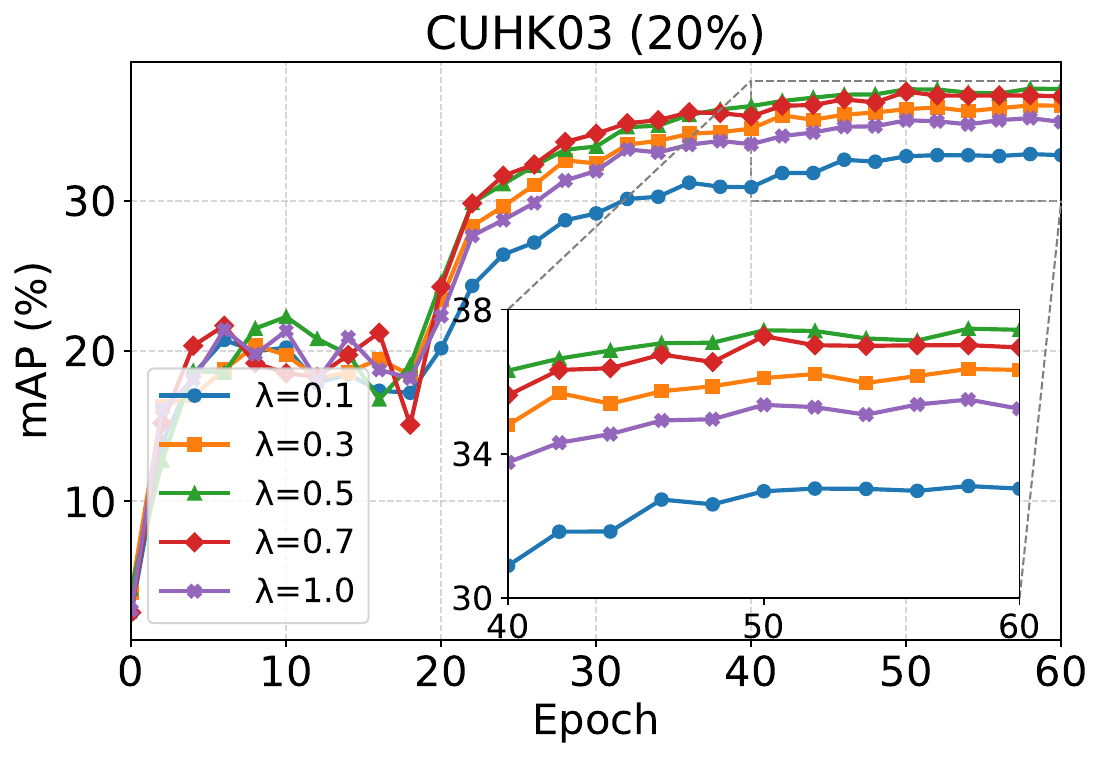}%
  \hspace{2pt}
  \includegraphics[width=0.23\linewidth, height=2.5cm]{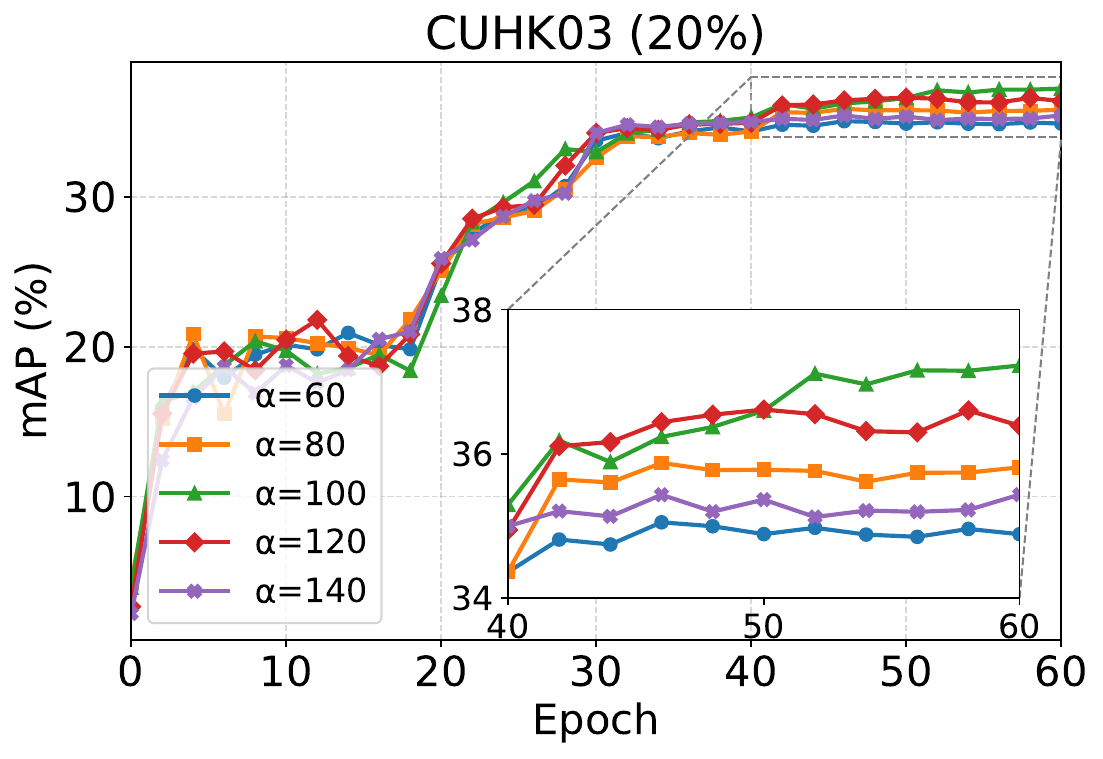}%
  \hspace{2pt}
  \includegraphics[width=0.23\linewidth, height=2.5cm]{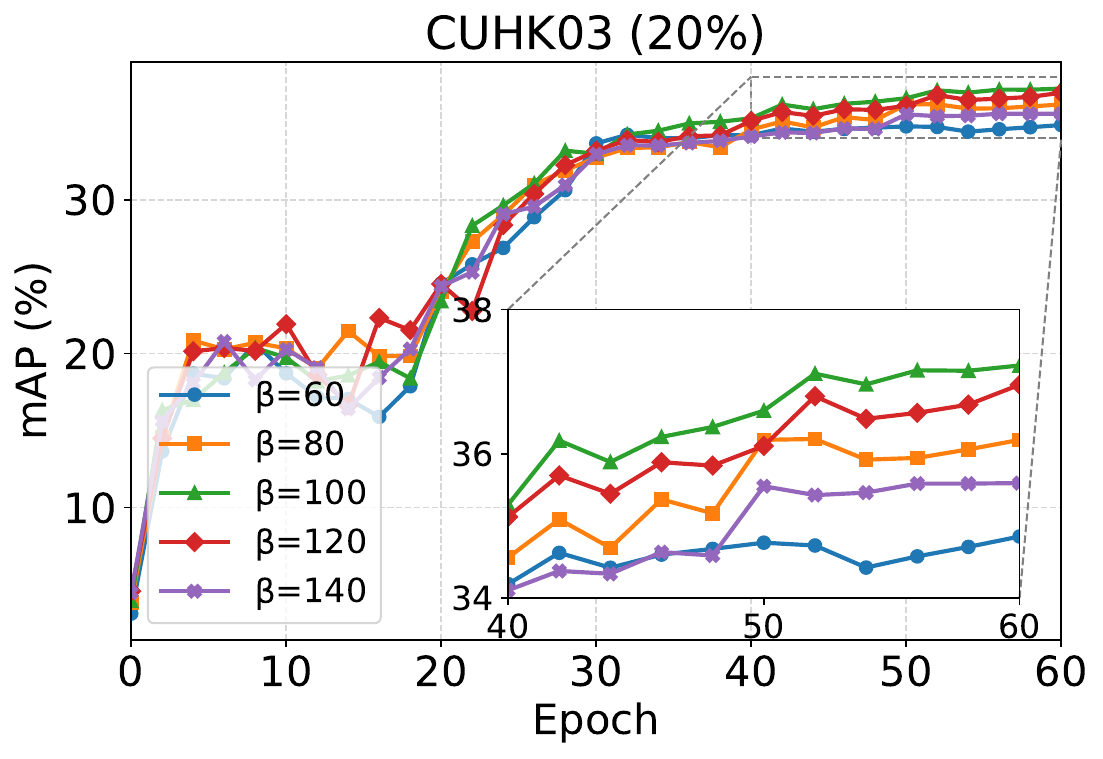}%
  \hspace{2pt}
  \includegraphics[width=0.23\linewidth, height=2.5cm]{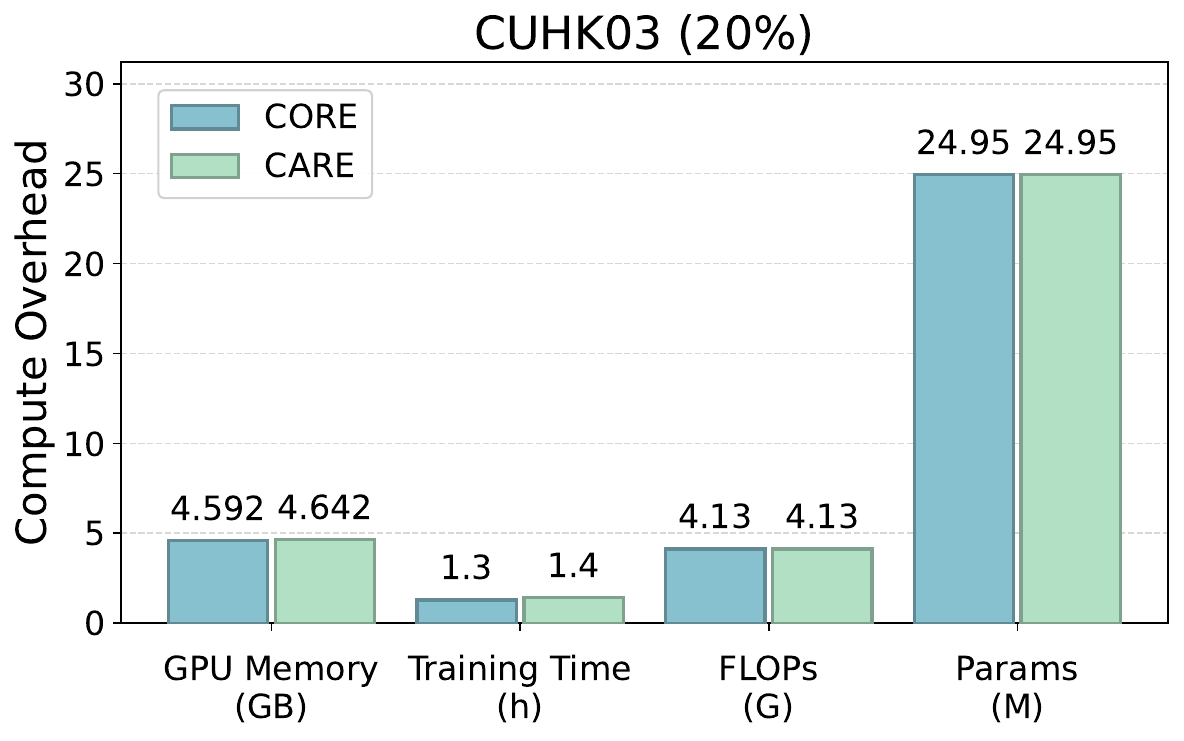}
  
  \caption{
  Ablation results of hyperparameters ($\lambda, \alpha, \beta$) and compute overhead under 20\% random noise ratio. 
  Each row shows results on one person Re-ID dataset: Market1501 (top), DukeMTMC-ReID (middle), and CUHK03 (bottom). From left to right, the columns detail: the ablation studies for hyperparameters \(\lambda\), \(\alpha\), and \(\beta\), followed by the compute overhead.
  }
  \vspace{-3mm}
  \label{fig:hyperparams_compact}
\end{figure*}

\begin{figure*}[t]
  \centering
  \includegraphics[width=0.23\linewidth, height=2.5cm]{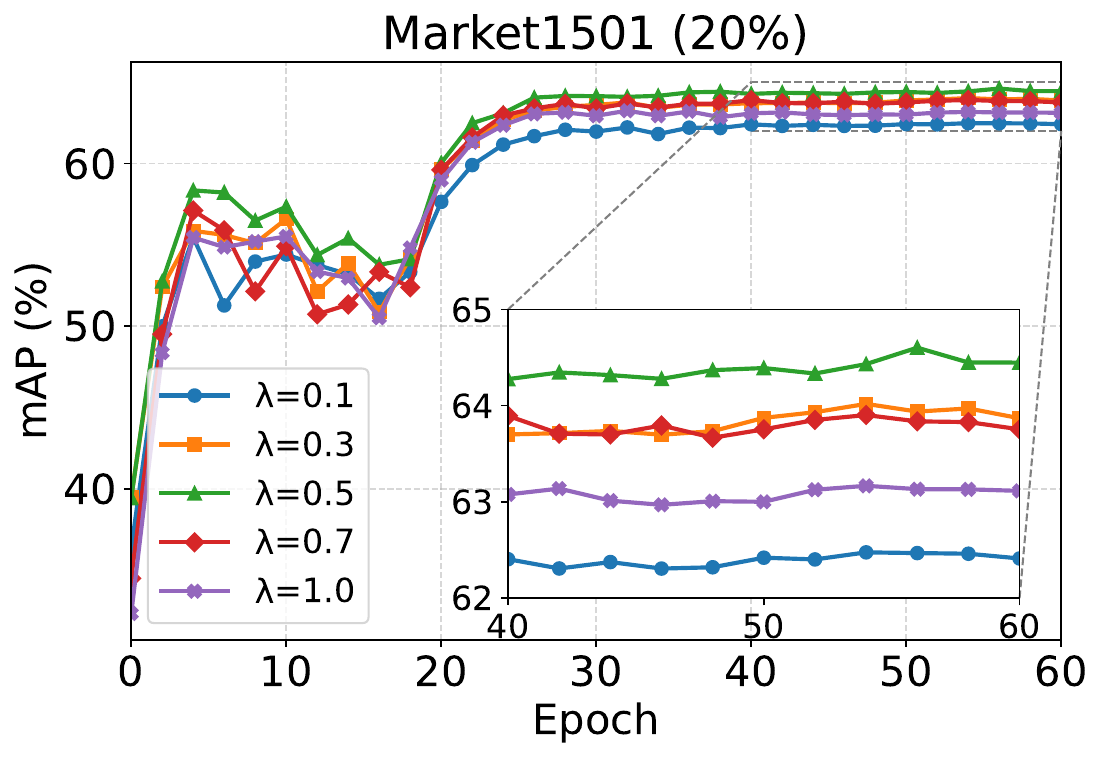}%
  \hspace{2pt}
  \includegraphics[width=0.23\linewidth, height=2.5cm]{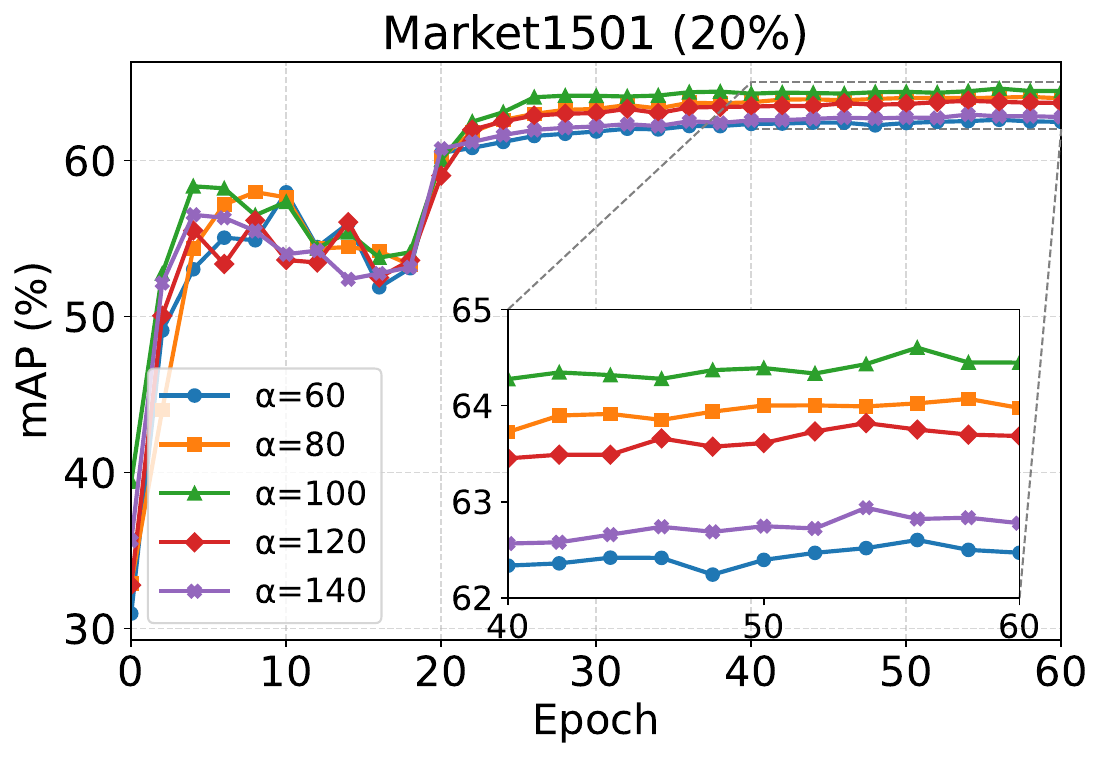}%
  \hspace{2pt}
  \includegraphics[width=0.23\linewidth, height=2.5cm]{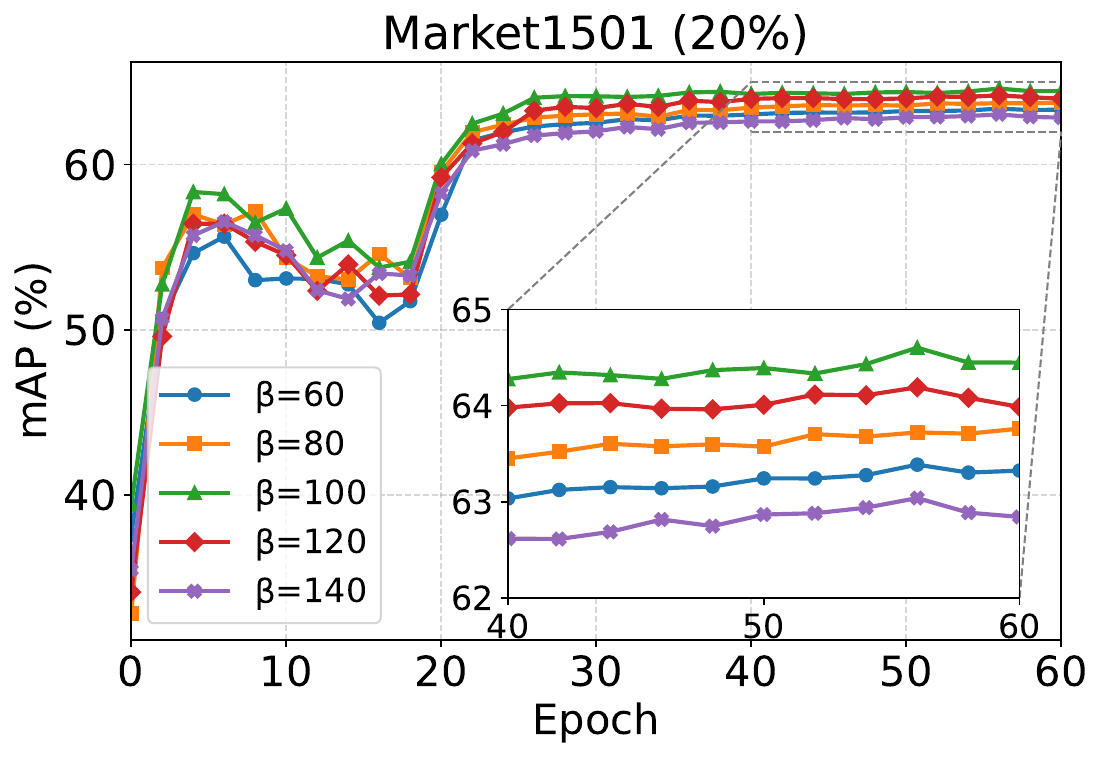}%
  \hspace{2pt}
  \includegraphics[width=0.23\linewidth, height=2.5cm]{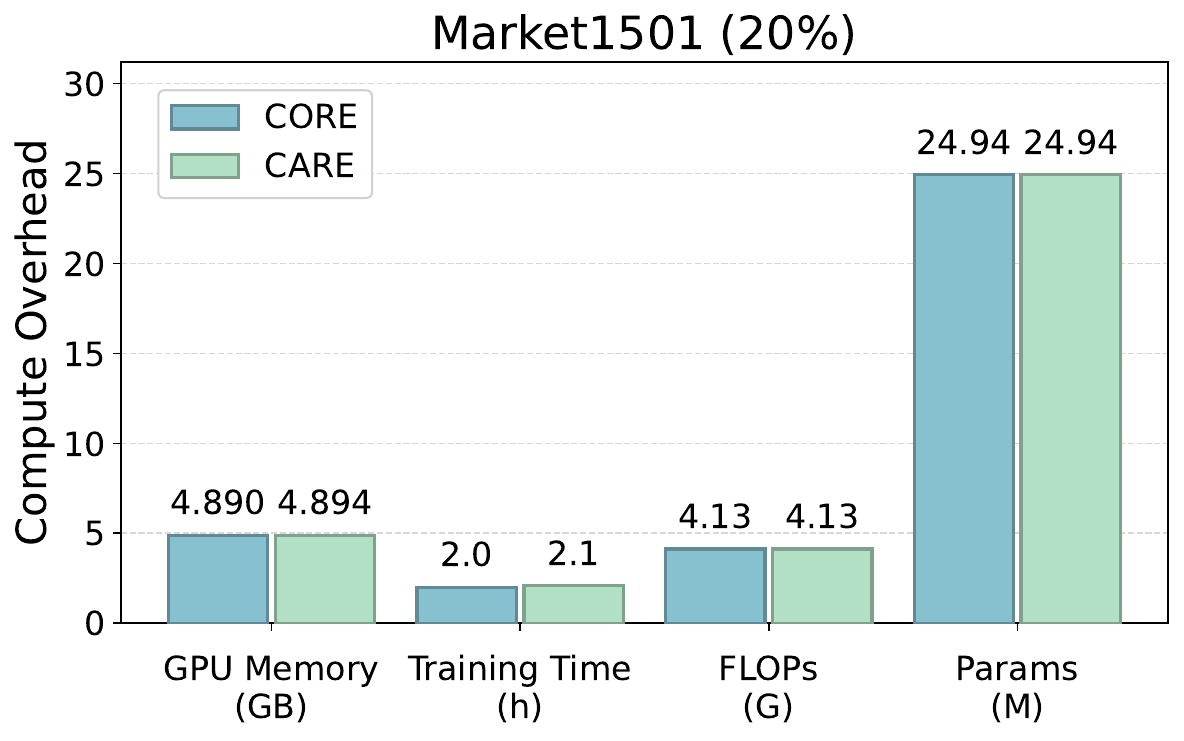}\\[\smallskipamount]
  
  \includegraphics[width=0.23\linewidth, height=2.5cm]{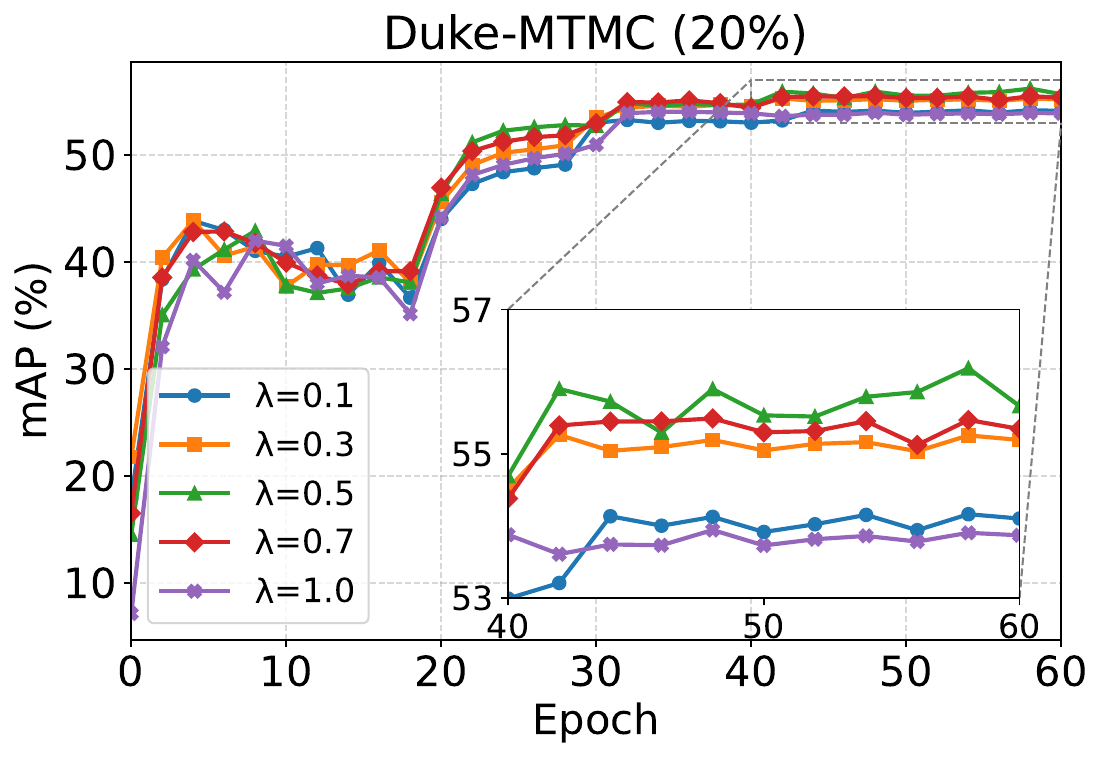}%
  \hspace{2pt}
  \includegraphics[width=0.23\linewidth, height=2.5cm]{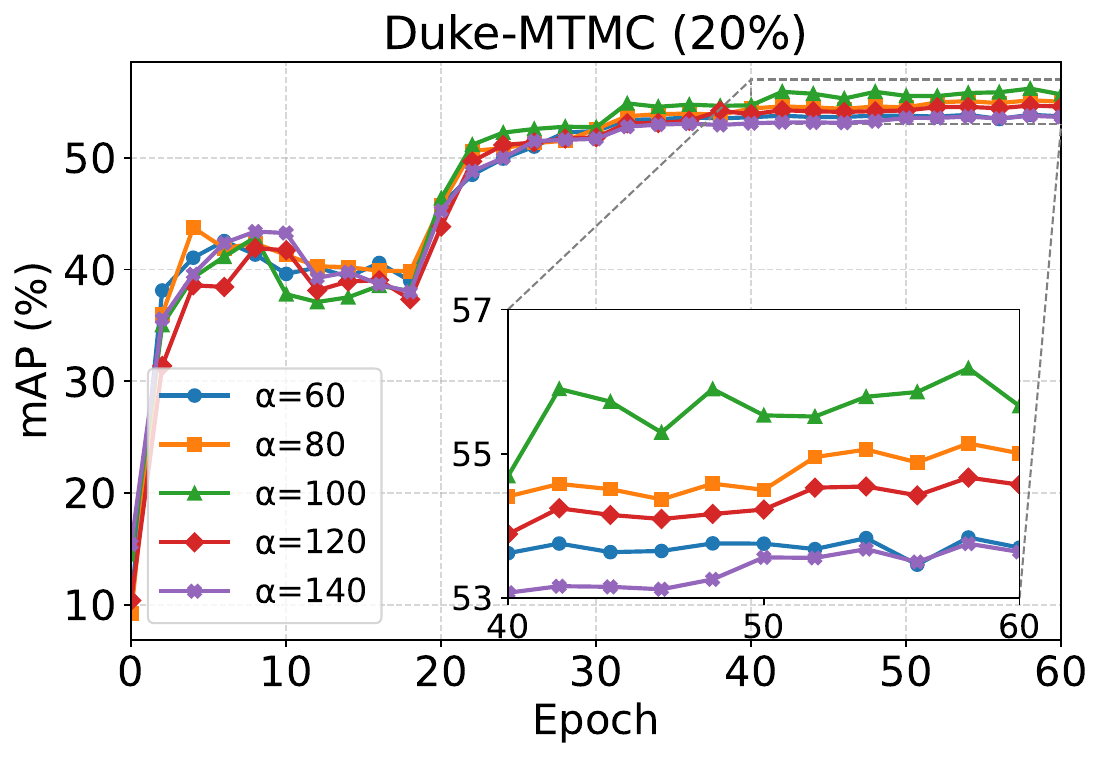}%
  \hspace{2pt}
  \includegraphics[width=0.23\linewidth, height=2.5cm]{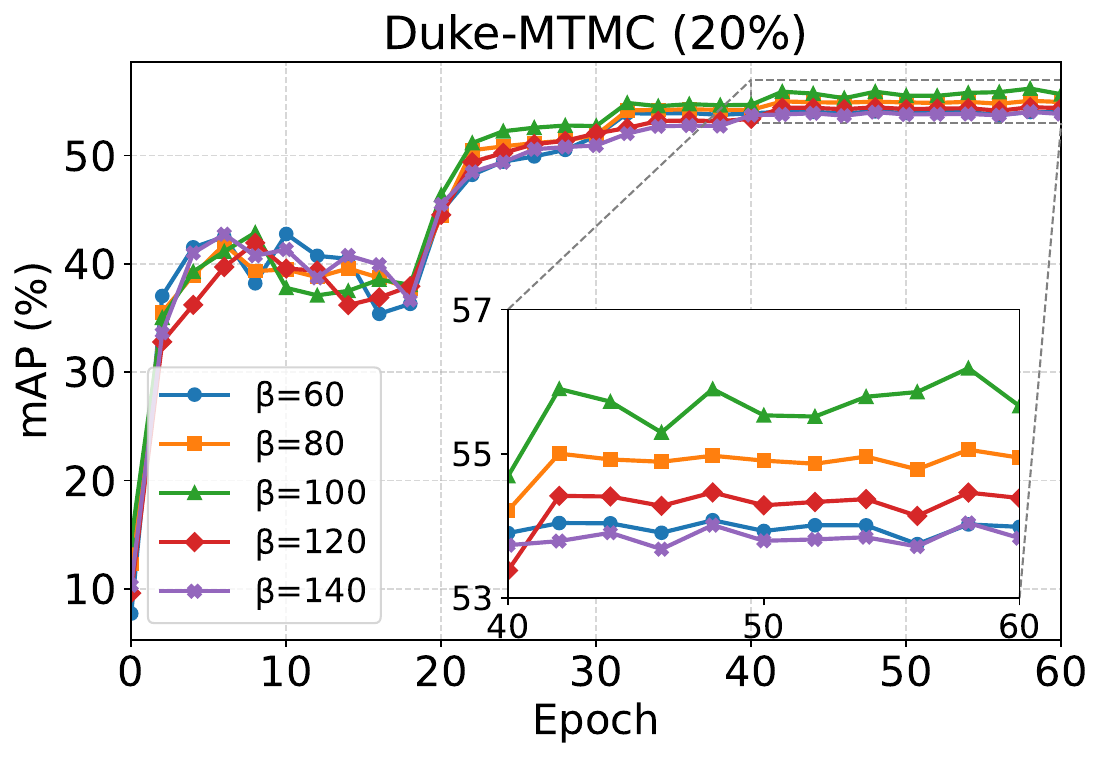}%
  \hspace{2pt}
  \includegraphics[width=0.23\linewidth, height=2.5cm]{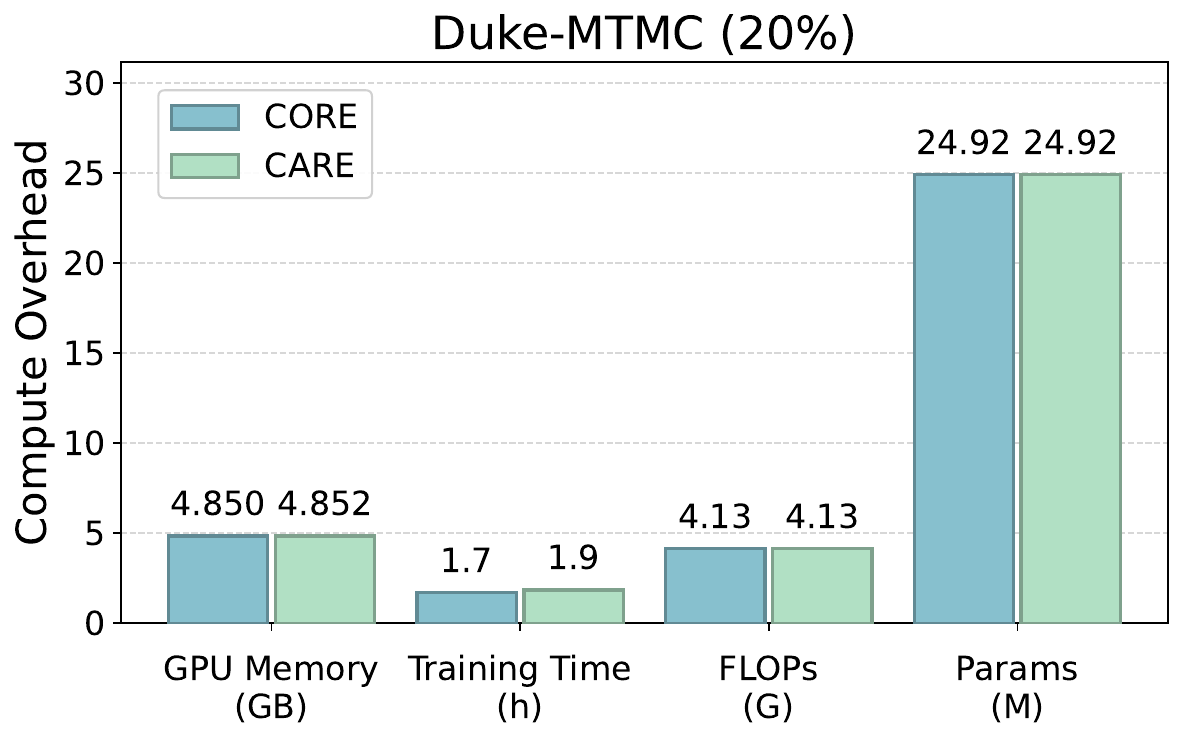}\\[\smallskipamount]
  
  \includegraphics[width=0.23\linewidth, height=2.5cm]{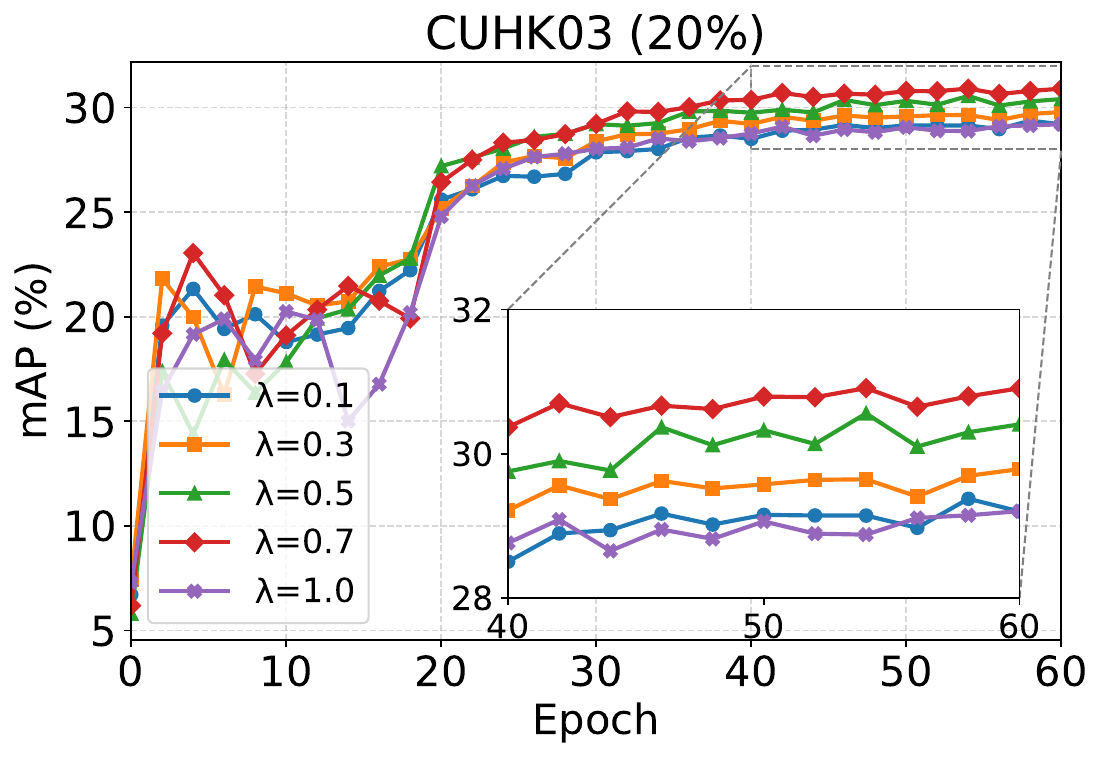}%
  \hspace{2pt}
  \includegraphics[width=0.23\linewidth, height=2.5cm]{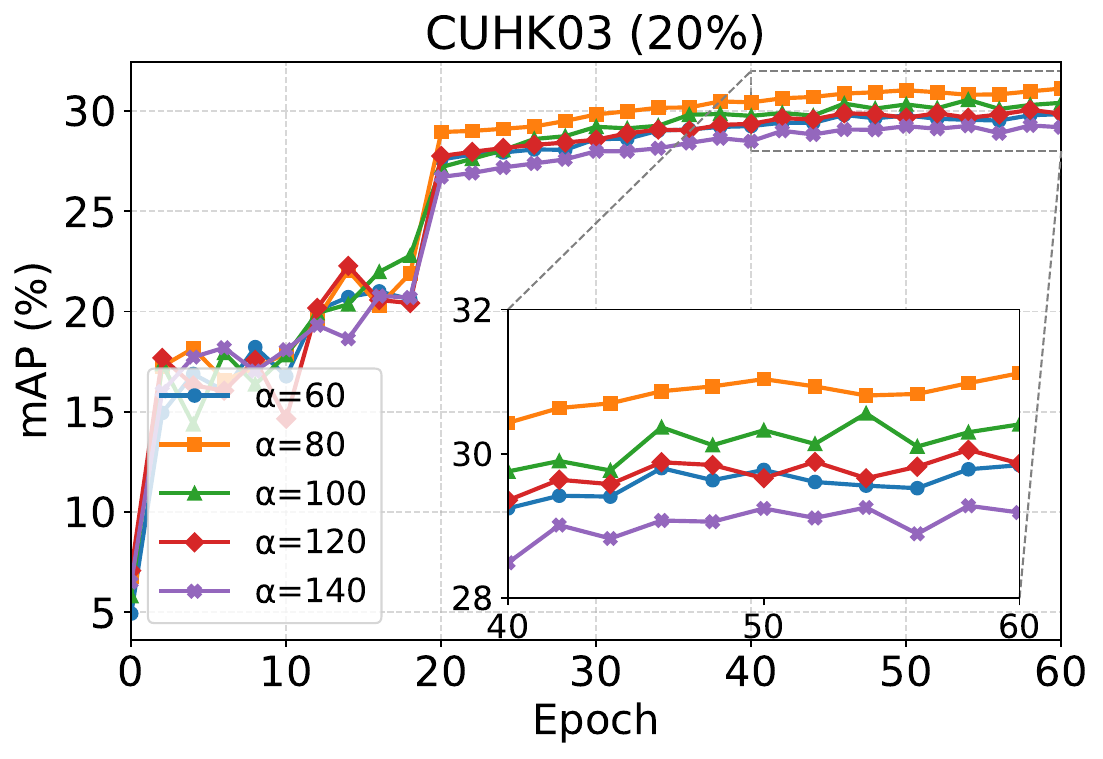}%
  \hspace{2pt}
  \includegraphics[width=0.23\linewidth, height=2.5cm]{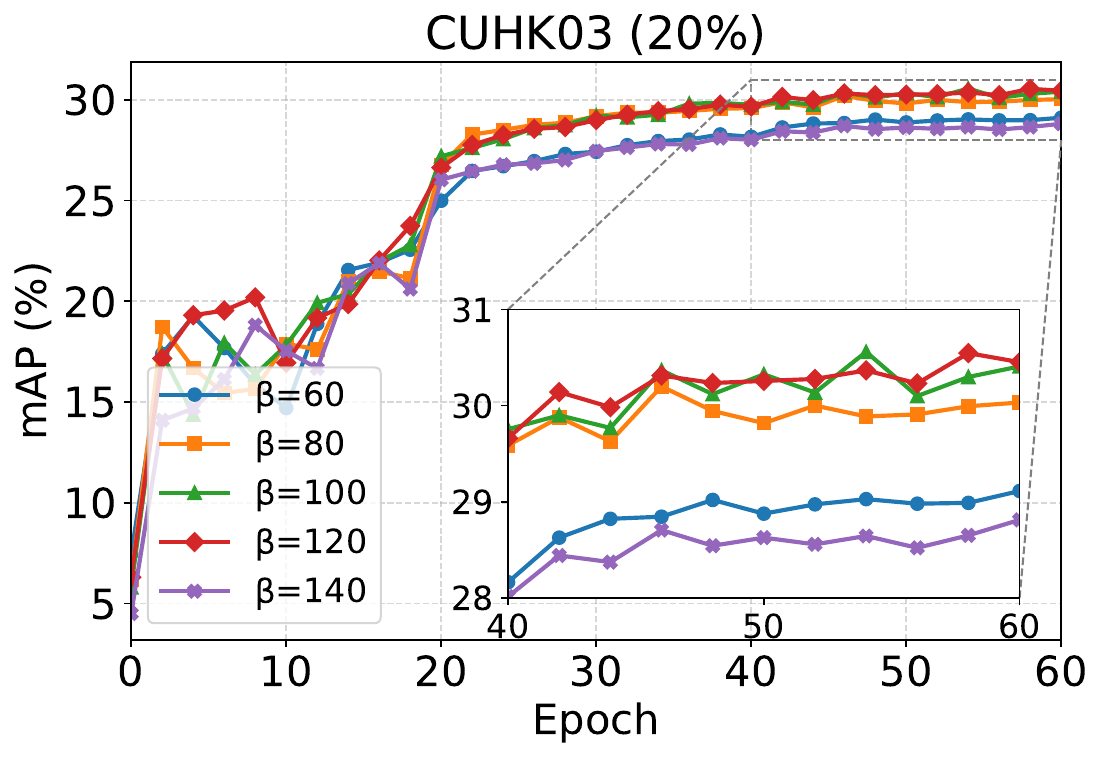}%
  \hspace{2pt}
  \includegraphics[width=0.23\linewidth, height=2.5cm]{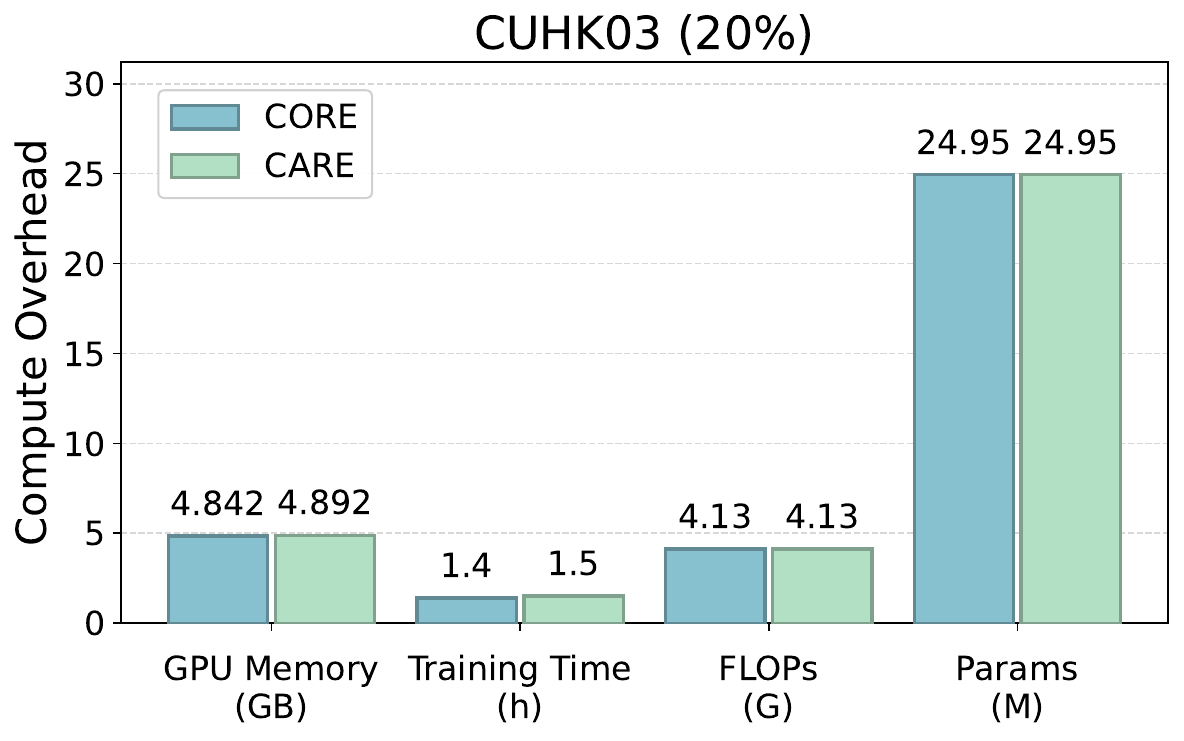}
  
  \caption{
  Ablation results of hyperparameters ($\lambda, \alpha, \beta$) and compute overhead under 20\% patterned noise ratio. 
  Each row shows results on one person Re-ID dataset: Market1501 (top), DukeMTMC-ReID (middle), and CUHK03 (bottom). From left to right, the columns detail: the ablation studies for hyperparameters \(\lambda\), \(\alpha\), and \(\beta\), followed by the compute overhead.}
  \vspace{-2.8mm}
  \label{fig:p_hyperparams_compact}
\end{figure*}

\begin{figure*}[htbp]
  \centering
  \resizebox{0.9\linewidth}{!}{
  \begin{tabular}{c @{\hskip 0.5em}
                  c @{\hskip 0.5em}
                  c @{\hskip 0.5em}
                  c @{\hskip 0.5em}
                  c}
    & \textbf{CORE (rand.)} & \textbf{CARE (rand.)} & \textbf{CORE (pat.)} & \textbf{CARE (pat.)} \\[4pt]

    \rotatebox{90}{\parbox{4cm}{\centering\textbf{Market1501}}} &
    \subcaptionbox{$V_c=0.92,\, V_a=5.24$}{\includegraphics[width=0.23\textwidth]{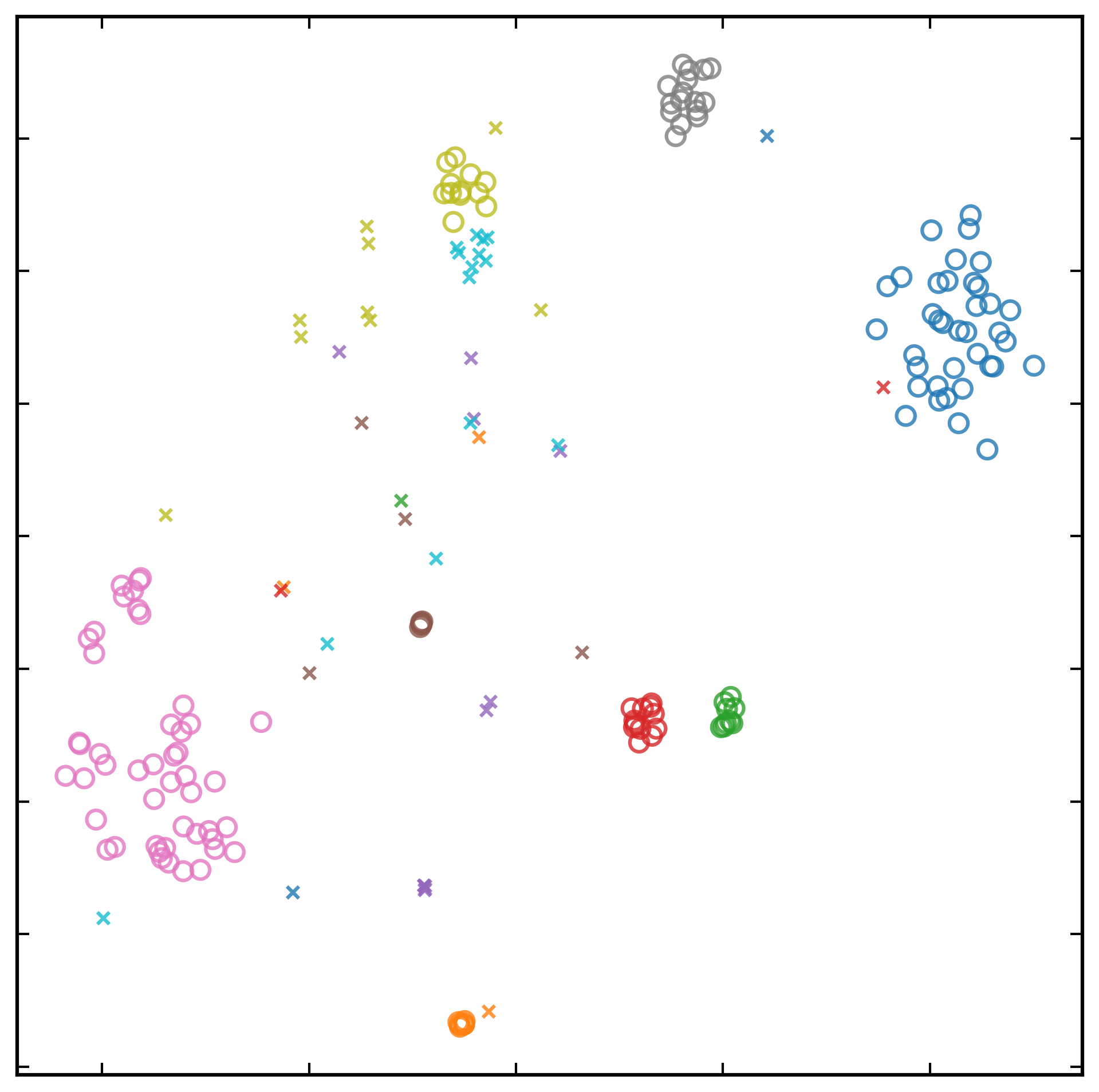}} &
    \subcaptionbox{$V_c=0.88,\, V_a=8.33$}{\includegraphics[width=0.23\textwidth]{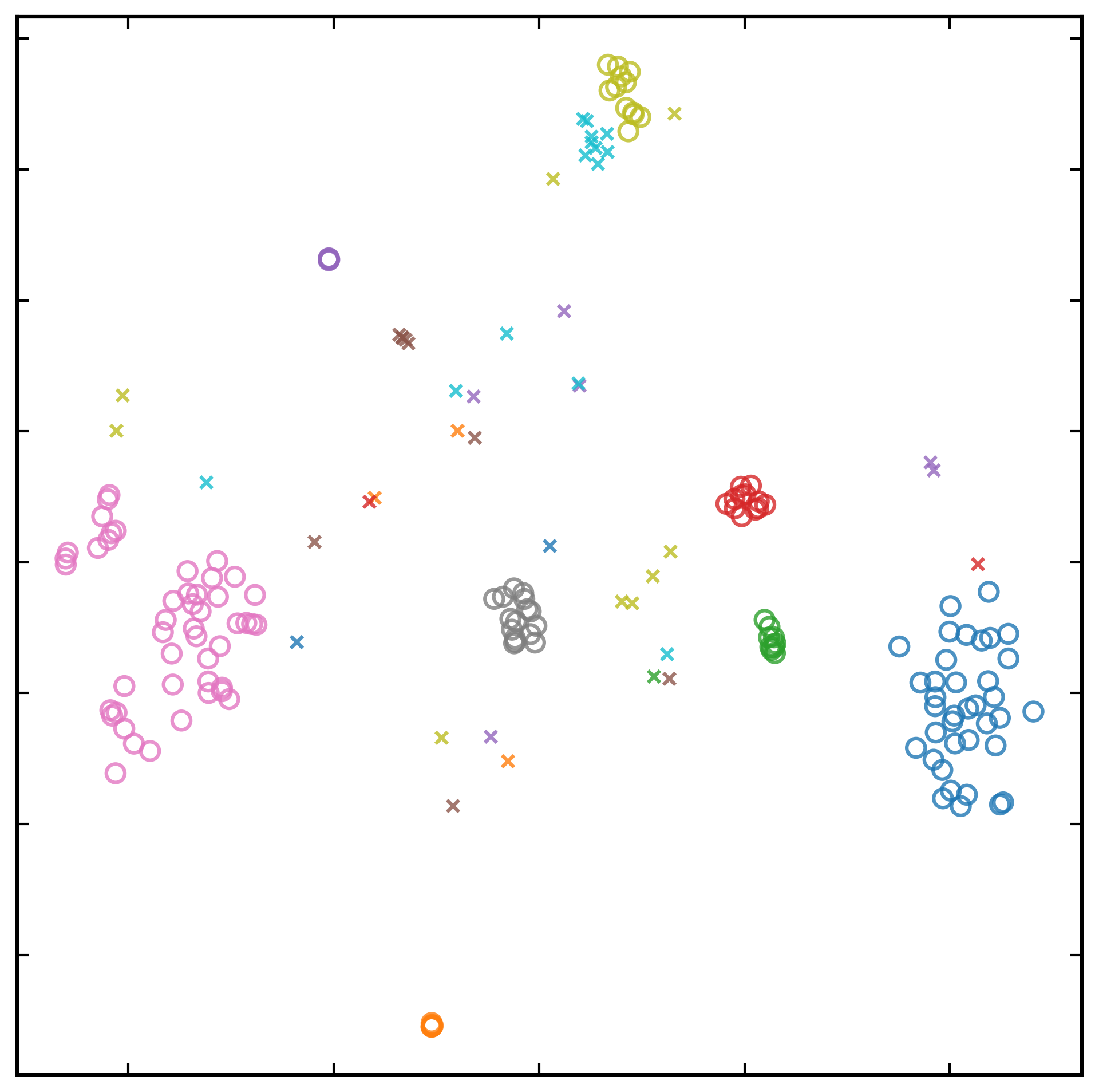}} &
    \subcaptionbox{$V_c=0.76,\, V_a=4.90$}{\includegraphics[width=0.23\textwidth]{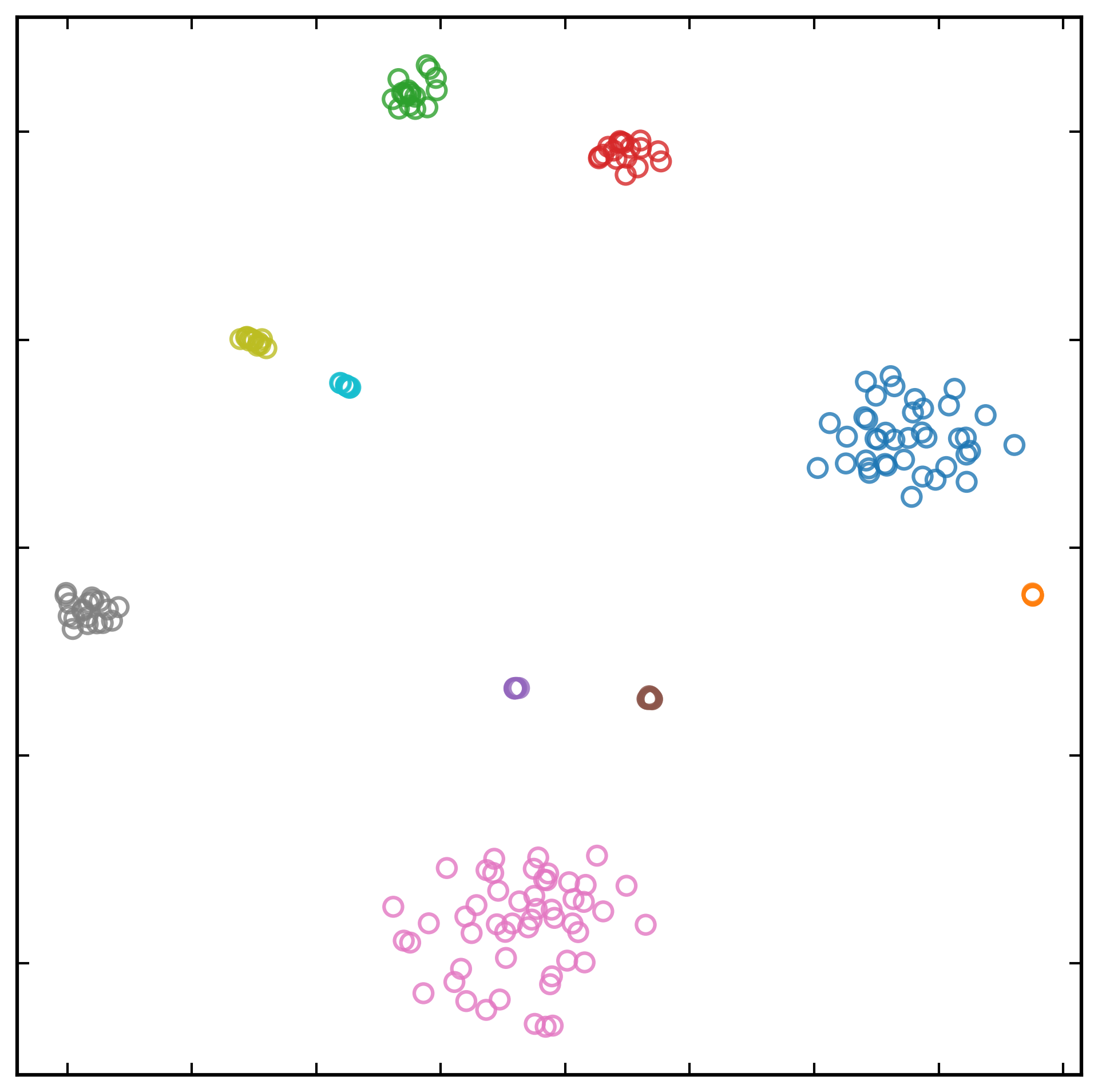}} &
    \subcaptionbox{$V_c=0.66,\, V_a=6.62$}{\includegraphics[width=0.23\textwidth]{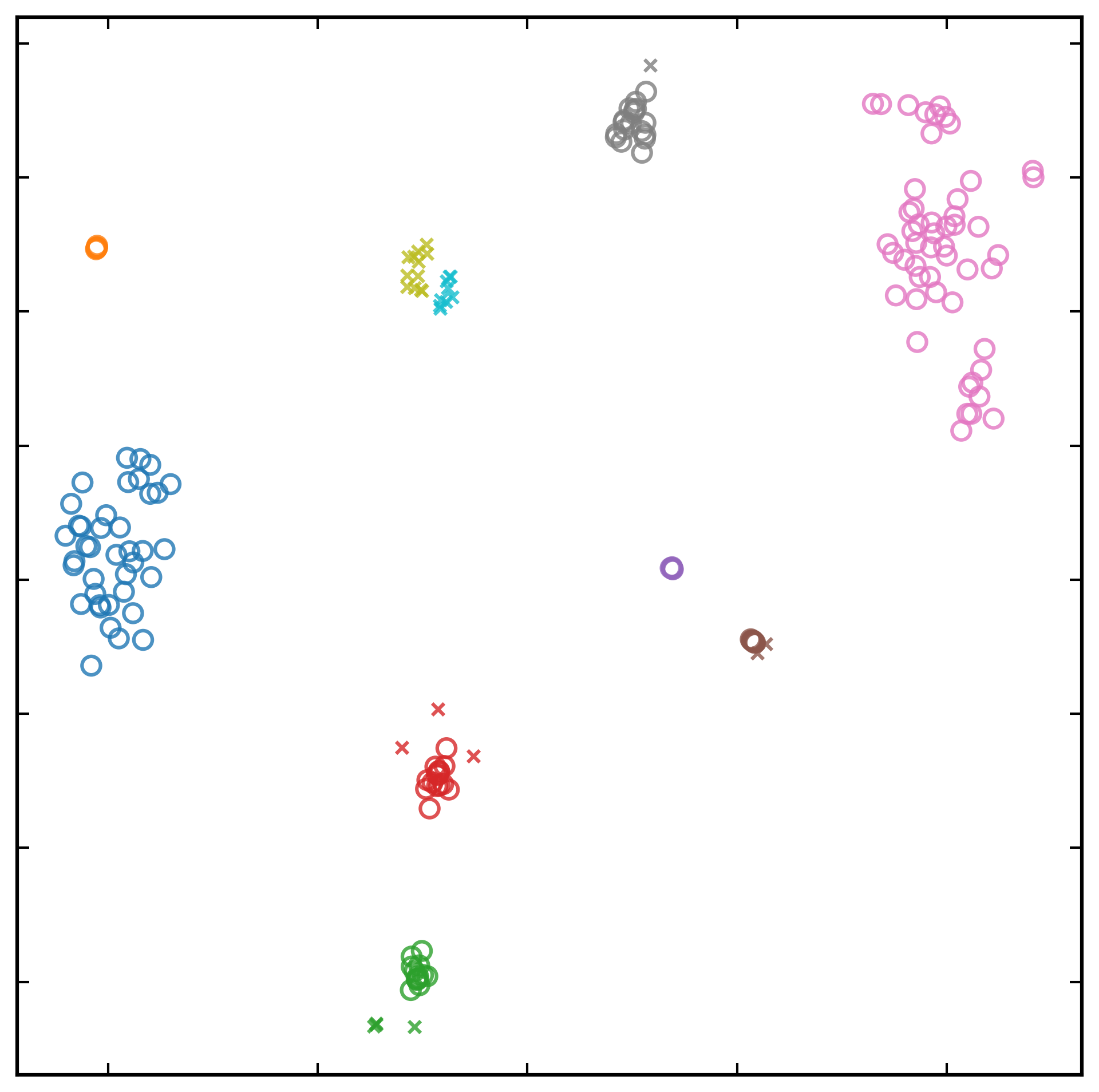}} \\[6pt]

    \rotatebox{90}{\parbox{4cm}{\centering\textbf{DukeMTMC-ReID}}} &
    \subcaptionbox{$V_c=0.78,\, V_a=5.65$}{\includegraphics[width=0.23\textwidth]{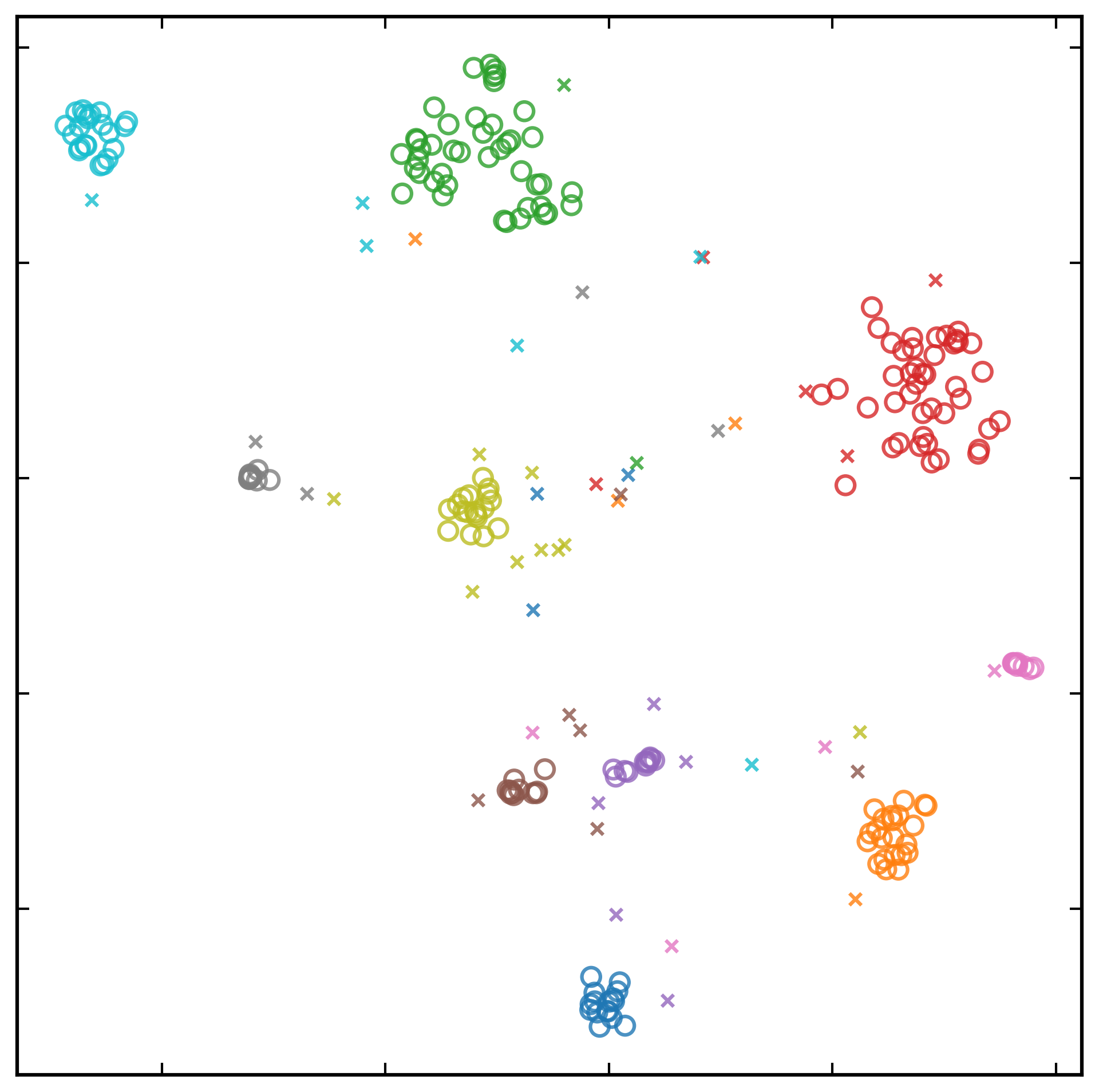}} &
    \subcaptionbox{$V_c=0.75,\, V_a=7.71$}{\includegraphics[width=0.23\textwidth]{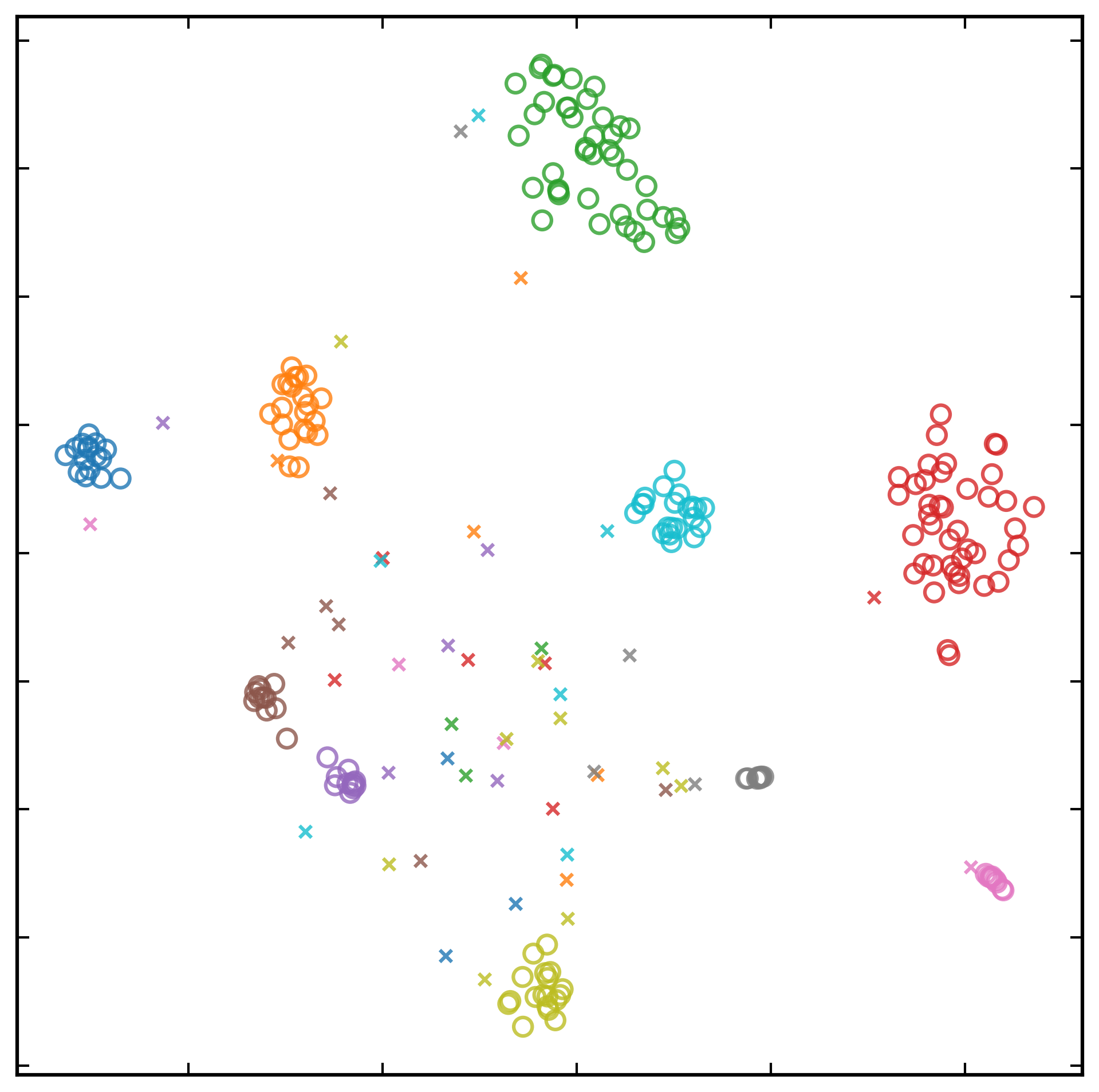}} &
    \subcaptionbox{$V_c=1.34,\, V_a=5.63$}{\includegraphics[width=0.23\textwidth]{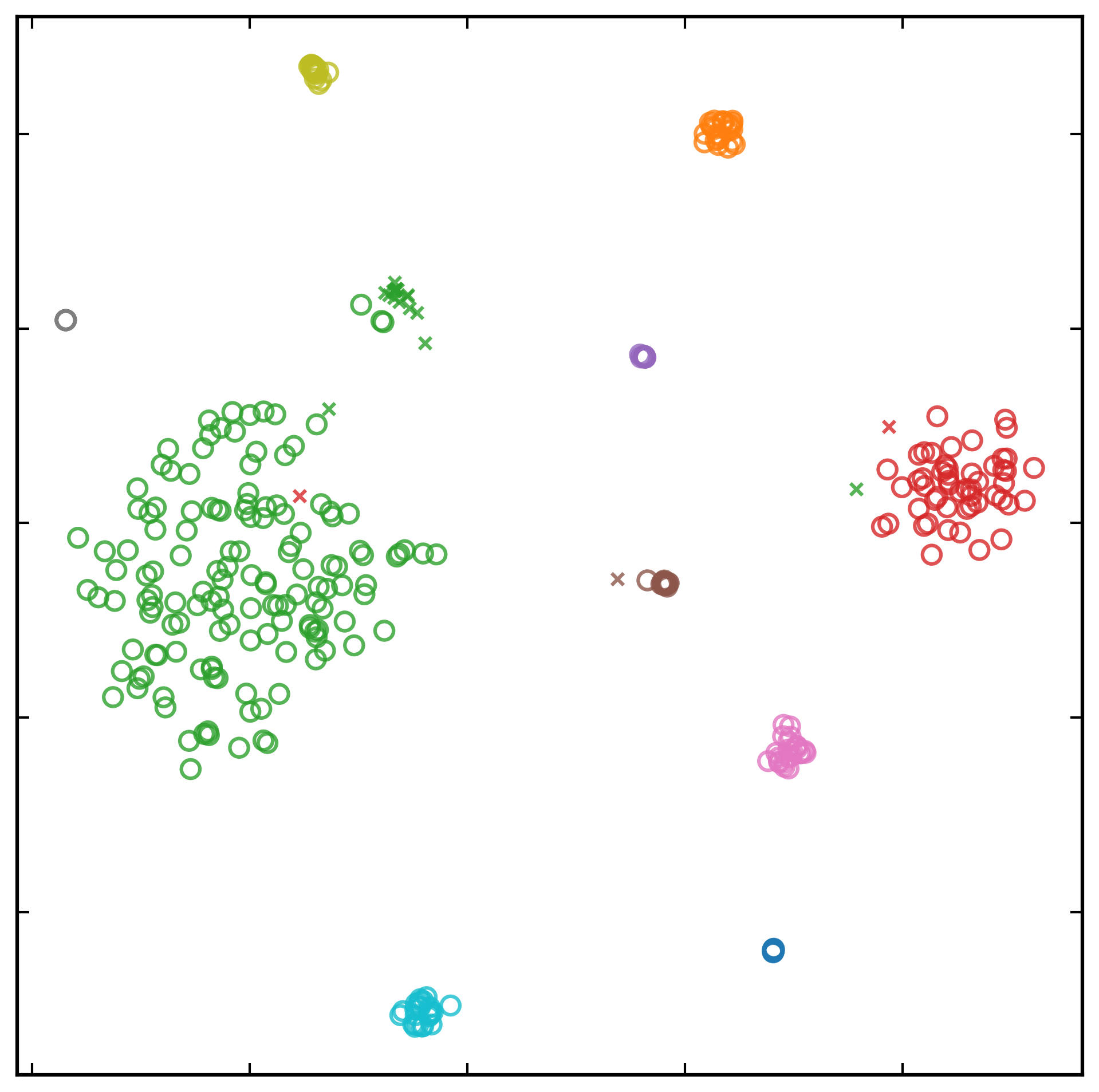}} &
    \subcaptionbox{$V_c=1.07,\, V_a=7.63$}{\includegraphics[width=0.23\textwidth]{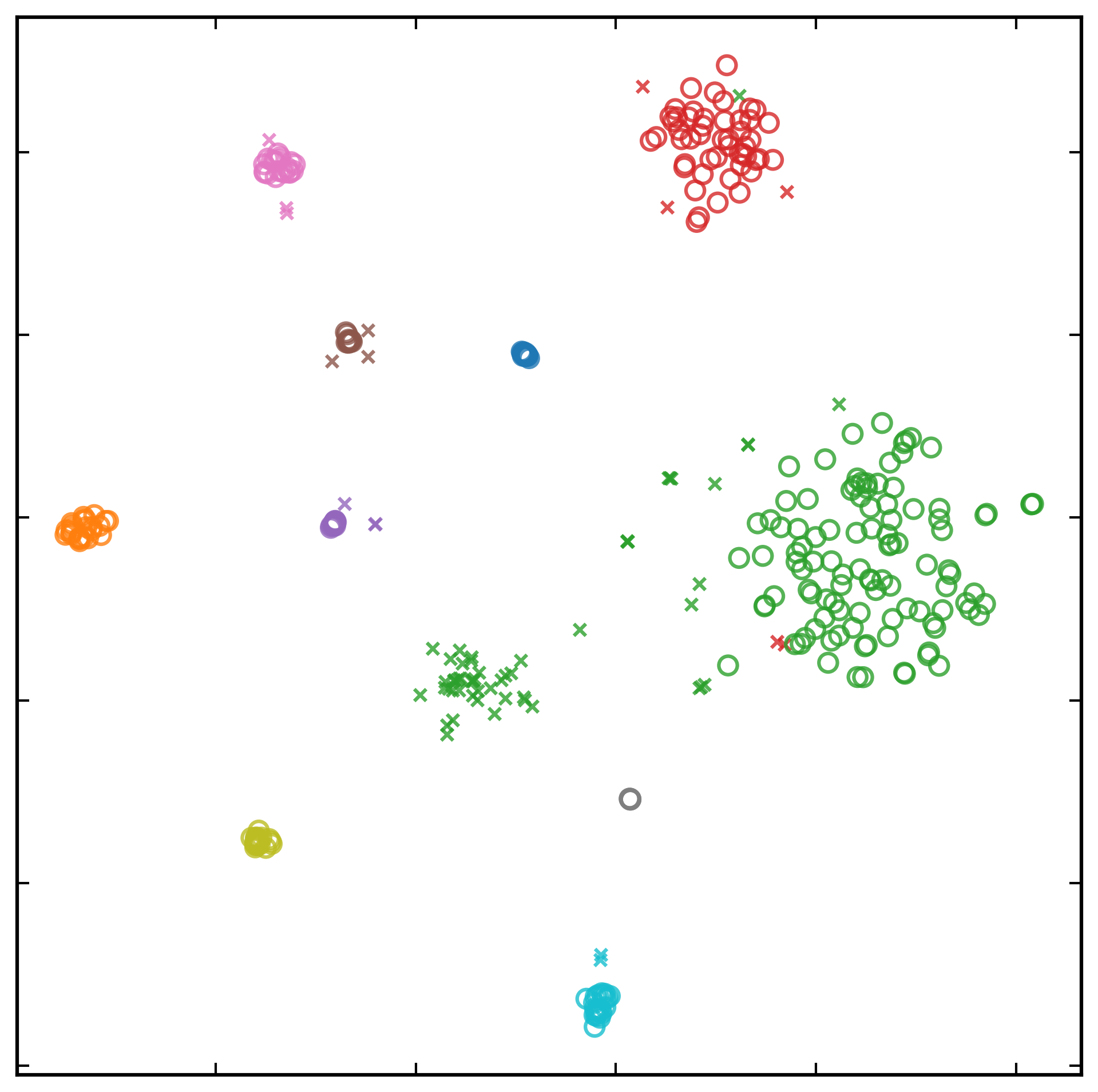}} \\[6pt]

    \rotatebox{90}{\parbox{4cm}{\centering\textbf{CUHK03}}} &
    \subcaptionbox{$V_c=0.49,\, V_a=3.86$}{\includegraphics[width=0.23\textwidth]{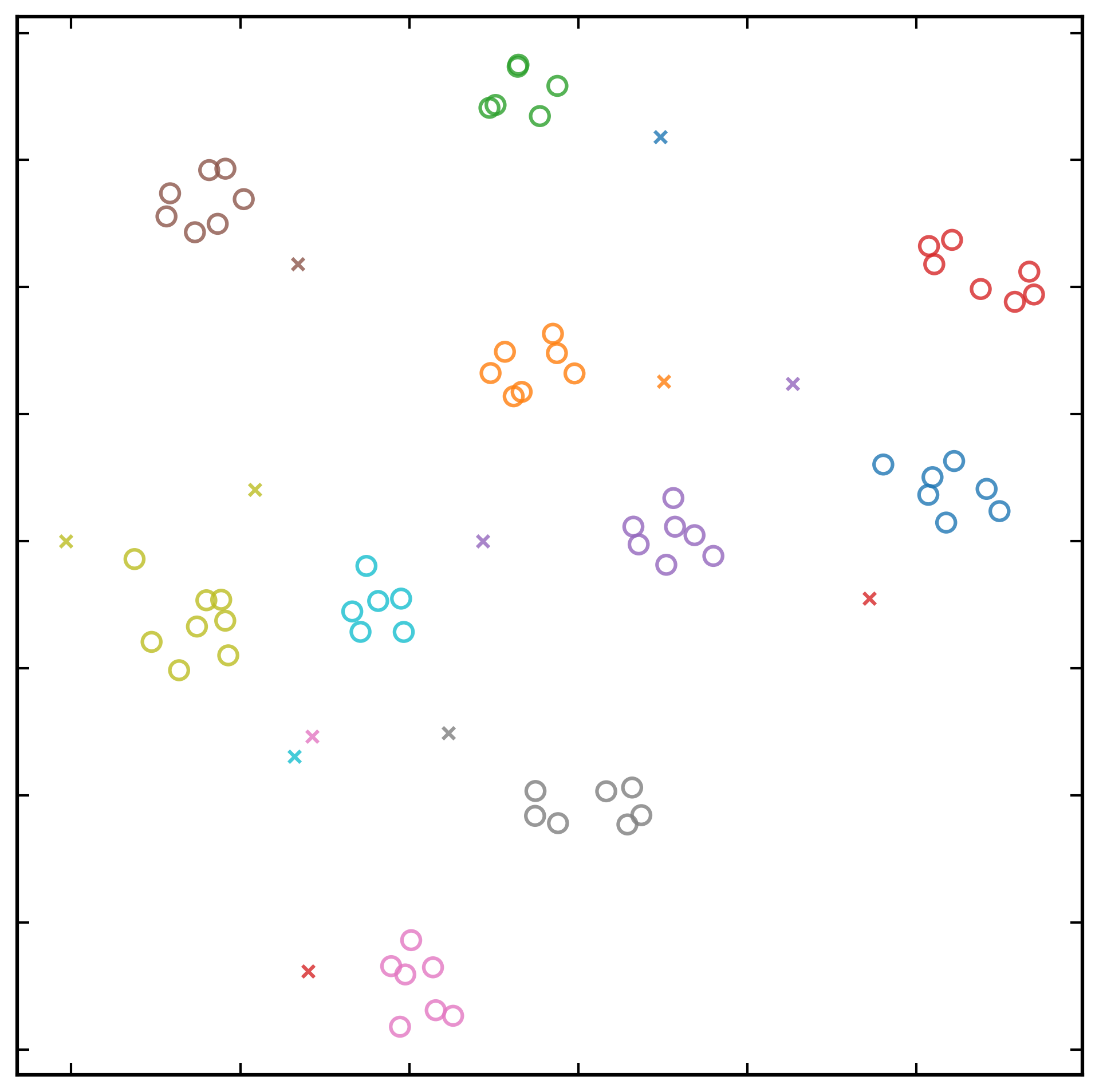}} &
    \subcaptionbox{$V_c=0.48,\, V_a=4.11$}{\includegraphics[width=0.23\textwidth]{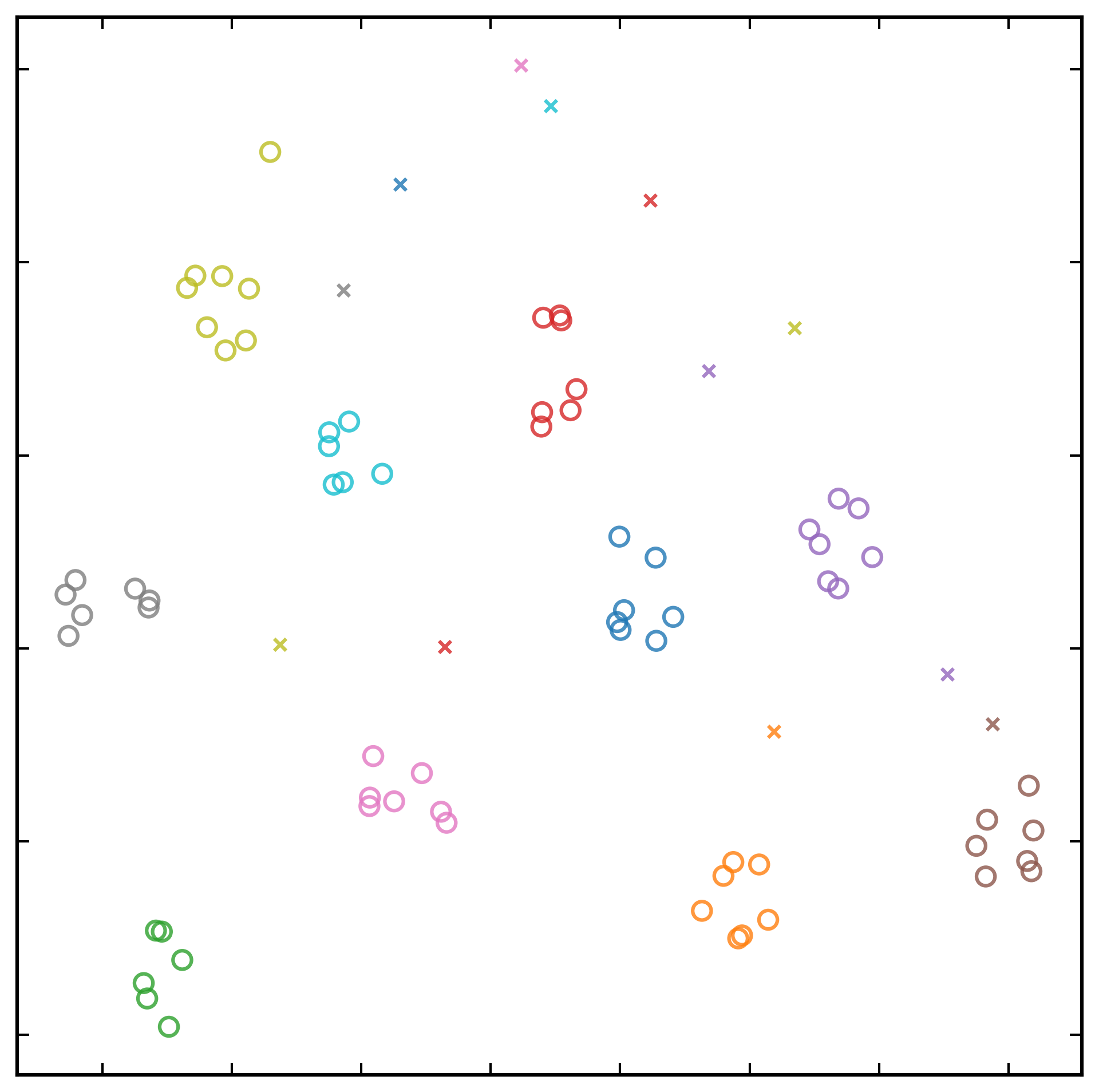}} &
    \subcaptionbox{$V_c=0.43,\, V_a=1.68$}{\includegraphics[width=0.23\textwidth]{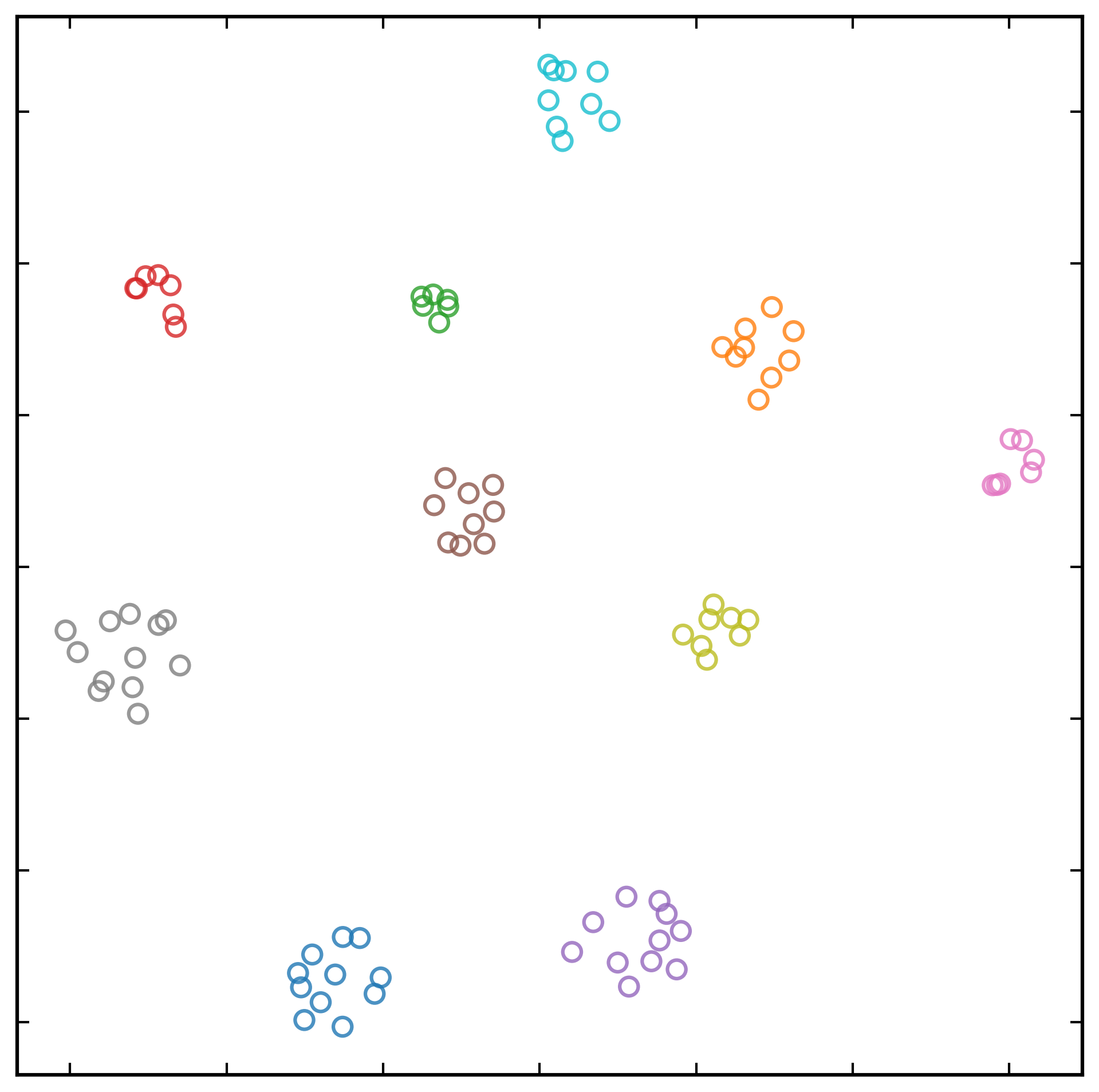}} &
    \subcaptionbox{$V_c=0.42,\, V_a=2.15$}{\includegraphics[width=0.23\textwidth]{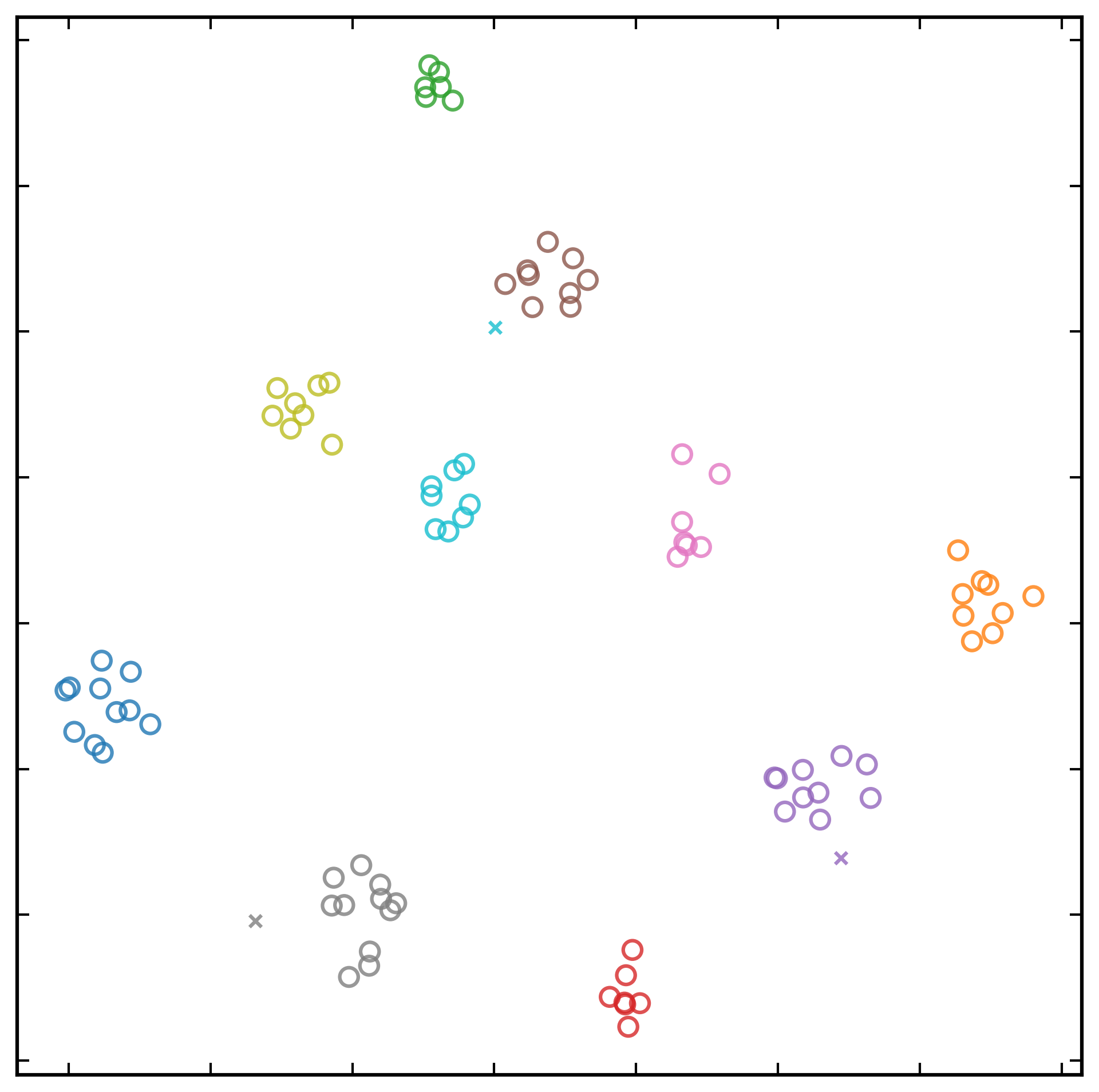}} \\
  \end{tabular}}
  \caption{The t-SNE visualizations of 10 identities for CORE and \textbf{CARE} under 20\% random and patterned ratios on three person Re-ID benchmarks.  The rows correspond to datasets (top: Market1501; middle: DukeMTMC-ReID; bottom: CUHK03). The columns correspond to CORE (random), \textbf{CARE} (random), CORE (patterned), and \textbf{CARE} (patterned). 
  Circles denote correctly labeled samples and crosses denote mislabeled ones. $V_c$ and $V_a$ quantify cluster compactness and mislabel dispersion, respectively.}
  \label{fig:tsne-3x4}
  \vspace{-2.5mm}
\end{figure*}

\subsubsection{Effect of Batch Size}
We further investigate the influence of training batch size under both random and patterned noise settings, as shown in Table~\ref{tab:results_batchsize}. 
Overall, excessively small batches (\textit{e.g.}, 8) consistently degrade performance. This is primarily due to high gradient variance and poor identity diversity within the mini-batch, which adversely affect the CAM and COSW mechanisms, thereby reducing their efficacy.
Meanwhile, a moderate batch size (\textit{i.e.}, 32) yields the most stable and competitive performance across all datasets. 
For instance, under 20\% random noise with a batch size of 32, the results achieve the best Rank-1/mAP scores of 90.4\%/70.5\%, 77.1\%/59.7\%, and 37.7\%/37.2\% on Market-1501, DukeMTMC-reID, and CUHK03, respectively. For patterned noise, when the batch size is 32, the performance is typically the optimal across the three datasets.
In addition, even under 50\% random noise, the results with a batch size of 32 also show nearly the best performance across the three datasets. 
A degradation in refinement-stage discrimination is also observed with overly large batches (\textit{e.g.}, 128). This stems from reduced parameter update frequency and weakened stochastic regularization, thereby impairing the model’s capacity to separate hard positives from noise.
This consistency holds across both random and patterned noise, with batch sizes between 16 and 32 yielding optimal performance. We therefore adopt 32 as the standard batch size, and recommend 16 as a memory-efficient alternative.

\subsubsection{Hyperparameter Sensitivity of $\lambda$, $\alpha$, and $\beta$}
As shown in Fig.~\ref{fig:hyperparams_compact}, we assess the hyperparameter sensitivity of \(\lambda\), \(\alpha\), and \(\beta\) on Market1501, DukeMTMC-ReID, and CUHK03 under 20\% random noise. For the hyperparameter \(\lambda\), which controls the trade-off between \(\mathcal{L}_{\mathrm{ENLL}}\) and \(\mathcal{L}_{\mathrm{KL}}^{(\mathrm{Dir})}\) in Eq. (\ref{eq:ecl_final}), a moderate setting \(\lambda=0.5\) consistently yields the best balance between fitting noisy supervision and maintaining calibrated uncertainty. When \(\lambda\) is too small, for example \(\lambda=0.1\), the regularization effect is negligible and the model tends to overfit spurious labels; conversely, an excessively large \(\lambda\), for example \(\lambda=1.0\), overemphasizes distribution alignment and degrades retrieval performance. The two CAM coefficients \(\alpha\) and \(\beta\) govern the instantaneous composite angular margin in Eq.~(\ref{eq:cam_inst}) and thereby the separation between clean but hard-to-learn and noisy samples. Across the three datasets, we observe a coherent trend: very small values, such as \(\alpha = \beta = 60\), weaken either the margin or the top-\(k\) ambiguity term and reduce noise discrimination, while very large values, such as \(\alpha = \beta = 140\), allow one term to dominate and bias scores, harming final mAP. Empirically, intermediate settings around \(\alpha=\beta=100\) produce the most stable and competitive results on Market1501, DukeMTMC-ReID, and CUHK03 under random noise, we thus adopt these defaults in our comparisons.

Under patterned noise, the overall behavior is similar but with slightly increased sensitivity to \(\alpha\) and \(\beta\) on datasets that exhibit richer intra-identity variation, as shown in Fig.~\ref{fig:p_hyperparams_compact}. In particular, \(\lambda=0.5\) remains the best trade-off between calibration and fidelity, while \(\alpha\) and \(\beta\) values near 100 again tend to maximize mAP across the three datasets. The patterned corruptions can create locally compact yet incorrect clusters; therefore, overly small CAM weights fail to separate mislabeled groups, whereas excessively large weights over-penalize difficult but correct samples. The mid-range setting around 100 provides a practical balance between these noise modes. Overall, the trends reported in Figs.~\ref{fig:hyperparams_compact} and~\ref{fig:p_hyperparams_compact} indicate that the \textbf{CARE}'s performance is robust to modest hyperparameter variations while being sensitive to extreme settings.

\subsubsection{Compute Overhead}
As shown in the rightmost column of Figs.~\ref{fig:hyperparams_compact} and \ref{fig:p_hyperparams_compact}, we measured the practical compute overhead of \textbf{CARE} versus the CORE baseline across Market1501, DukeMTMC-ReID, and CUHK03 under both random and patterned noises. 
The results show that the \textbf{CARE} introduces negligible architectural overhead. The CORE network remains unchanged ($\sim$25M parameters, $\sim$4.1G FLOPs). The only measurable costs are a very small increase in GPU memory (generally tens to hundreds of MB) and a modest lengthening of training time, attributable to the lightweight calibration and refinement modules. Since these do not increase forward/backward complexity and are substantially outweighed by the robustness improvements, the \textbf{CARE} is deemed highly practical for common person Re-ID setups.

\subsubsection{Hyperparameter Tuning Analysis}
To ensure reproducible and efficient hyperparameter tuning, the following procedure is recommended. First, we should perform a coarse logarithmic sweep across \(\alpha\) and \(\beta\) (candidate set: \(\{0,\,0.1,\,0.5,\,1,\,10,\,50,\,100,\,500,\,1000,\,5000\}\)) and \(\lambda\) (\(\{0.1,\,0.25,\,0.5,\,0.75,\,1.0\}\)) to identify a stable performance region. Second, we then conduct a fine-grained search within a narrowed neighborhood (\textit{e.g.}, \(\alpha,\beta\in[60,140]\) and \(\lambda\in[0.25,0.75]\)). In practice, since \(\alpha\) and \(\beta\) exhibit similar influence, they can be tied to reduce the search space. We suggest \(\lambda=0.5\) and \(\alpha=\beta=100\) as robust defaults, offering a reliable trade-off between noise resilience and informative sample retention. If the computational cost is limited, fixing \(\alpha=\beta\) and tuning only \(\lambda\) preserves the majority of performance improvements while substantially lowering the hyperparameter tuning cost.

\begin{figure}[t]
  \centering
  \includegraphics[width=0.97\linewidth,height=8cm]{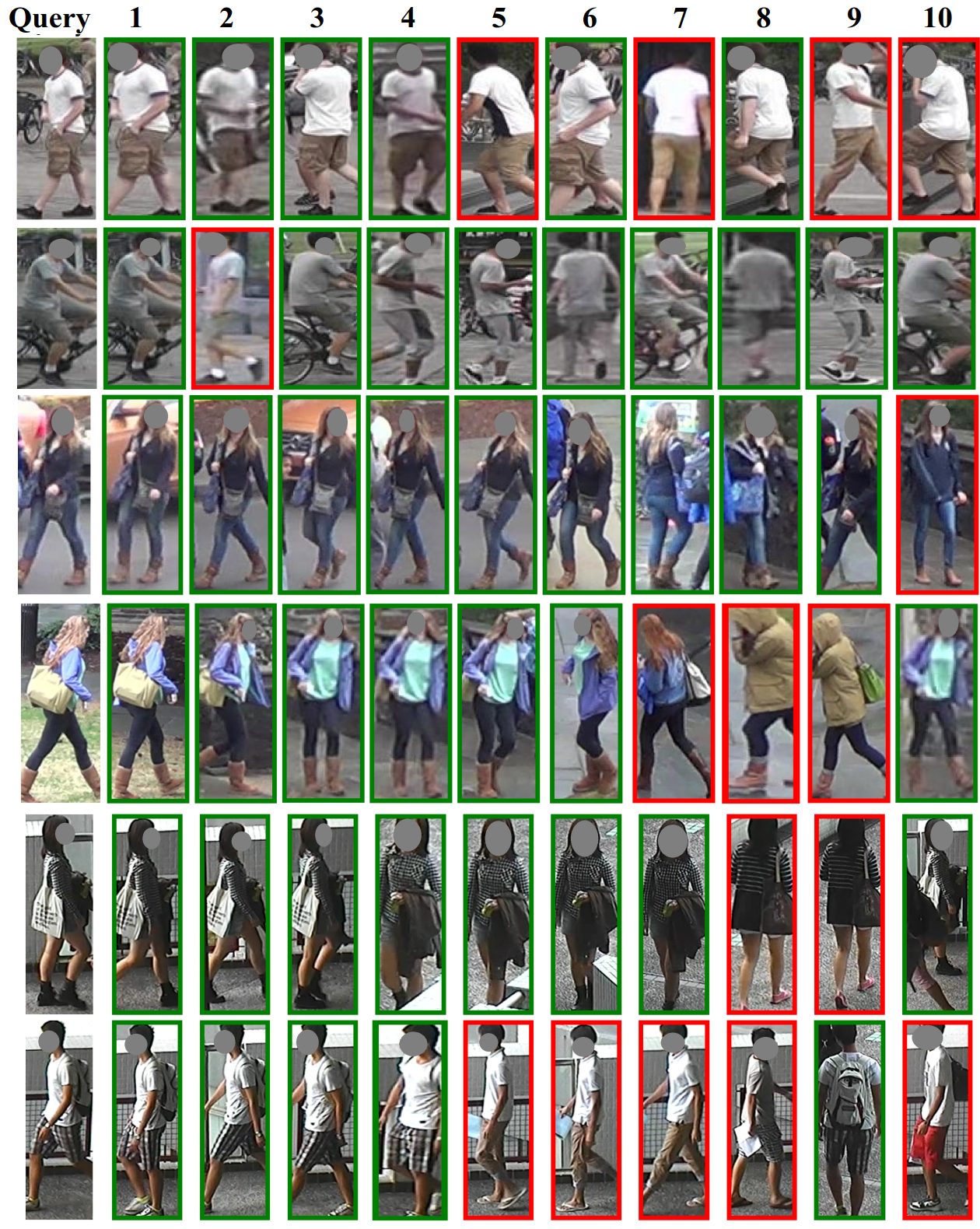} 
    \caption{Top-10 retrieved results on three person Re-ID datasets, \textit{i.e.}, 
  Rows 1–2: Market1501; rows 3–4: DukeMTMC-ReID; rows 5–6: CUHK03.
  Each row shows one query (\textbf{left}) and its top-10 retrieved results.
  The green and red bounding boxes denote correct and wrong retrieval results, respectively.
  }
  \vspace{-4mm}
  \label{fig:ranklist}
\end{figure}

\subsection{Visualization Analysis}
\subsubsection{Statistical Distribution Analysis.} 
 
To quantify the clustering of correctly labeled samples and the dispersion of noisy samples across different training settings, we employ two evaluation metrics, namely $V_c$ and $V_a$, on the Market1501, DukeMTMC-ReID, and CUHK03 datasets under 20\% random and patterned noise ratios. $V_c$ quantifies the average intra-class variance of correctly labeled samples, where a lower value indicates tighter clustering (better performance). In contrast, $V_a$ measures the average variance between mislabeled samples and the centers of their true identities, with a higher value suggesting greater separation from incorrect clusters (better performance). As shown in Fig.~\ref{fig:tsne-3x4}, on Market1501 under random noise, our method \textbf{CARE} achieves $V_c$=0.88 and $V_a$=8.33, compared to CORE's values of $V_c$=0.92 and $V_a$=5.24.
Similarly, under patterned noise, \textbf{CARE} attains $V_c$=0.66 and $V_a$=6.62, outperforming CORE, which yields $V_c$=0.76 and $V_a$=4.90.
On DukeMTMC-ReID, $V_c$ is decreased from 0.78 to 0.75 while $V_a$ increases from 5.65 to 7.71 under random noise; similarly, $V_c$ decreases from 1.34 to 1.07 while $V_c$ changes from 5.63 to 7.63 under patterned noise. 
Notably, on CUHK03 under random noise, $V_c$ decreases from 0.49 to 0.48 while $V_a$ increases from 3.86 to 4.11. Similarly, under patterned noise, $V_c$ is reduced from 0.43 to 0.42, whereas $V_a$ shows a more pronounced improvement, rising from 1.68 to 2.15.
This indicates that under different noise settings, our proposed \textbf{CARE} method can more effectively concentrate on correctly labeled samples compared to the baseline CORE, while mislabeled samples are pushed away from the center.

\subsubsection{Retrieved Results Analysis}
Fig.~\ref{fig:ranklist} presents a visualization of the top ten retrieval results from the Market1501, DukeMTMC-ReID, and CUHK03 datasets. It can be observed that the incorrectly retrieved instances are highly similar to the correct instances, making them challenging even for human discrimination. Moreover, our method successfully identifies the majority of correct sample images, significantly improving retrieval accuracy. While a small number of wrongly matched instances may exist due to factors like similar clothing styles, camera views, or insufficient illumination, these visualization results still demonstrate the effectiveness of our method.

\section{Conclusion}
\label{sec:Conclusion}
In this paper, we propose the \underline{\textbf{CA}}libration-to-\underline{\textbf{RE}}finement (\textbf{CARE}) method, a two-stage calibration-to-refinement framework for noisy-label person Re-ID. In the \textbf{Calibration} stage, we introduced PEC, a Dirichlet-informed calibration that breaks softmax translation invariance and employs an evidential calibration loss to mitigate over-confident predictions on corrupted labels. In the \textbf{Refinement} stage, we presented EPR, which leverages the CAM metric to distinguish hard but correctly labeled positives from mislabeled instances in a hyperspherical embedding, and applies COSW to allocate samples according to CAM-derived certainty. Extensive experiments on Market1501, DukeMTMC-ReID, and CUHK03 under both random and patterned noise demonstrate that \textbf{CARE} consistently improves Rank-1 and mAP over the baseline and achieves the state-of-the-art methods. The ablation studies verify the complementary benefits of PEC and EPR, showing that the \textbf{Calibration} and \textbf{Refinement} stages jointly enhance robustness and convergence. Overall, \textbf{CARE} offers an effective and generalizable paradigm for improving metric learning under label corruption.
In terms of limitations, it can be seen that we mainly focus on single-modal scenarios with minimal illumination variations. Therefore, when addressing more complex open-world scenarios, our method may still have significant improvement room.


\bibliographystyle{IEEEtran}
\bibliography{references}

\newpage

\end{document}